\documentclass[twoside]{article}

\usepackage{aistats2020}

\usepackage[unicode=true,bookmarks=true,bookmarksnumbered=true,bookmarksopen=true,bookmarksopenlevel=1,breaklinks=false,pdfborder={0 0 0},pdfborderstyle={},backref=section,colorlinks=true]{hyperref}
\usepackage{booktabs}       % professional-quality tables
\usepackage{nicefrac}       % compact symbols for 1/2, etc.
\usepackage{microtype}      % microtypography

\usepackage{graphicx}
\usepackage{amsmath,amssymb,mathtools,bm,etoolbox}
\usepackage{caption}
\usepackage{subcaption}
\usepackage{xcolor}
\usepackage{wrapfig}
\usepackage[title]{appendix}

\definecolor{citecol}{rgb}{0,0.5,0}
\definecolor{linkcol}{rgb}{0.8,0,0}
\definecolor{blcolor}{rgb}{0,0,0.65}
\hypersetup{
    colorlinks,
    linkcolor={blcolor},
    citecolor={blcolor},
    urlcolor={blcolor}
}

\usepackage{tikz}
\usetikzlibrary{fit,positioning}

\usepackage[round,authoryear]{natbib}

\usepackage{algorithm, algpseudocode}
\floatname{algorithm}{Algorithm}

\algtext*{EndFunction}

\newcommand{\remove}[1]{}

\newcommand{\bfv}[1]{{\bf #1}}

\newcommand{\deriv}[2]{\frac{\textup{d}#1\hfill}{\textup{d}#2\hfill}}

\newcommand{\pderiv}[3]{\left.\frac{\partial#1}{\partial#2\hfill}\right\vert_{#3}}
\newcommand{\pderivw}[2]{\frac{\partial#1}{\partial#2}}
\newcommand{\expect}[1]{\mathbb{E}\left[#1\right]}
\newcommand{\expectw}[2]{\mathbb{E}_{#1}\left[#2\right]}
\newcommand{\variance}[1]{\mathbb{V}\left[#1\right]}
\newcommand{\variancew}[2]{\mathbb{V}_{#1}\left[#2\right]}

\newcommand{\p}[1]{p\left(#1\right)}
\newcommand{\pind}[2]{p_{#1}\left(#2\right)}
\newcommand{\dpx}{\deriv{p}{\bfv{x}}}

\title{A unified view of \\likelihood ratio and reparameterization gradients
\\and an optimal importance sampling scheme}

\author{
  Paavo Parmas%\thanks{Use footnote for providing further
  \\
  Neural Computation Unit\\
  Okinawa Institute of Science and Technology Graduate University\\
  Okinawa, Japan \\
  \texttt{paavo.parmas@oist.jp} \\
}

\begin{document}

%\maketitle

\twocolumn[

\aistatstitle{A unified view of likelihood ratio and
  reparameterization gradients \\and an optimal importance sampling
  scheme}

\aistatsauthor{ Paavo Parmas$^*$ \And Masashi Sugiyama } %  \And Author 2 \And  Author 3 

\aistatsaddress{Okinawa Institute of Science and Technology \\
  \texttt{paavo.parmas@oist.jp} \And
  RIKEN and The University of Tokyo\\\texttt{sugi@k.u-tokyo.ac.jp}
}] % 1 \And  Institution 2 \And Institution 3 } ]

\begin{abstract}

  Reparameterization (RP) and likelihood ratio (LR) gradient
  estimators are used throughout machine and reinforcement learning;
  however, they are usually explained as simple mathematical tricks
  without providing any insight into their nature. We use a first
  principles approach to explain LR and RP, and show a connection
  between the two via the divergence theorem. The theory motivated us
  to derive optimal importance sampling schemes to reduce LR gradient
  variance. Our newly derived distributions have analytic probability
  densities and can be directly sampled from.  The improvement for
  Gaussian target distributions was modest, but for other
  distributions such as a Beta distribution, our method could lead to arbitrarily large
  improvements, and was crucial to obtain competitive performance in
  evolution strategies experiments.

\end{abstract}

\section{Introduction}
\label{intro}
Both likelihood ratio (LR) gradients
\citep{glynn1990likelihood,williams1992reinforce} and
reparameterization (RP) gradients
\citep{rezende2014stochasticBP,kingma2013autoencoder} can be used to
obtain unbiased estimates of the gradient of an expectation w.r.t. the
parameters of the distribution:
$\deriv{}{\theta}\expectw{x\sim\p{x;\theta}}{\phi(x)}$. This problem
is fundamental in machine learning \citep{mcgradrev}, and the gradients are
used for optimization in a wide range of tasks
\citep{schulman2015stocgraph,weber2019credit,parmas2018total}, e.g. reinforcement learning
\citep{sutton1998reinforcement,schulman2015trpo,schulman2017ppo,sutton2000policy,peters2008polgrad}, stochastic variational inference
\citep{hoffman2013stochastic} and evolutionary algorithms
\citep{wierstra2008natural,salimans2017oaies,ha2018worldmodels,conti2018improving}.

The LR gradient is usually derived as
$\deriv{}{\theta}\expectw{x\sim\p{x;\theta}}{\phi(x)} =
\int\deriv{\p{x;\theta}}{\theta}\phi(x)\textup{d}x =
\int\p{x;\theta}\frac{\deriv{\p{x;\theta}}{\theta}}{\p{x;\theta}}\phi(x)\textup{d}x =
\int\p{x;\theta}\deriv{\log\p{x;\theta}}{\theta}\phi(x)\textup{d}x =
\expectw{x\sim\p{x;\theta}}{\deriv{\log\p{x;\theta}}{\theta}\phi(x)}$. On the other
hand, the RP gradient is derived by defining a mapping
$g(\epsilon) = x$, where $\epsilon$ is sampled from a fixed simple
distribution, but $x$ ends up being sampled from the desired
distribution. For example, if  $x$ is Gaussian
$x\sim \mathcal{N}(\mu,\sigma)$, then the required mapping is
$g(\epsilon) = \mu +\sigma\epsilon$, where
$\epsilon \sim \mathcal{N}(0,1)$, and the RP gradient is derived as
$\deriv{}{\theta}\expectw{x\sim\p{x;\theta}}{\phi(x)} =
\deriv{}{\theta}\expectw{\epsilon\sim\mathcal{N}(0,1)}{\phi\left(g(\epsilon)\right)}
= \expectw{\epsilon\sim\mathcal{N}(0,1)}{\deriv{\phi\left(g(\epsilon)\right)}{\theta}}=
\expectw{\epsilon\sim\mathcal{N}(0,1)}{\deriv{g}{\theta}\deriv{\phi\left(g(\epsilon)\right)}{g}}$,
where $\theta=[\mu,\sigma]$, $\deriv{g}{\mu} = 1$, $\deriv{g}{\sigma} = \epsilon$ and
$\deriv{\phi\left(g(\epsilon)\right)}{g} = \deriv{\phi\left(x\right)}{x}$.

%\footnote{Note that $\deriv{g}{\mu} = 1$, $\deriv{g}{\sigma} = \epsilon$ and
%$\deriv{\phi\left(g(\epsilon)\right)}{g} = \deriv{\phi\left(x\right)}{x}$.}

What do these derivations mean, and what is the relationship between
the methods?  We give two possible answers to this question in
Secs.~\ref{boxtheor} and \ref{flowtheor}, then explain that the LR
gradient is the unique unbiased estimator that weights the function
values $\phi(x)$, and motivate importance sampling from a different distribution
$q(x)$ to reduce LR gradient
variance. Our optimal importance sampling scheme is reminiscent of the
optimal reward baseline for reducing LR gradient variance
\citep{weaver2001optimalbaseline} (App.~\ref{lrbasicsapp}), but our result is
orthogonal, and can be combined with such prior methods.

% Why these gradients are important. The typical derivation and explanation.
%I have a better view. Using the stuff

\paragraph{Further background and related work:}

The variance of LR and RP gradients has been of central importance in
their research. Typically, RP is said to be more accurate and scale
better with the sampling dimension
\citep{rezende2014stochasticBP}---this claim is also backed by theory
\citep{xu2018rpvar,nesterov2017randomtheory}; however, {\bf there is
  no guarantee that RP outperforms LR}. In particular, for multimodal
$\phi(x)$ \citep{gal2016uncertainty} or chaotic systems \citep{pipps},
LR can be arbitrarily better than RP (e.g., the latter showed that LR
can be $10^6$ more accurate in practice). Moreover, RP is not directly
applicable to discrete sampling spaces, but requires continuous
relaxations
\citep{maddison2016concrete,jang2016categorical,tucker2017rebar}. Differentiable
RP is also not always possible, but implicit RP gradients have
increased the number of usable distributions
\citep{figurnov2018implicitRP}. Techniques for variance reduction have
been extensively studied, including control variates/baselines
\citep{greensmith2004cv,grathwohl2017bpthroughvoid,tucker2018mirage,gu2015muprop,geffner2018using,gu2016q}
as well as Rao-Blackwellization
\citep{aueb2015local,ciosek2018expected,asadi2017meanac}. One can also
combine the best of both LR and RP gradients by dynamically
reweighting them \citep{pipps,metz2019und}. Importance sampling for
reducing LR gradient variance was previously considered in variational
inference \citep{ruiz2016impinvarinf}, but they proposed to sample from
the same distribution while tuning the variance, whereas in our work
we derive an optimal distribution. In reinforcement learning,
importance sampling has been studied for sample reuse via off-policy
policy evaluation
\citep{thomas2016offpolicy,jiang2016doubly,gu2017interpolatedpol,munos2016safe,jie2010connection},
but modifying the policy to improve gradient accuracy has not been
considered. The flow theory in Sec.~\ref{flowtheor} was concurrently
derived by \citet{jankowiak2018rpflow}, but their work focused on
deriving new RP gradient estimators, and they do not discuss the
duality. Our derivation is also more visual.

\section{A probability ``boxes" view of LR and RP gradients}
\label{boxtheor}

\begin{figure*}[h]
        \centering
	\begin{subfigure}{.49\textwidth}
		\includegraphics[width=\textwidth]{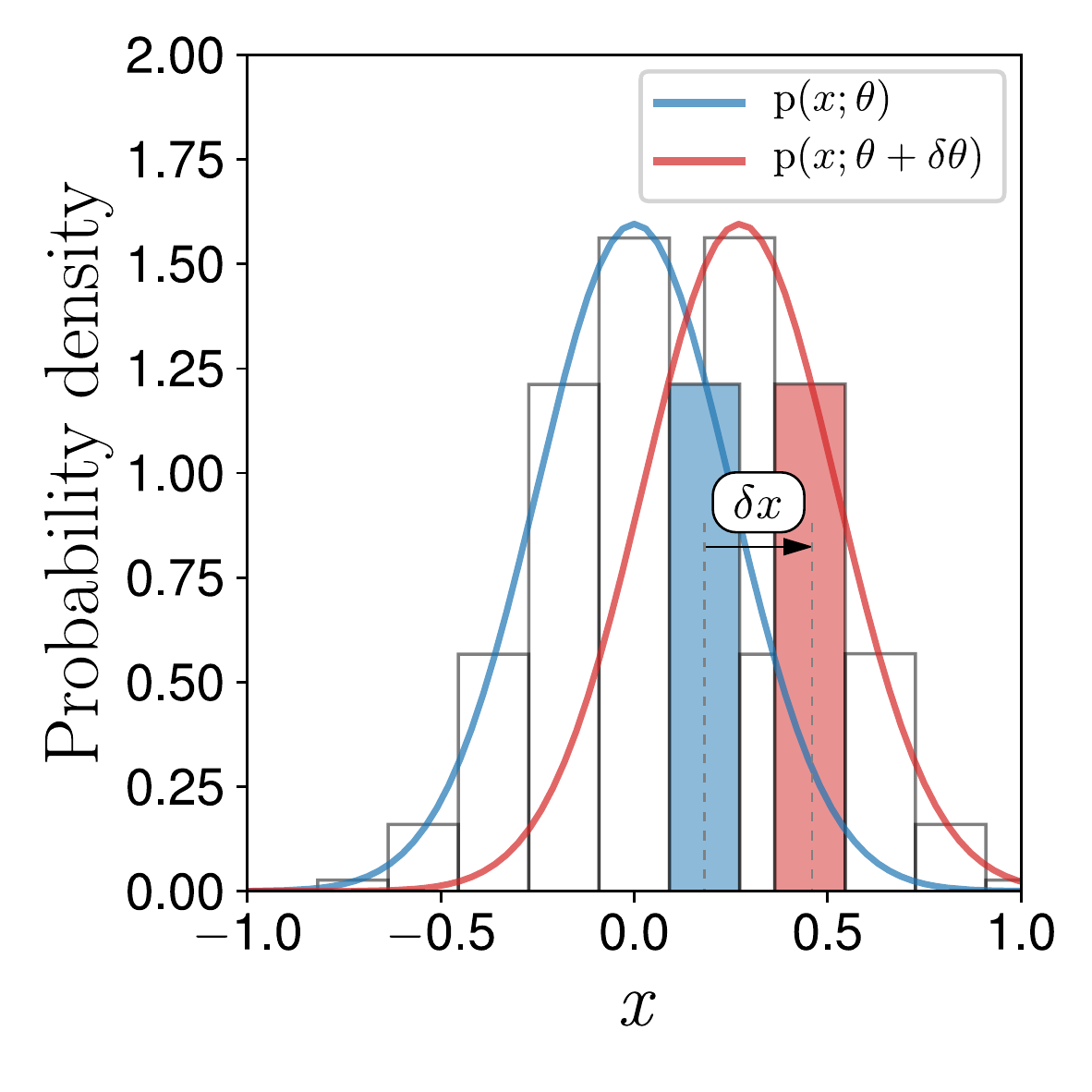}
		\caption{RP probability ``boxes''}
          \label{rpbox}
        \end{subfigure}
%%%%%%%%%%%%%%
	\begin{subfigure}{.49\textwidth}
		\includegraphics[width=\textwidth]{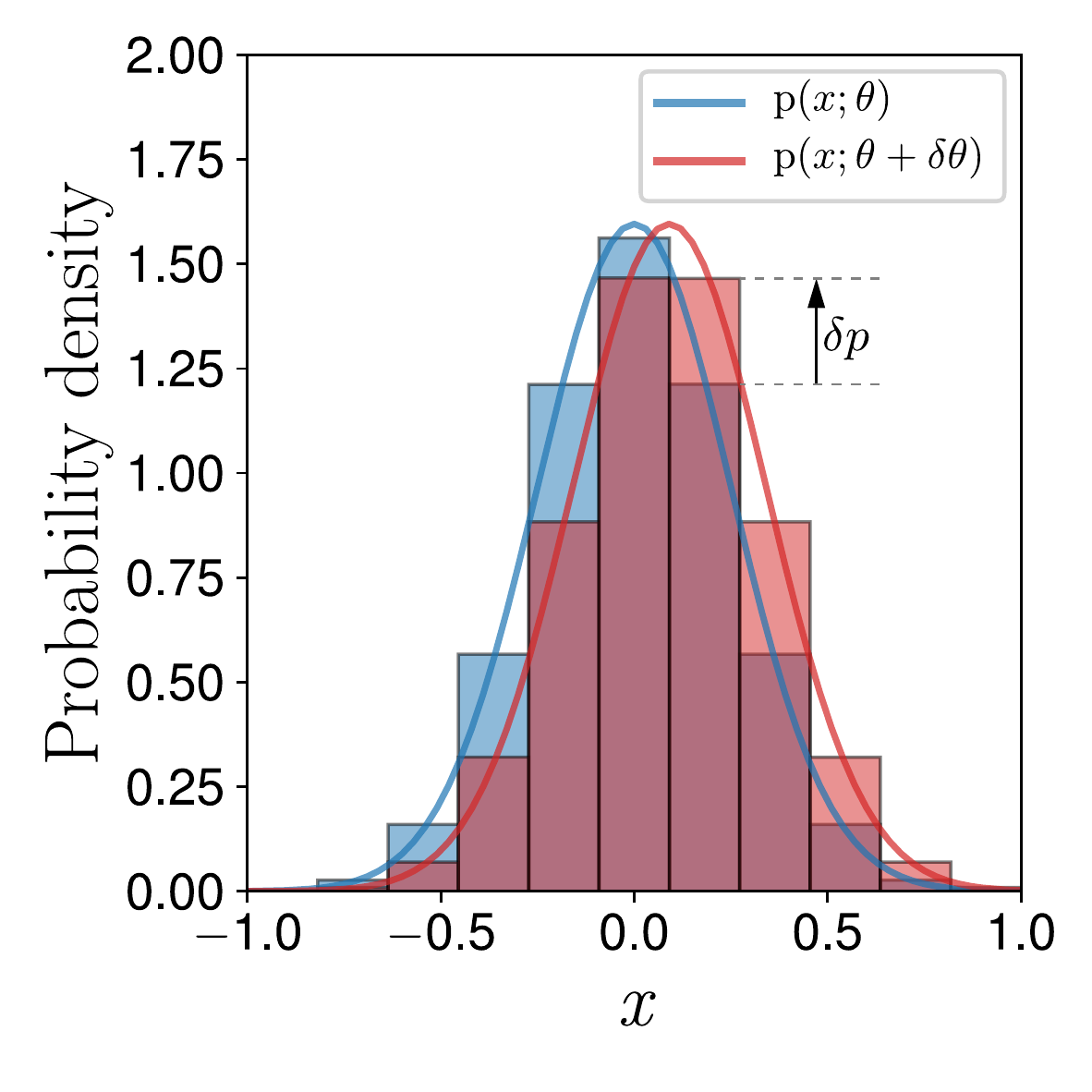}
		\caption{LR probability ``boxes''}
          \label{lrbox}
	\end{subfigure}
%%%%%%%%%%%%%%
	\caption{LR keeps the boundaries of the ``boxes" fixed, while
          RP keeps the probability mass fixed.}
          \label{probboxes}
\end{figure*}

Here we give the first of our two explanations of the link between LR
and RP gradients.  The explanation relies on a first principles
thinking about the effect that changing the parameters of a
probability distribution $\theta$ has on infinitesimal ``boxes" of
probability mass (Fig.~\ref{probboxes}). 
Both LR and RP are trying to estimate
$\deriv{}{\theta}\int \p{x;\theta}\phi(x)\textup{d}x$.  A typical
finite explanation of Riemann integrals is performed by discretizing
the integrand into ``boxes" of size $\Delta x$, and summing:
$\deriv{}{\theta}\sum_{i=1}^N \p{x_i;\theta}\Delta x_i\phi(x_i)$.
Taking the limit as $N \rightarrow \infty$ recovers the true integral.
In this equation $\p{x_i;\theta}\Delta x_i$ is the amount of
probability mass inside the ``box", and $\phi(x_i)$ is the function
value inside the ``box".

Such a view can be used to explain RP gradients.  In this case, the
boundaries of the ``box" are fixed with reference to the shape of the
probability distribution, i.e. for each $i$ we define the center of
the box as $x_i = g(\epsilon_i;\theta)$, and the boundaries as
$g(\epsilon_i\pm \Delta\epsilon/2;\theta)$, where $\epsilon_i$ is the
reference position on a fixed simple distribution $\p{\epsilon}$.
the amount of probability mass
assigned to each ``box" stays fixed at
$  \Delta p_i = \p{\epsilon}\Delta\epsilon~;
$ however, the center of the
``box" moves, so the function value $\phi(x_i)$ inside each ``box"
changes by
$\delta\phi_i = \phi\left(g(\epsilon_i;\theta+\delta\theta)\right) -
\phi\left(g(\epsilon_i;\theta)\right) = \phi(x_i+\delta x_i) -
\phi(x_i)$.
The full derivative can then be expressed as
$\deriv{}{\theta}\expectw{x\sim\p{x;\theta}}{\phi(x)} \approx 
\frac{1}{\delta\theta}\sum_{i=1}^N\Delta p_i\delta\phi_i =
\sum_{i=1}^N\Delta p_i\frac{\delta\phi_i}{\delta x_i}\frac{\delta
  x_i}{\delta\theta}$.
Taking the infinitesimal limit as
$N\rightarrow\infty$, and noting
$\Delta p_i = \p{x_i;\theta}\Delta x_i$, we obtain the RP gradient
estimator
$\int \p{x;\theta}\deriv{\phi(x)}{x}\deriv{x}{\theta}~\textup{d}x$.
We see that RP essentially estimates the
gradient by keeping the probability mass inside each ``box" fixed, but
estimating how the function value $\phi$ inside the ``box" changes as
the parameters $\theta$ are perturbed.

\remove{As the parameters of the probability
distribution are shifted by $\delta\theta$, the RP method keeps the
amount of probability mass inside each ``box" fixed, but the position
of the ``box" changes meaning that the function value $\phi(x_i)$
inside each ``box" changes. The gradient is estimated by estimating
this change $\deriv{\phi(x_i)}{x}\delta x_i$.%\footnote{The change is correctly
%estimated by the derivative in the infinitesimal limit.}
}

The LR gradient, on the other hand, keeps the boundaries of the
``boxes" fixed, i.e. the centre of the box is at $x_i$, and the
boundaries at $x_i \pm \Delta x_i/2$. Now, as the boundaries are
independent of $\theta$, the function value $\phi(x_i)$ inside the box
stays fixed, even as $\theta$ is perturbed by $\delta\theta$; however,
the probability mass inside the box changes, because the density
changes by
$\delta p_i = \p{x_i;\theta+\delta\theta} - \p{x_i;\theta}$.
The full
derivative can be expressed as
$\deriv{}{\theta}\expectw{x\sim\p{x;\theta}}{\phi(x)} \approx
  \frac{1}{\delta\theta}\sum_{i=1}^N\Delta x_i\delta p_i\phi(x_i) =
\sum_{i=1}^N \p{x_i;\theta}\Delta x_i\frac{\delta
  p_i/\delta\theta}{\p{x_i;\theta}}\phi(x_i)$.
Where we have multiplied and divided by $\p{x_i;\theta}$. Taking the
infinitesimal limit recovers the LR gradient
$\int\p{x;\theta}\frac{\deriv{\p{x;\theta}}{\theta}}
{\p{x;\theta}}\phi(x)~\textup{d}x =
\expectw{x\sim\p{x;\theta}}{\frac{\deriv{\p{x;\theta}}{\theta}}
  {\p{x;\theta}}\phi(x)}$. The transformation
$\p{x;\theta}\frac{\deriv{\p{x;\theta}}{\theta}} {\p{x;\theta}} =
\p{x;\theta}\deriv{\log\p{x;\theta}}{\theta}$ is known as the
log-derivative trick, and it may appear to be the essence behind the
LR gradient, but actually the multiplication and division by
$p(x;\theta)$ is just a special case of the more general Monte Carlo
integration principle. Any integral $\int f(x)~\textup{d}x$ can be
approximated by sampling from a distribution $q(x)$ as
$\int f(x)~\textup{d}x = \int q(x) \frac{f(x)}{q(x)}~\textup{d}x =
\expectw{x\sim q(x)}{\frac{f(x)}{q(x)}}$. Rather than thinking of the
LR gradient in terms of the log-derivative term, it may be better
to think of it as simply estimating the integral
$\int\deriv{\p{x;\theta}}{\theta}\phi(x)~\textup{d}x$ by applying the
appropriate importance weights to samples from $p(x;\theta)$. Thus, we
see that in the discretized case, the LR gradient picks
$q(x) = \p{x;\theta}$ \citep{jie2010connection} and performs Monte
Carlo integration to approximate
$\frac{1}{\delta\theta}\sum_{i=1}^N\Delta x_i\delta p_i\phi(x_i)$ by
sampling from $P(x_i) = \Delta x_i\p{x_i;\theta}$. To summarize: LR
estimates the gradient by keeping the boundaries of the boxes fixed,
measuring the change in probability mass in each box, and weighting by
the function value: $\phi(x_i)\delta p$.

Sometimes, the LR gradient is described as being ``kind of like a
finite difference gradient"
\citep{salimans2017oaies,mania2018simplers}, but here we see that it is
a different concept, which does not rely on fitting a straight line
between differences of $\phi$ (App.~\ref{lrbasicsapp}), but estimates
how probability mass is reallocated among different $\phi$ values via
Monte Carlo integration by sampling from $\p{x;\theta}$.

\remove{The LR gradient, on the other hand, keeps the boundaries of the
``boxes" fixed, but measures the change in probability mass in
each box, $\phi(x_i)\delta p$.  Sometimes, the LR gradient is
described as being ``kind of like a finite difference gradient"
\citep{salimans2017oaies,mania2018simplers}, but here we see that it is
a different concept, which does not rely on fitting a straight line
between differences of $\phi$ (App.~\ref{lrbasicsapp}), but
estimates how probability mass is reallocated among different
$\phi$ values via Monte Carlo integration by sampling from $\p{x;\theta}$.}

\section{A unified probability flow view of
  LR and RP gradients}
\label{flowtheor}

\begin{figure*}[!t]
        \centering
	\begin{subfigure}{.49\textwidth}
		\includegraphics[width=\textwidth]{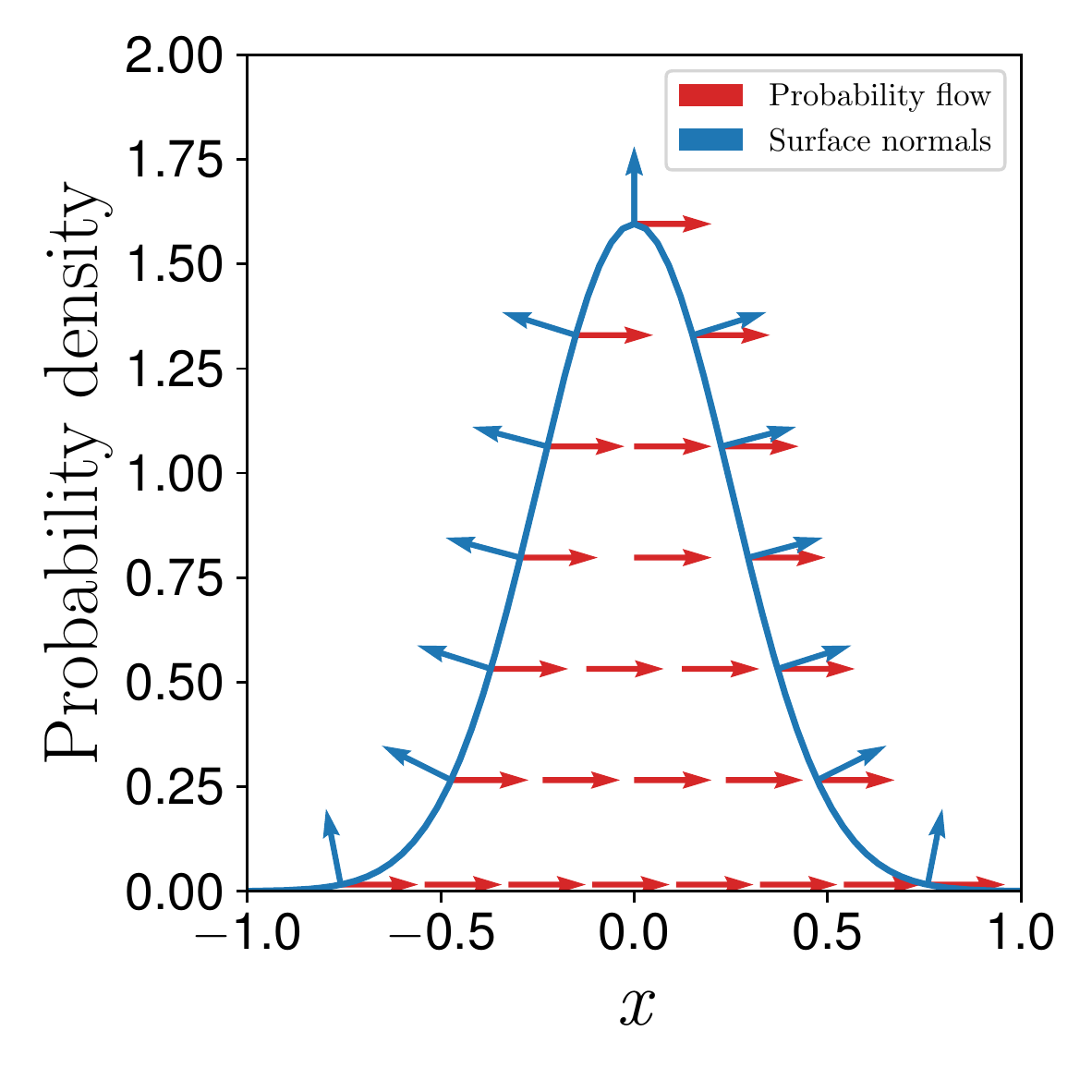}
		\caption{$\mu$ flow lines}
          \label{mulines}
        \end{subfigure}
%%%%%%%%%%%%%%
	\begin{subfigure}{.49\textwidth}
		\includegraphics[width=\textwidth]{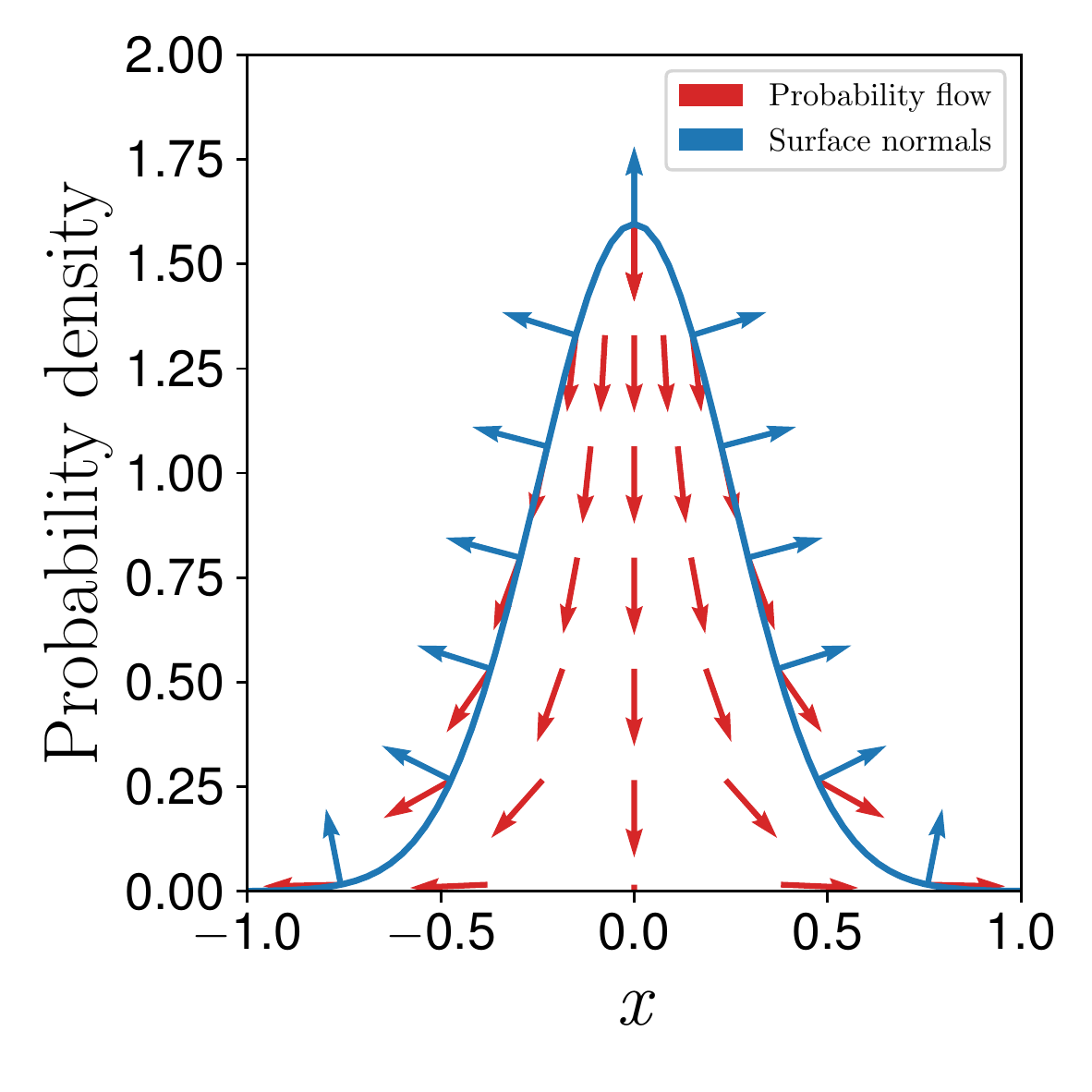}
		\caption{$\sigma$ flow lines}
          \label{sigmalines}
	\end{subfigure}
%%%%%%%%%%%%%%
	\caption{Probability flow lines when $\mu$ and $\sigma$ are perturbed.}
          \label{problines}
\end{figure*}

Here we give another explanation of LR and RP.
The appeal of this theory is that both LR and RP come out of the
same derivation, thus showing a link between the two. In particular,
we define a virtual incompressible flow of probability mass imposed by
perturbing the parameters $\theta$ of $\p{x;\theta}$, which can
be used to express the derivative of the expectation as an integral
over this flow.
LR and RP estimators correspond to duals of this
integral under the divergence theorem (App.~\ref{probflowderivapp}).

The main idea resembles RP, but in addition to sampling the
$\bfv{x}$ location, we sample a height $h$ for each point:
$h = \epsilon_h\p{\bfv{x};\theta}$, where
$\epsilon_h \sim \textup{unif}(0,1)$, i.e., the sampling space is
extended with an additional dimension for the height
$\tilde{\bfv{x}} = [\bfv{x}^T, h]^T$, and we are uniformly sampling in
the volume under $\p{\bfv{x};\theta}$. The
definition of $g$ in the introduction is extended, s.t.
$\tilde{g}(\epsilon_x,\epsilon_h) = \tilde{\bfv{x}}$. The expectation
turns into:

\begin{equation}
\begin{aligned}
  &\deriv{}{\theta}\int\p{\bfv{x};\theta}\phi(\bfv{x})\textup{d}\bfv{x}
   \\
  &= \deriv{}{\theta}\int_{\epsilon_x}\int_{\epsilon_h}
  \p{\epsilon_x}\p{\epsilon_h}\phi\left(\tilde{g}(\epsilon_x,\epsilon_h)\right)
    \textup{d}\epsilon_x\textup{d}\epsilon_h\\
  &=
  \int_{V}\nabla_{\tilde{\bfv{x}}}\phi(
  \tilde{\bfv{x}})\nabla_\theta \tilde{g}(\epsilon_x,\epsilon_h)
    \textup{d}V.
  \end{aligned}
\label{volint}
\end{equation}

In Eq.~(\ref{volint}), $V$ is the volume under the curve, and
$\phi([\bfv{x}^T, h]^T) \coloneqq \phi(\bfv{x})$ ignores the
$h$-component. Each column $i$ of
$\nabla_\theta \tilde{g}(\epsilon_x,\epsilon_h)$ corresponds to a vector field
induced by perturbing the $i^{\textup{th}}$ component of $\theta$. The
red lines in Fig.~\ref{problines} show the induced flow fields for a
Gaussian distribution as the mean and variance are perturbed.  The
other member of the integral,
$\nabla_{\tilde{\bfv{x}}}\phi(\tilde{\bfv{x}})$ is the grad of the
scalar field $\phi(\tilde{\bfv{x}})$. As $\phi$ does not depend on
$h$, the grad will always be parallel to the $\bfv{x}$ axes with
magnitude $\deriv{\phi}{\bfv{x}}$.

According to the divergence theorem, the \emph{volume integral} in
Eq.~(\ref{volint}) can be turned into a \emph{surface integral} over the
boundary $\bfv{S}$ ($\textup{d}\bfv{S}$ is a
shorthand for $\hat{\bfv{n}}\textup{d}S$, where $\hat{\bfv{n}}$ is the surface normal
vector):

\begin{equation}
\label{divergenceeq}
\int_V \nabla\cdot\bfv{F}\textup{d}V = \int_S\bfv{F}\textup{d}\bfv{S}.
\end{equation}

In Eq.~(\ref{divergenceeq}), $\bfv{F}$ is any vector field. A common corollary
arises by picking $\bfv{F} = \phi\bfv{v}$, where $\phi$ is a scalar
field, and $\bfv{v}$ is a vector field. We choose
$\bfv{v} = \nabla_\theta \tilde{g}(\epsilon_x,\epsilon_h)\delta\theta$, where
$\delta\theta$ is an arbitrary perturbation in $\theta$, so that
$\bfv{F} = \phi(\tilde{\bfv{x}})\nabla_\theta
\tilde{g}(\epsilon_x,\epsilon_h)\delta\theta$, in which case
$\nabla\cdot\bfv{F} =
\nabla\cdot\left(\phi(\tilde{\bfv{x}})\nabla_\theta
  \tilde{g}(\epsilon_x,\epsilon_h)\delta\theta\right) =
\nabla_{\tilde{\bfv{x}}}\phi(\tilde{\bfv{x}})\nabla_\theta
\tilde{g}(\epsilon_x,\epsilon_h)\delta\theta +
\phi(\tilde{\bfv{x}})\nabla_{\tilde{\bfv{x}}}\cdot\nabla_\theta
\tilde{g}(\epsilon_x,\epsilon_h)\delta\theta$. Note that the term
$\nabla_\theta \tilde{g}(\epsilon_x,\epsilon_h)\delta\theta$ corresponds to an
incompressible flow (because the probability density does not change
at any point in the augmented space). As the div of an incompressible
flow is 0, then
$\nabla_{\tilde{\bfv{x}}}\cdot\nabla_\theta
\tilde{g}(\epsilon_x,\epsilon_h)\delta\theta = 0$, and the second term
disappears. Noting that $\delta\theta$ can be canceled, because it is
arbitrary, we are left with the equation:

\begin{equation}
\begin{aligned}
  \int_{V}\nabla_{\tilde{\bfv{x}}}\phi(
  \tilde{\bfv{x}})\nabla_\theta \tilde{g}(\epsilon_x,\epsilon_h)
    \textup{d}V &= \int_{S} \phi(
  \tilde{\bfv{x}})\nabla_\theta \tilde{g}(\epsilon_x,\epsilon_h)\textup{d}\bfv{S}.
  \end{aligned}
\label{surfint}
\end{equation}

Now we explain how the left-hand side of Eq.~(\ref{surfint}) gives rise
to the RP gradient estimator, while the right-hand side corresponds to
the LR gradient estimator. \remove{The RP estimator follows quite easily, while the
LR estimator is slightly more involved.}

\paragraph{RP estimator:}
Consider the
$\nabla_{\tilde{\bfv{x}}}\phi( \tilde{\bfv{x}})\nabla_\theta
\tilde{g}(\epsilon_x,\epsilon_h)$ term. As the scalar field
$\phi( \tilde{\bfv{x}})$ is independent of the height location $h$,
the component of the grad in that direction is 0, and
$\phi( \tilde{\bfv{x}}) = \phi(\bfv{x})$.  As the $h$-component is 0,
then the value of $\tilde{g}$ in the $h$-direction is multiplied by 0, and is
irrelevant for the product, so
$\nabla_{\tilde{\bfv{x}}}\phi( \tilde{\bfv{x}})\nabla_\theta
\tilde{g}(\epsilon_x,\epsilon_h) = \nabla_{\bfv{x}}\phi(\bfv{x}) \nabla_\theta
g(\epsilon_x)$, which is just the term used in the RP
estimator. Hence, the left-hand side of Eq.~(\ref{surfint}) corresponds
to the RP gradient.

\paragraph{LR estimator:} We will show that the LR estimator tries to
integrate
$\int_S\phi( \tilde{\bfv{x}})\nabla_\theta
\tilde{g}(\epsilon_x,\epsilon_h)\textup{d}\bfv{S}$. To do so, note that
$\textup{d}\bfv{S} = \hat{\bfv{n}}\textup{d}S$. It is necessary to
express the normalized surface vector $\hat{\bfv{n}}$, and then perform
the integral over the surface. The derivation is in App.~\ref{surfDeriv},
and the final result is:

\begin{equation}
\begin{aligned}
  \int_{S} \phi(\tilde{\bfv{x}})\nabla_\theta
  \tilde{g}(\epsilon_x,\epsilon_h)~\textup{d}\bfv{S} =
  \int_{X}
  \phi(\bfv{x})\deriv{\p{\bfv{x};\theta}}{\theta}
  ~\textup{d}\bfv{x}.
\label{eq:subs}
\end{aligned}
\end{equation}

We have already seen that a Monte Carlo integration of
the right-hand side of Eq.~(\ref{eq:subs}) using samples from
$\p{\bfv{x};\theta}$ yields the LR gradient
estimator. Thus, the RP and LR
are duals under the divergence theorem. To further strengthen this
claim we prove that the LR gradient estimator is the unique estimator
that takes weighted averages of the function values $\phi(\bfv{x})$.

\newtheorem{lrunique}{Theorem}
\begin{lrunique}[Uniqueness of LR estimator]
$\psi(\bfv{x}) = \p{\bfv{x};\theta}\deriv{\log \p{\bfv{x};\theta}}{\theta}$ is
  the unique function $\psi(\bfv{x})$, s.t.
  $\int \psi(\bfv{x})\phi(\bfv{x})\textup{d}\bfv{x} =
  \deriv{}{\theta}\int\p{\bfv{x};\theta}\phi(\bfv{x})\textup{d}\bfv{x}$
  for any $\phi(\bfv{x})$.

  Proof. \textup{Suppose that there exist $\psi(\bfv{x})$ and $f(\bfv{x})$, s.t.
  $\int \phi(\bfv{x})\psi(\bfv{x})~\textup{d}\bfv{x} = \int
  \phi(\bfv{x})f(\bfv{x})~\textup{d}\bfv{x}$ for any $\phi(\bfv{x})$.
  Rearrange the equation into
  $\int
  \phi(\bfv{x})\left(\psi(\bfv{x})-f(\bfv{x})\right)~\textup{d}\bfv{x} =
  0$, then pick $\phi(\bfv{x}) = \psi(\bfv{x})-f(\bfv{x})$ from which we
  get
  $\int \left(\psi(\bfv{x})-f(\bfv{x})\right)^2~\textup{d}\bfv{x} = 0$.
  Therefore, $\psi = f$. Q.E.D.}
  \label{lrunique}
\end{lrunique}

We see that Eq.~(\ref{eq:subs}) was immediately clear
without having to go through the derivation. \remove{(given that the
  RP gradient estimator is unbiased).} The same analysis does not work
for RP (App.~\ref{rpNotUnique}). Indeed, there are infinitely many
RP gradients \citep{jankowiak2018rpflow}. Moreover, the analysis does not
consider coupled sampling of $\bfv{x}$ \citep{walder2019newtricks}.

\section{Slice ratio importance sampling}
\label{sliceratio}

\begin{figure*}[!t]
        \centering
	\begin{subfigure}{.24\textwidth}
		\includegraphics[width=\textwidth]{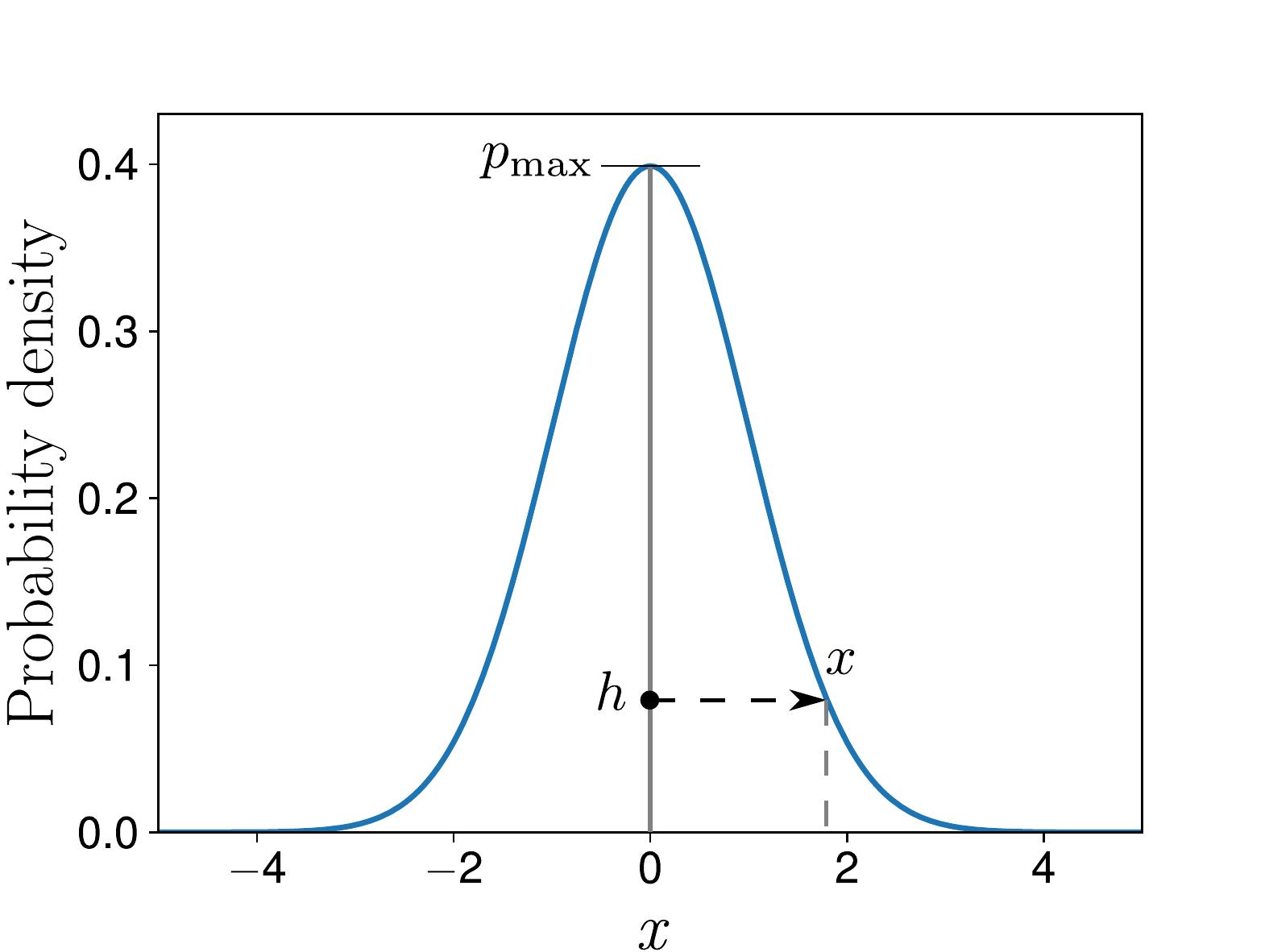}
		\caption{Slice ratio sampling\remove{Gaussian as reference}}
          \label{gdist}
        \end{subfigure}
%%%%%%%%%%%%%%
	\begin{subfigure}{.24\textwidth}
		\includegraphics[width=\textwidth]{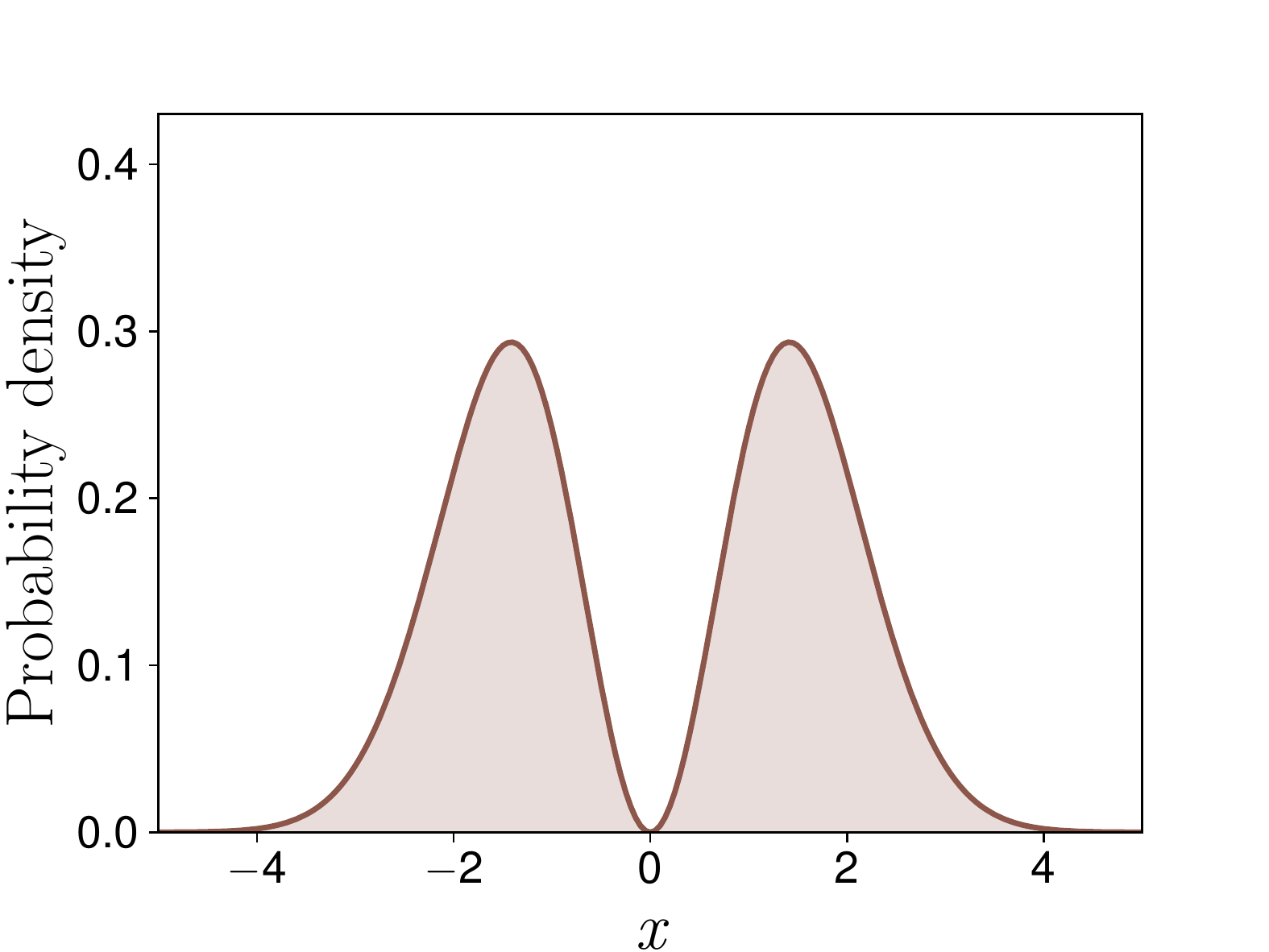}
		\caption{L-distribution}
          \label{ldist}
	\end{subfigure}
%%%%%%%%%%%%%%
	\begin{subfigure}{.24\textwidth}
		\includegraphics[width=\textwidth]{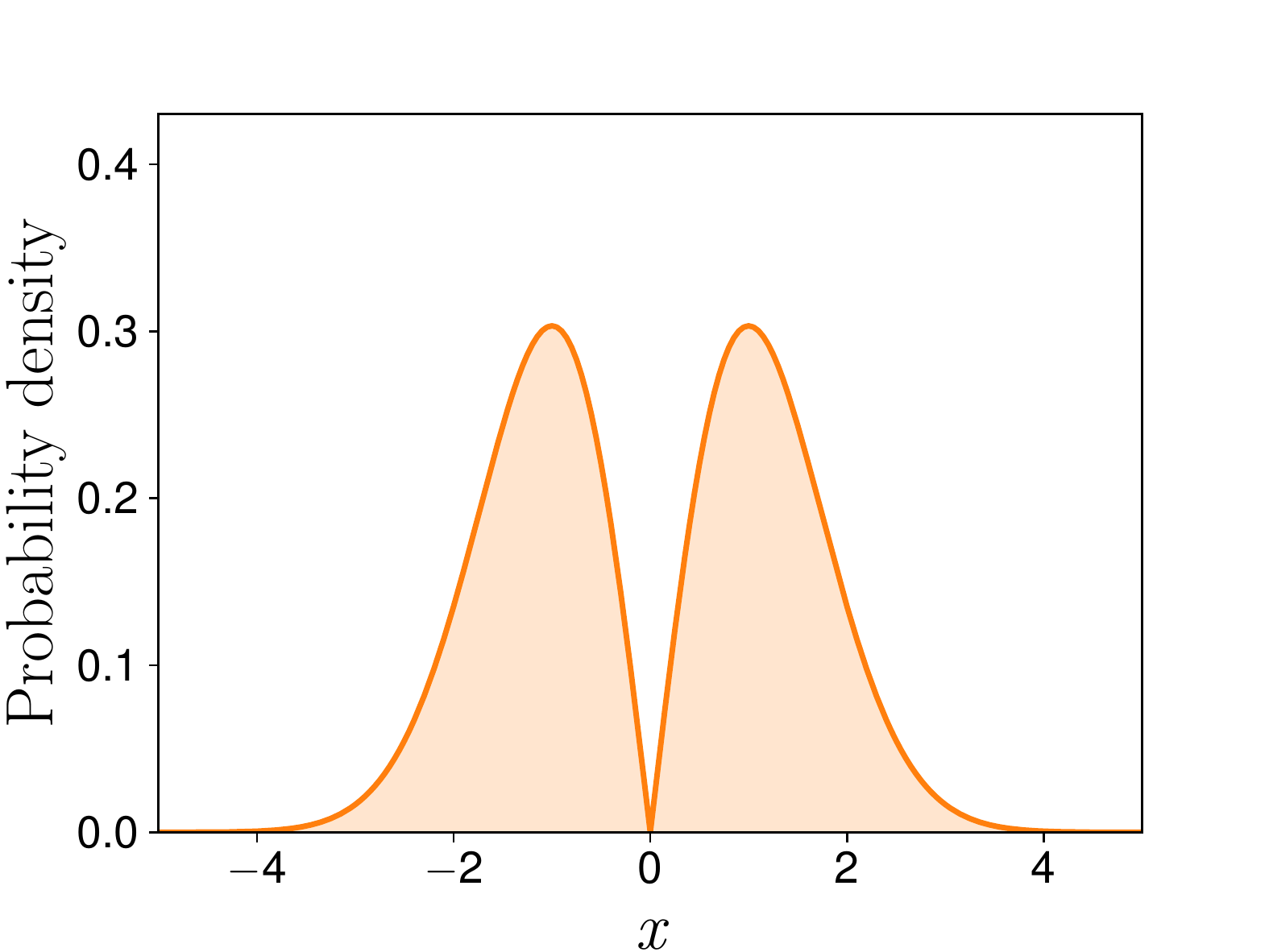}
		\caption{B-distribution}
          \label{bdist}
        \end{subfigure}
        \begin{subfigure}{.24\textwidth}
		\includegraphics[width=\textwidth]{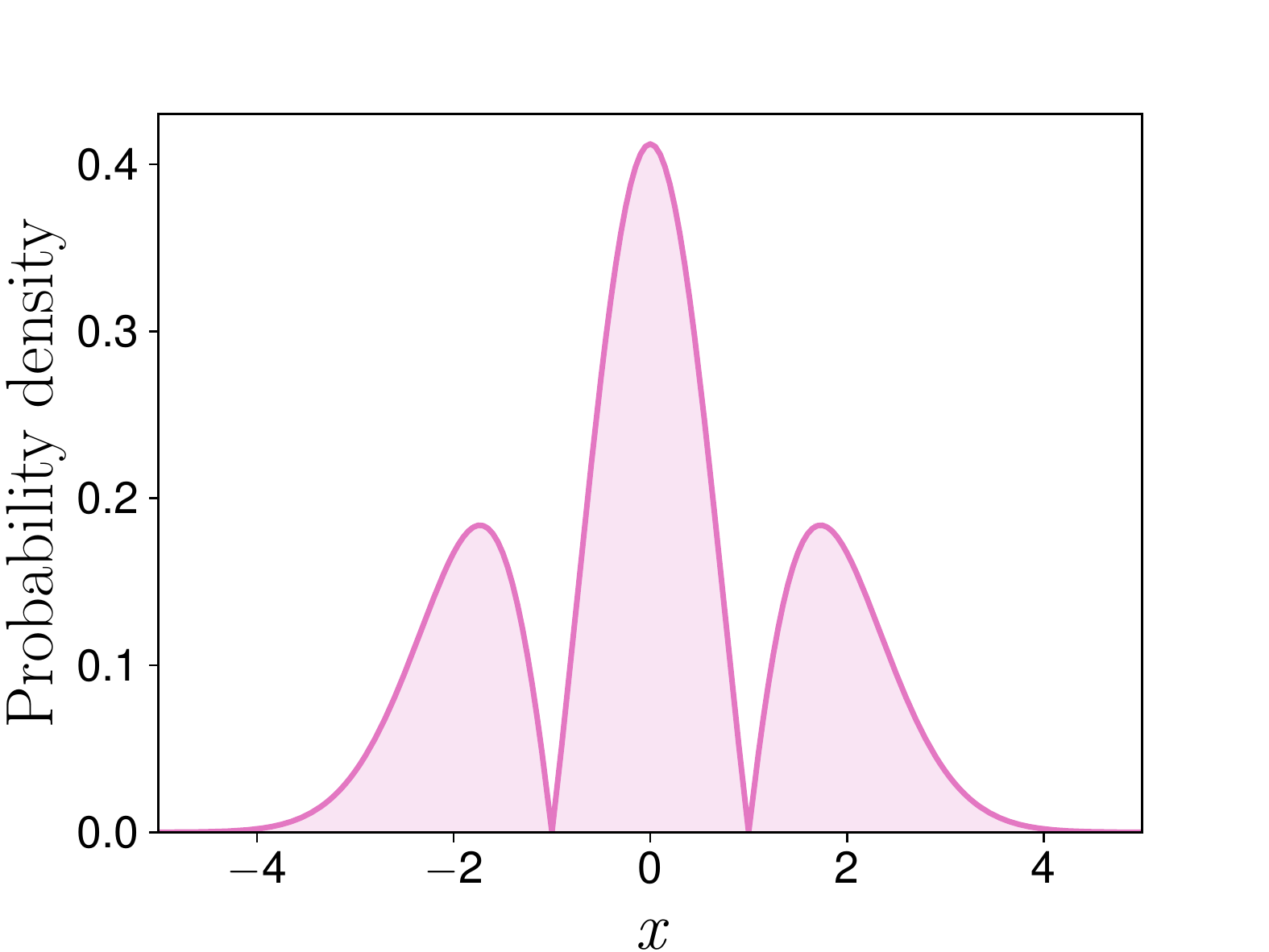}
		\caption{W-distribution}
                \label{wdist}
	\end{subfigure}\\
	\caption{Slice ratio sampling method Fig.~\ref{gdist}, and
          new importance sampling distributions to reduce LR gradient
          variance (Figs.~\ref{ldist},\ref{bdist},\ref{wdist}). The
          shaded regions are histograms generated with the direct
          sampling methods, and the solid lines are the analytic
          densities. Both methods match, demonstrating the correctness
          of the derivations. B and W distributions are
            optimal for $\deriv{}{\mu}$ and $\deriv{}{\sigma}$
            respectively.}
          \label{newdistr}
\end{figure*}

As LR is the only unbiased gradient estimator that weights samples of
$\phi(x)$ (Sec.~\ref{flowtheor}), what could be done to reduce its
variance? One underexplored option is to keep the product
$\p{\bfv{x};\theta}\deriv{\log \p{\bfv{x};\theta}}{\theta}$ the same,
but sample from a different distribution $q(x)$ using importance
sampling
$q(x)\p{\bfv{x};\theta}\deriv{\log
  \p{\bfv{x};\theta}}{\theta}\nicefrac{1}{q(x)}$.  How to pick $q(x)$?
Our first attempt (App.~\ref{sliceintegral}, Fig.~\ref{ldist}) was
suboptimal.

\remove{What more can be done?}

\remove{The derivation in Sec.~\ref{sliceintegral} appeared quite ad-hoc, and it
is unclear whether it is a good distribution to sample from. In this section
we derive an optimal sampling distribution to minimize the variance. The
concept of sampling the height introduced in the previous section will
prove useful in this derivation, and in fact the L-distribution turns out
to be an optimal distribution for a 2-dimensional Gaussian base distribution.}

\paragraph{Optimal importance sampling for minimum gradient variance:}
We seek a distribution $q(x)$, s.t. the variance of
$\deriv{\p{x;\theta}}{\theta}\phi(x)/q(x)$ is minimized. The
derivation is analogous to the standard result for optimal importance
sampling in statistics \citep{mcbook}. As $\phi(x)$ is not known {\it
  a priori}\remove{Note that other methods, such as adaptive
  importance sampling could be used to also take into account for
  $\phi(x)$ when sampling multiple points, but here we focus on the
  non-adaptive case. Of course adaptive methods could be combined with
  our methods to achieve even better performance.}, we minimize the
variance of $\deriv{\p{x;\theta}}{\theta}/q(x)$. The omission is
well-justified in the multidimensional setting, as most of the
variation in $\phi(\bfv{x})$ is caused by the other dimensions and can
thus be viewed as noise. See also App.~\ref{sec:phiom} for several
other justifications.\remove{This omission is well-justified in the
  multidimensional setting, where most of the variation of $\phi$ is
  caused by the variation in the other dimensions, and can thus be
  viewed as noise.}  The variance can be expressed as
$\int q(x)
\left(\frac{\deriv{\p{x;\theta}}{\theta}}{q(x)}\right)^2~\textup{d}x$.
Adding in the constraint $\int q(x)~\textup{d}x = 1$ with a Lagrange
multiplier $\lambda$, and performing a variational optimization by
setting the derivative w.r.t. $q$ to 0 we have:

\begin{equation}
  -\left(\frac{\deriv{\p{x;\theta}}{\theta}}{q(x)}\right)^2
  + \lambda = 0 ~~\Rightarrow~~ q(x) =
  \left|\deriv{\p{x;\theta}}{\theta}\right|/\sqrt{\lambda}~.
  \label{eq:optdist}
\end{equation}

Eq.~(\ref{eq:optdist}) tells that the optimal importance sampling
distribution is proportional to the magnitude of the gradient of the
base distribution. How to normalize this
distribution, and how to sample from it?

For a Gaussian distribution, we can derive two possible distributions:
one for $\mu$ (Fig.~\ref{bdist}) and one for $\sigma$
(Fig.~\ref{wdist}). The derivative w.r.t $\mu$ appears more important,
so we derive it first. Note that
$\deriv{\p{x}}{\mu} = -\deriv{\p{x}}{x}$, and that by sampling a
height $h$ and transforming from the $h$-coordinate to the
$x$-coordinate via $x = p^{-1}(h)$, the probability density is
weighted: $\textup{d}h = \left|\deriv{\p{x}}{x}\right|\textup{d}x$
(App.~\ref{sliceintegral}). This insight allows us to derive the
distribution and a sampling method (App.~\ref{slrgderiv}). Namely, to
sample from the distribution: 1) sample
$h \sim \textup{unif}(0, p_{\mathrm{max}})$, where
$p_{\mathrm{max}} = \p{\mu;\mu,\sigma}$ is the peak probability
density, 2) compute the location of the edge of the slice
$x = p^{-1}(h)$ (Fig.~\ref{gdist}). Putting these results
together, one obtains the pdf, a sampling method and the LR gradient
estimator:

\begin{equation}
\begin{aligned}
\pind{B}{x;\mu,\sigma} = \frac{|x-\mu|}{2\sigma^2}
\exp\left(\frac{-(x-\mu)^2}{2\sigma^2}\right),&\\
x = \mu \pm \sigma\sqrt{-2\log(\epsilon_h)} ~~~\textup{where}~~~
\epsilon_h \sim \textup{unif}(0,1),&\\
\deriv{}{\mu}\expectw{x\sim\p{x}}{\phi(x)} =
\expectw{x\sim q(x)}{\textup{sgn}(x-\mu)\frac{2}{\sigma\sqrt{2\pi}}\phi(x)}.&
\end{aligned}
\label{eq:bdist}
\end{equation}

We call the derived distribution the B-distribution
(Fig.~\ref{bdist}), and the resulting gradient estimator the slice
ratio gradient (SLRG). Notice that the B-distribution is the
Rayleigh distribution symmetrized about the origin.  The derivation
for $\sigma$ is similar (App.~\ref{wdistDeriv}), but note,
$\deriv{\p{x}}{\sigma} = \sigma\deriv{^2\p{x}}{x^2}$ for a Gaussian.

\remove{Similarly
  to how the L-distribution was related to the Maxwell-Boltzmann
  distribution, the B-distribution is the Rayleigh distribution
  symmetrized about the origin.}

\paragraph{Slice ratio sampling for the symmetric Beta distribution:}
The slice ratio sampling method is crucial in some situations. For
example, consider a distribution, such as the symmetric Beta distribution:

\begin{equation}
  \pind{\beta}{x} =
  \frac{x^{\alpha-1}(1-x)^{\alpha-1}}{B(\alpha,\alpha)}.
\end{equation}

When $\alpha$ tends to 1 from above, this distribution tends to the
uniform distribution between 0 and 1. Consider a distribution with the
same shape, but where the mean is shifted, s.t. it
is symmetric about a parameter $\mu$, instead of
$x=\nicefrac{1}{2}$. In this case, as $\alpha$ tends to 1, the
variance of the gradient w.r.t. $\mu$ will tend to $\infty$, because
$\deriv{\pind{\beta}{x}}{\mu}$ is around 0 in most of the sampling
range, but very large at the edges of the distribution. We derived the optimal
pdf, sampling method and gradient estimator
(App.~\ref{slicebetaapp}):

\begin{equation}
\begin{aligned}
  &\pind{\beta R}{x} =
  \frac{\alpha-1}{2\times 0.25^{\alpha-1}}
  (x-x^2)^{\alpha-2}|1-2x|,\\
  &x =
  0.5 \pm 0.5\sqrt{1-
    \epsilon_h^{\nicefrac{1}{(\alpha-1)}}} ~~~\textup{where}~~~
\epsilon_h \sim \textup{unif}(0,1),\\
&\deriv{}{\mu}\expectw{x\sim\pind{\beta}{x}}{\phi(x)} \\
&=
\expectw{x\sim q(x)}{\textup{sgn}(x-0.5)
  \frac{2\times 0.25^{\alpha-1}}{B(\alpha,\alpha)}\phi(x)},
   \textup{for } \alpha>1.
\end{aligned}
\label{eq:betadist}
\end{equation}

For a shifted, stretched and centered distribution, replace $x$ with
$k(x-0.5)+\mu$, and the gradient estimator needs to be scaled down by
$k$. To obtain a variance $\sigma^2$, set $k$ to
$2\sigma\sqrt{2\alpha+1}$.

\paragraph{Multidimensional case:} For a Gaussian $\p{x;\theta}$, as the dimension
increases, the optimal $q(x)$ tends to the original distribution
(App.~\ref{multdim}). For this reason, we propose to sample each dimension
separately from the B-distribution, potentially allowing for a bias,
but while reducing the variance of the gradient estimator
(see also App.~\ref{sec:multjust} for more justification). In general,
we believe that such a technique will be necessary for other distributions
as well if the dimension grows high. To see this, consider the importance
weighted likelihood ratio gradient estimator for
a factorized distribution $\p{\bfv{x};\theta} = \prod_i\pind{i}{x_i;\theta_i}$:
\begin{equation}
  \begin{aligned}
    &\deriv{}{\theta_i}\expectw{\bfv{x}\sim\p{\bfv{x}}}{\phi(\bfv{x})}\\
    &= \expectw{\bfv{x}\sim q(\bfv{x})}{\frac{\p{\bfv{x}}}
      {q(\bfv{x})}\deriv{\log p}{\theta_i}\phi(\bfv{x})} \\ &=
    \expectw{\bfv{x}\sim q(\bfv{x})}{\frac{\pind{\backslash
          i}{\bfv{x}_{\backslash i}}\pind{i}{x_i}} {q_{\backslash
          i}(\bfv{x}_{\backslash i})q_i(x_i)}\deriv{\log
        \pind{i}{x_i}}{\theta_i}\phi(\bfv{x})}, \\ &~~~~~~~~~~~~~~\textup{where }p_{\backslash i}
    \textup{ is } \prod_{j\neq i}\pind{j}{x_j;\theta_j}.
  \label{multdimeq}
  \end{aligned}
\end{equation}

While $q$ can be modified to reduce the variance of
$\frac{\pind{i}{x_i}}{q_i(x_i)}\deriv{\log \pind{i}{x_i}}{\theta}$,
this will increase the variance of the
$\frac{\pind{\backslash j}{\bfv{x}_{\backslash j}}}{q_{\backslash
    j}(\bfv{x}_{\backslash j})}$ terms for $j\neq i$. If each $q_j$ is
modified, then the variance of these terms grows exponentially with
the dimension, and any decrease in variance from having modified $q_i$
becomes negligible. Our proposed solution is
to replace
$\frac{\pind{\backslash j}{\bfv{x}_{\backslash j}}} {q_{\backslash
    j}(\bfv{x}_{\backslash j})}$ with its expected value, which is
1. Note that our technique is not just a convenience, but it is
a necessity.
In practice, such a scheme may introduce a small bias, but
drastically reduce the variance. Next we show some fairly general
conditions under which this method still gives an unbiased gradient
estimator.

\paragraph{Sufficient conditions for an unbiased gradient 
  estimator in high dimensions with our scheme:}

\begin{enumerate}
\item If $\phi(\bfv{x}) = \sum_{i=1}^D\phi_i(x_i)$, then our estimation scheme is unbiased.
\item If $\phi(\bfv{x})$ is quadratic, then our estimation is scheme is unbiased.
\end{enumerate}
Both conditions are independently sufficient for unbiasedness (derivations in App.~\ref{suffcon}).

\paragraph{Effect of greater variance of $q_i$:} Lastly, we point
toward another issue with modifying $q_i$ in Eq.~(\ref{multdimeq}).  The
variance of $q_i$ may be larger than the variance of $p_i$, and this
could manifest as a larger variance of $\phi(\bfv{x})$, which would
act as additional noise on the other dimensions $j\neq i$. Our
proposed solution is to optimize the reduction in gradient variance
while constraining the variance of $q$. Assuming the mean $\mu=0$,
this can be performed using a variational optimization with an
additional Lagrange multiplier for
$\int q(x)x^2\textup{d}x = k\sigma^2$ analogously to
Eq.~(\ref{eq:optdist}). The general equation is

\begin{equation}
\label{varconstoptimality}
q(x) = \left|\deriv{\p{x;\theta}}{\theta}\right|/\sqrt{\lambda_1 + \lambda_2x^2}.
\end{equation}

\begin{figure*}
        \centering
	\begin{subfigure}{.32\textwidth}
		\includegraphics[width=\textwidth]{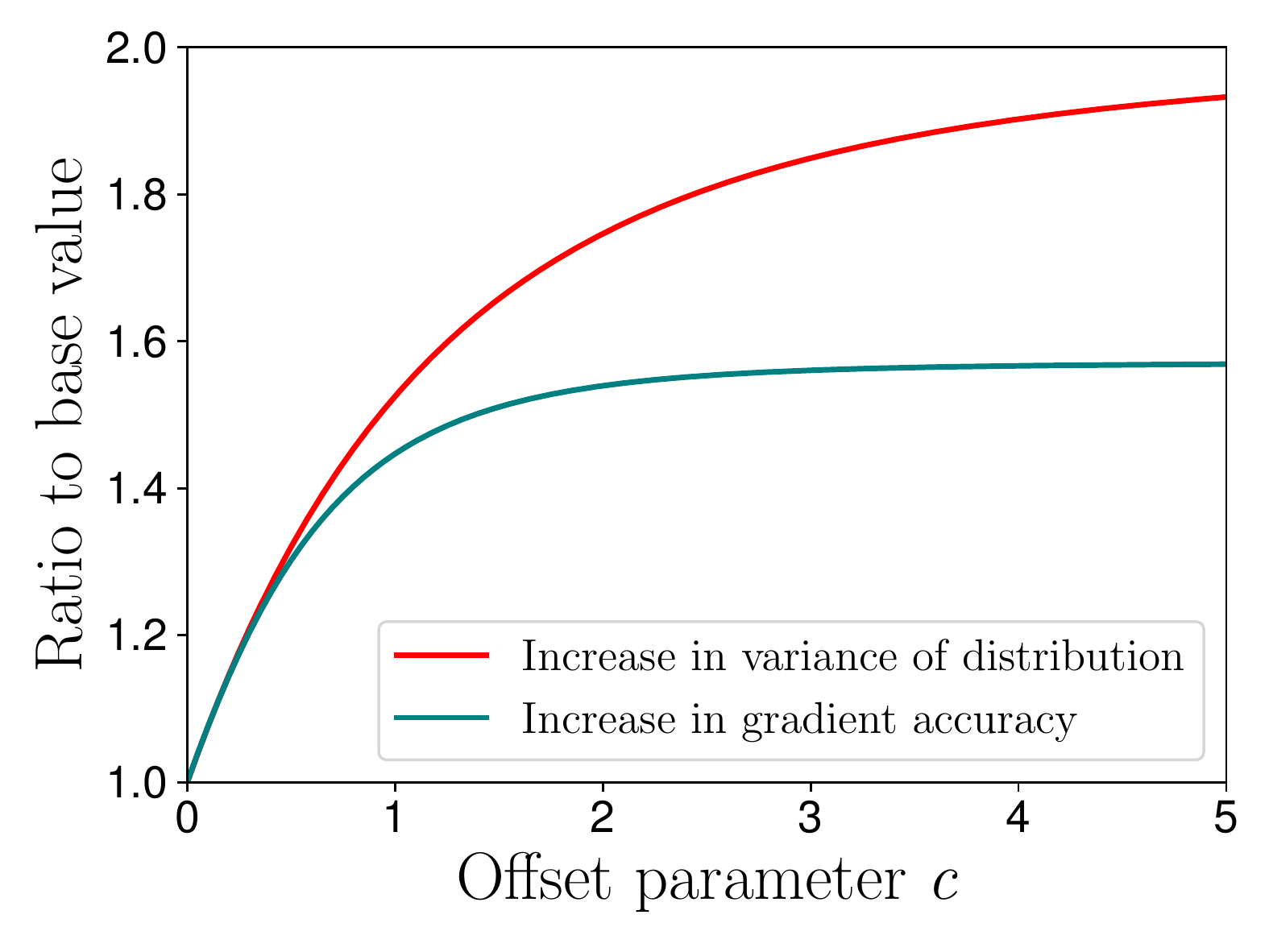}
		\caption{$t(c)$ in teal and $v(c)$ in red}
          \label{varscaleplot}
        \end{subfigure}
%%%%%%%%%%%%%%
	\begin{subfigure}{.32\textwidth}
		\includegraphics[width=\textwidth]{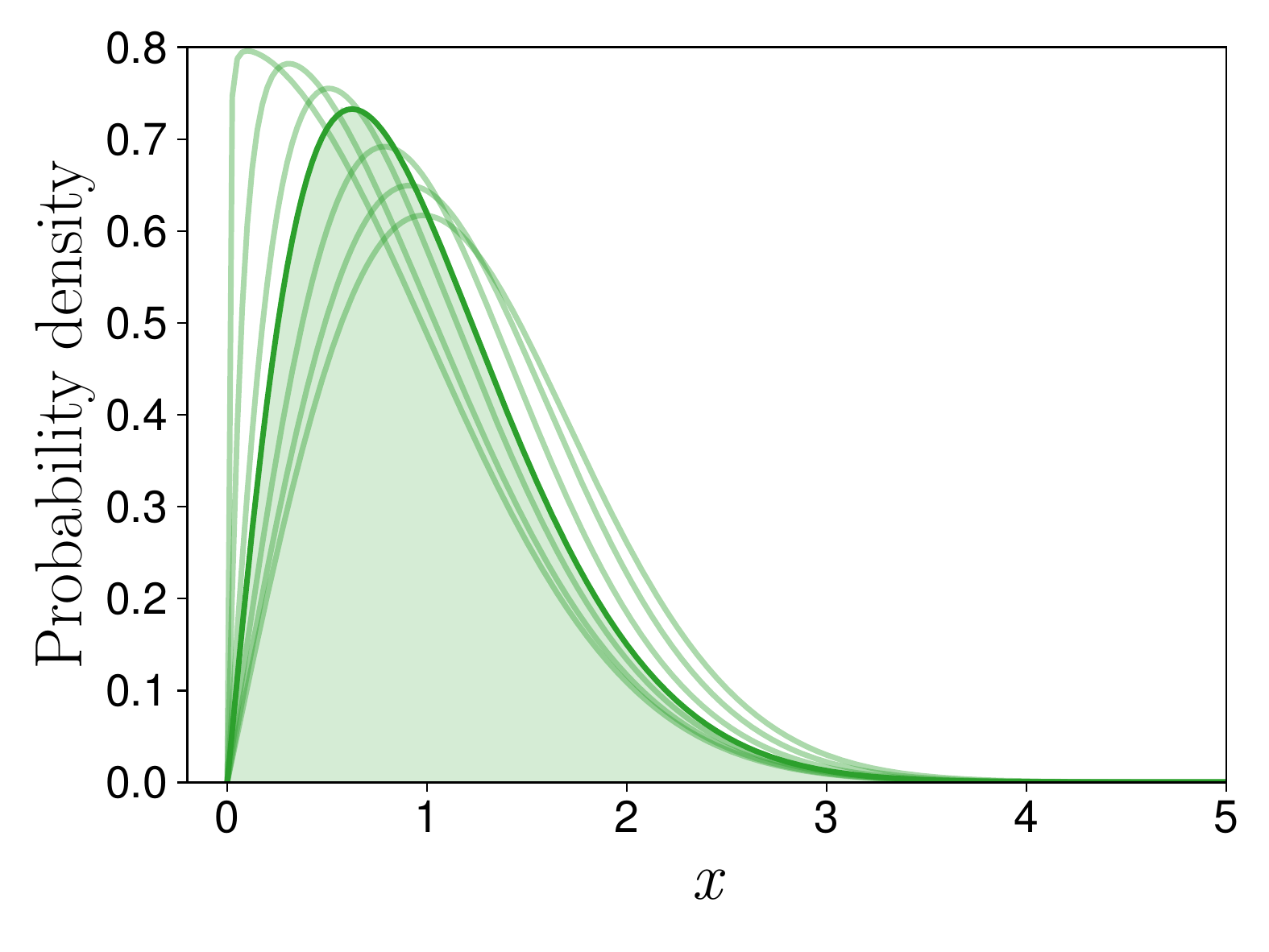}
		\caption{\tiny $c \in [0.01,0.1,0.3,{\bf 0.5},1.0,2.0,5.0]$}
          \label{trunkplot}
	\end{subfigure}
%%%%%%%%%%%%%%
	\begin{subfigure}{.32\textwidth}
		\includegraphics[width=\textwidth]{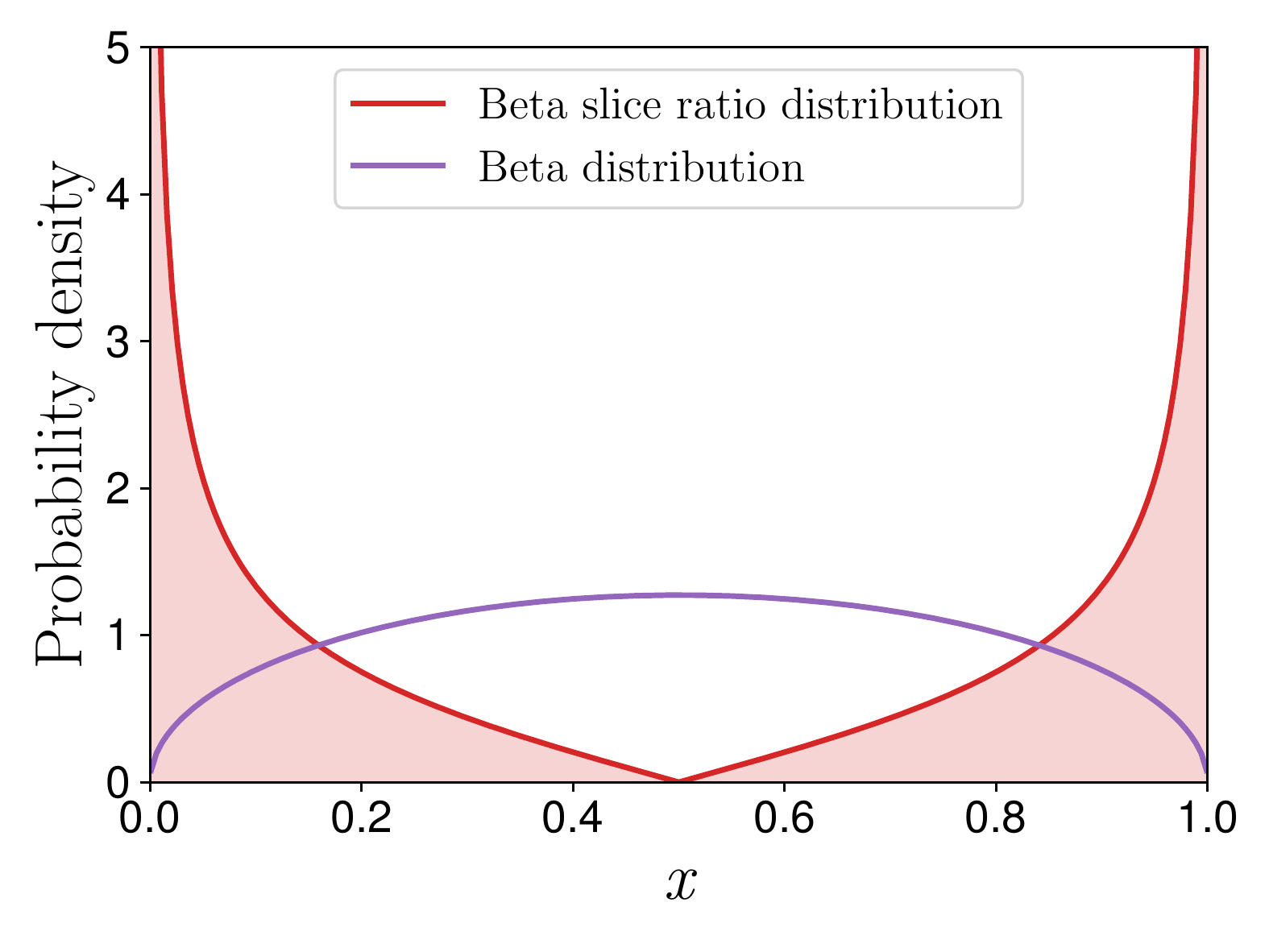}
		\caption{\tiny $\alpha = 1.5$}
          \label{betaplot}
	\end{subfigure}
%%%%%%%%%%%%%%
	\caption{Scaling of truncated ratio gradient accuracy with the dimension (\ref{varscaleplot}), truncated ratio distribution for various $c$ (\ref{trunkplot}) and
          the slice ratio distribution for the Beta distribution (\ref{betaplot}).}
          \label{truncandbeta}
\end{figure*}

For a Gaussian $\p{x;\theta}$, this equation can be solved
(App.~\ref{truncderivapp}). We call
the result the truncated ratio gradient (TRRG). The pdf, sampling
method and gradient estimator are below:

\begin{equation}
\begin{aligned}
  &\pind{tr}{x;c,\mu,\sigma} \\&= \frac{\exp(-\frac{c^2}{2})}{1-\Phi(c)}
  \frac{1}{\sigma2\sqrt{2\pi}}\frac{|x-\mu|}{\sqrt{(x-\mu)^2+\sigma^2c^2}}
  \exp(-\frac{(x-\mu)^2}{2\sigma^2}) \\ &\textup{where}~ \Phi(c) \textup{ is the cdf
  of a unit normal distribution},\\
  &x =
  \mu \pm \sigma\sqrt{\epsilon_c^2 - c^2} ~~~\textup{where}~~~
  \epsilon_c \sim \textup{truncG}(c,\infty), \textup{and}\\
  & \textup{truncG}(a,b)
  \textup{ is the unit normal truncated\footnotemark between } [a,b]\\
&\deriv{}{\mu}\expectw{x\sim\p{x}}{\phi(x)} \\&=
\expectw{x\sim q(x)}{\textup{sgn}(x-\mu)
    \frac{2\epsilon_c}{\sigma}\frac{1-\Phi(c)}{\exp(-\frac{c^2}{2})}\phi(x)}.
\end{aligned}
\end{equation}
This distribution interpolates between a Gaussian distribution and the
B-distribution.
\footnotetext{By truncated we mean that the probability density is set
  to 0 outside these bounds, and the remaining probability
  distribution is renormalized. Such a distribution is
  implemented e.g. in MATLAB and scipy \citep{scipy}.}
The interpolation is controlled by the $c$ parameter: for $c=0$ the
distribution is Gaussian, and for $c \rightarrow \infty$ the
distribution tends to the B-distribution.  One half of the
distribution is plotted in Fig.~\ref{trunkplot} for several values of
$c$, and Fig.~\ref{varscaleplot} shows how the accuracy of
$\deriv{p}{\mu}/q$, and the variance of the distribution scale with
$c$ (we name these functions $t(c)$ and $v(c)$ respectively). These
functions were computed analytically (App.~\ref{truncderivapp}). How
should one pick the parameter $c$? A simple choice may be to pick $c$
around $0.5$, where the accuracy starts increasing slower than the
variance of the distribution.  But is there a more principled method
based on the dimensionality?

\begin{table*}
  \caption{Guidelines for choosing the offset parameter
    $c$ for the truncated ratio gradient.}
\vskip 0.1in
\label{cchoice}
\centering
\begin{tabular}{rllllllll}%{lcccccccccc}
\toprule
Suggested parameter $c$  & 0.1 & 0.2 & 0.3 & 0.4 & 0.5 & 0.6 & 0.8 & 1.0\\
%\midrule
Dimension $(D-1)$ & 4523 & 676 & 238 & 119 & 71 & 48 & 27 & 19\\
Exp. increase in accuracy $t$ & 1.076 & 1.144 & 1.204 & 1.257 & 1.302 & 1.341 & 1.402 & 1.447\\
\bottomrule
\end{tabular}
\vskip -0.1in
\end{table*}

We give a short analysis of the effect of the variance and guidelines
for picking $c$. For example, consider the case when $\phi$ is linear
with slope $a$ in every dimension, the dimensionality is $D$ and the
variance is scaled by $v_c$, then the variance of $\phi$ would
increase by a factor $v_c$ to $a^2\sigma^2Dv_c$. The noise from the
other dimensions would scale as roughly $v_c(D-1)a^2\sigma^2$. However,
the increase in accuracy $t_c$ counteracts this increase in noise, and
the gradient variance of this noise scales as
$(\nicefrac{v_c}{t_c})(D-1)a^2\sigma^2$. Now if we assume that the
gradient signal has a variance around $a^2\sigma^2$, and we want to
guarantee that the additional gradient noise from the other dimensions
does not exceed the maximum decrease in variance, then we could pick
$c$ s.t. $(\nicefrac{v_c}{t_c}-1)(D-1)\approx 1$. In
Tab.~\ref{cchoice}, we show several values of
$1/(\nicefrac{v_c}{t_c}-1)$ and the expected increase in accuracy
$t_c$, which can be used as a guideline for picking an appropriate $c$
for the dimensionality of your problem. We could also estimate the
reduction in gradient signal variance as
$(1-\nicefrac{1}{t_c})a^2\sigma^2$ for a more conservative estimate of
$D$, but in practice, the reduction in gradient signal variance is
greater than $\nicefrac{1}{t_c}$ because of structure in $\phi$.  In
general, for deterministic problems it may be better to be
conservative and aim for a smaller increase in accuracy with a smaller
$c$, whereas if $\phi$ is stochastic, then the additional variance
from other dimensions may be negligible and higher $c$ values can be
used.

\section{Experiments to verify theory}

We performed experiments on a quadratic $\phi(\bfv{x})$ to verify the
theory. In App.~\ref{evolstrategies} we also evaluate our methods in
evolution strategies experiments in reinforcement learning, but as our
work proved that importance sampling for Gaussian base distributions
can only lead to modest gains at best, it is difficult to obtain
statistically significant results. On the other hand, our method was
crucial for obtaining competitive results using a Beta
distribution, as emphasized by our experiment here. \remove{Here we
  emphasize that the slice ratio gradient method is crucial for some
  non-Gaussian distributions, e.g. for our example with a Beta
  distribution.}

\begin{figure*}[!t]
        \centering
	\begin{subfigure}{.49\textwidth}
		\includegraphics[width=\textwidth]{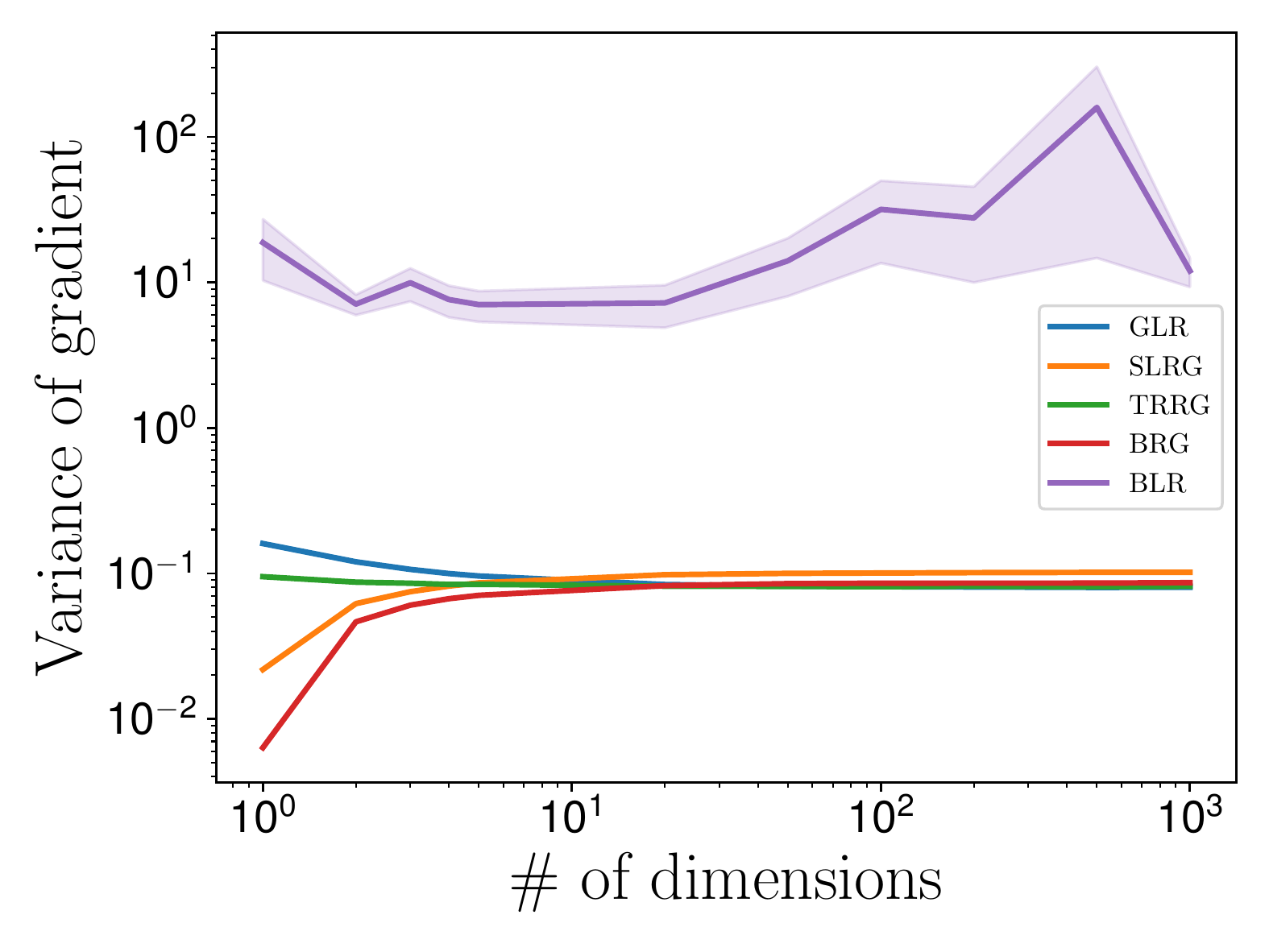}
		\caption{Deterministic}
          \label{detquad}
        \end{subfigure}
%%%%%%%%%%%%%%
	\begin{subfigure}{.49\textwidth}
		\includegraphics[width=\textwidth]{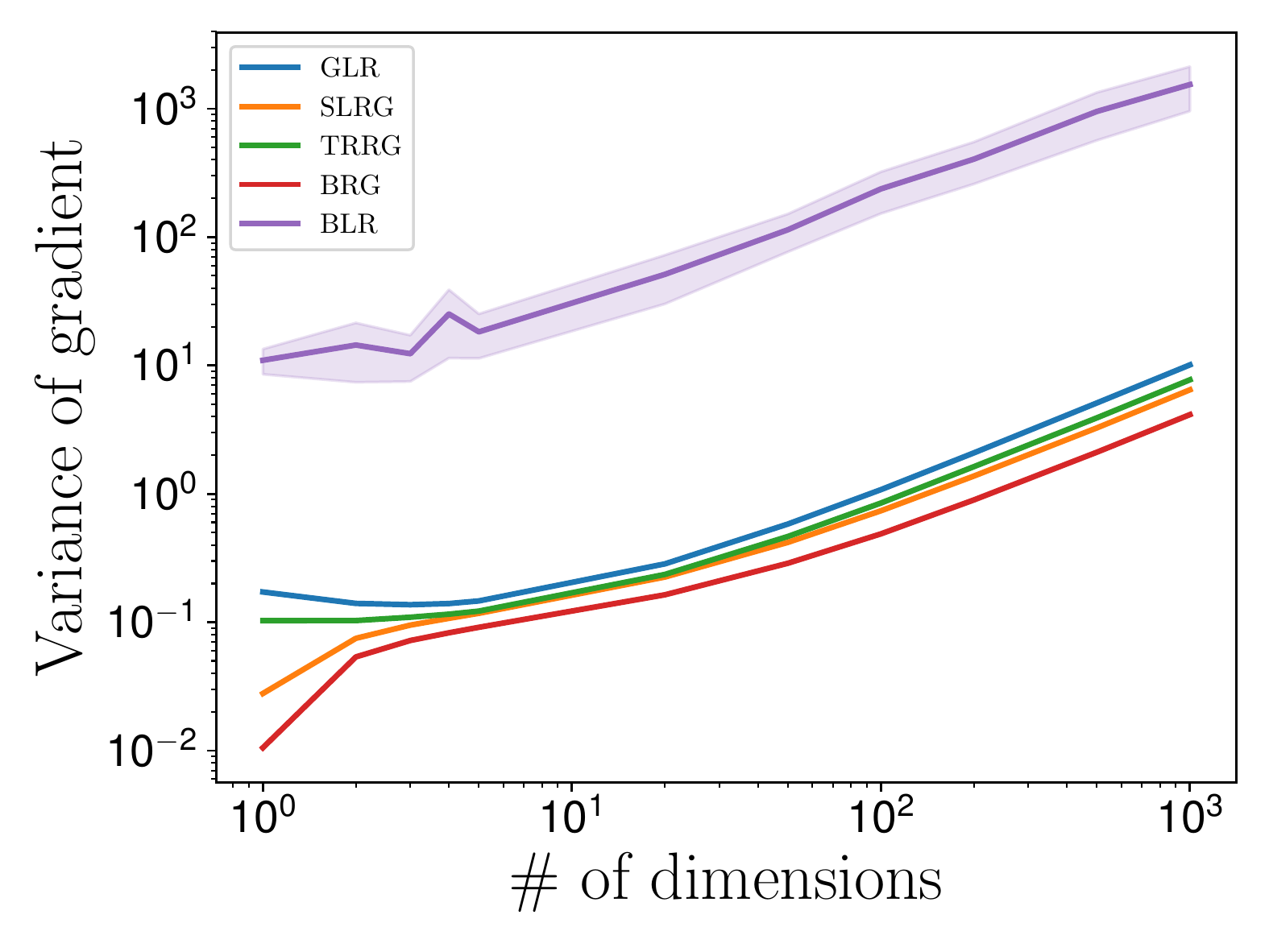}
		\caption{Noisy}
          \label{noisyquad}
	\end{subfigure}
%%%%%%%%%%%%%%
	\caption{Scaling of gradient estimator variance on a quadratic
          problem with deterministic and noisy function
          evaluations. The confidence intervals correspond to one
          standard deviation of the estimate.}
          \label{quadexps}
\end{figure*}

\paragraph{Setup:} $\phi(\bfv{x})$ is a quadratic
$(\bfv{x}-\bfv{a})^TQ(\bfv{x}-\bfv{a})$, where $\bfv{a} = \bfv{1}$ and
$Q = \textup{ones}(D,D)/D^2$ is a matrix of ones, which is scaled,
such that $\phi(\bfv{x})$ remains constant at $\bfv{x}=\bfv{0}$. We
evaluate a deterministic case, as well as a case when Gaussian noise
$\sigma_n^2=1$ is added on $\phi$. We vary the dimension between
1--1000, and plot the variance of the gradient estimators: GLR---LR
gradient with a Gaussian $p(x)$; SLRG---slice ratio gradient with a
Gaussian $p(x)$; TRRG---truncated ratio gradient with $c=0.5$;
BRG---slice ratio gradient with a Beta $p(x)$, and $\alpha=1.5$,
plotted in Fig.~\ref{betaplot}; BLR---LR gradient with a Beta $p(x)$,
and $\alpha=1.5$. The mean of the distributions was set to $0$ and the
variance was set to $1$ (the Beta distributions were stretched by
$k=2\sigma\sqrt{2\alpha+1}$ to achieve this). We used antithetic
sampling, so that the effect of any baseline could be ignored. The
gradient was estimated by averaging 100 samples, and this was repeated
for a large number of times to estimate the variance of the gradient
estimator. Bootstrapping was used to obtain confidence intervals. The
results are plotted in Fig.~\ref{quadexps}.

\paragraph{Results and analysis:} The main result is that using the
slice ratio method, the gradient accuracy for the Beta distribution
could be increased by 100--1000 times (compare BRG to BLR),
showing that our method is necessary for some
non-Gaussian distributions. In general, the increase in gradient
accuracy would tend to $\infty$ as the $\alpha$ parameter tends to 1
from above; however, even for moderately curved cases, such as
$\alpha=1.5$ (Fig.~\ref{betaplot}) the improvement in accuracy can be
drastic.

The results confirm our theoretical analysis: in the
deterministic case, the SLRG method outperforms the standard GLR
method, but as the dimensionality is increased, this reverses; whereas
in the high-noise case, SLRG always outperforms GLR. In the noisy
case, the gradient variances at $D=1000$ are GLR: $10.10 \pm 0.05$,
SLRG: $6.46 \pm 0.03$, TRRG: $7.73 \pm 0.04$, BRG: $4.14 \pm 0.02$
(the errorbars correspond to 1 standard deviation). The ratios
$10.10/6.46 = 1.563$ and $10.10/7.73=1.307$ match the theoretical
improvements in gradient accuracy for the SLRG gradient at large $c$
in Fig.~\ref{varscaleplot} and for the TRRG gradient at $c=0.5$ in
Tab.~\ref{cchoice}. In the deterministic case, the gradient variances
at $D=1000$ are GLR: $0.0803 \pm 0.0004$, SLRG: $0.1015 \pm 0.0005$,
TRRG: $0.0815 \pm 0.0004$, BRG: $0.0864 \pm 0.0004$, showing that TRRG
is more robust than SLRG to problems arising from increasing the
dimension, while it still allows reducing the variance in the
stochastic $\phi$ setting. Interestingly, BRG achieved a lower
gradient variance than SLRG in the deterministic setting, and was overall
the best in the stochastic setting even though the variances of the
base distributions were the same.

\section{Conclusions}

We have introduced a new unified theory of LR and RP gradients. The
theory explained that the sampling distribution $q(x)$ for LR
gradients is a separate matter to the distribution $p(x;\theta)$ used
to compute the objective function
$\expectw{x\sim\p{x;\theta}}{\phi(x)}$, and motivated us to search for
the optimal importance sampling distribution $q(x)$ to reduce gradient
variance. We derived these importance sampling distributions together
with sampling methods for them for Gaussian and Beta objective
distributions $\p{x;\theta}$ to reduce the variance of the gradient
w.r.t. a mean shifting parameter of the distribution. Optimal
sampling for other gradients is left for future work. We further
analyzed the scalability with the dimension of the sampling space.
Gaussian distributions are widely used in the literature, and we found
that our method is able to provide a modest improvement in gradient
accuracy. On the other hand, for distributions with a ``flat top",
which have found less use, our method can drastically improve the
accuracy, and is crucial for obtaining good results. Which objective
distributions outperform Gaussians in which situations is a
substantial research topic: e.g. clipped distributions
\citep{fujita2018clipped}, Beta distributions \citep{chou2017betapol},
exponential family distributions \citep{eisenach2019expfamily} or
normalizing flows \citep{tang2018normalizing,mazoure2019normalizing2}
have been considered, but they did not importance sample from
$q(x)$. Our slice ratio gradients will be essential to obtain a fair
comparison between different $\p{x;\theta}$.

\subsubsection*{Acknowledgments}
PP was supported by OIST Graduate School funding and by RIKEN.
MS was supported by KAKENHI 17H00757.
%We thank the anonymous reviewers for useful comments.
%This work was supported by OIST Graduate School funding and by
%JSPS KAKENHI Grant Number
%JP16H06563 and JP16K21738.

\bibliography{unifgrad}
\bibliographystyle{apalike}

\newpage

\begin{appendices}

\onecolumn
  
\section{Likelihood ratio gradient basics}
\label{lrbasicsapp}

The likelihood ratio (LR) gradient estimator is given by

\begin{equation}
  \deriv{}{\theta}\expectw{x\sim\p{x;\theta}}{\phi(x)} =
  \expectw{x\sim\p{x;\theta}}{\deriv{\log\p{x;\theta}}{\theta}\phi(x)}.
\end{equation}

For a Gaussian $\p{x;\theta}$:

\begin{equation}
\begin{aligned}
  \log\p{x;\theta} &= -\frac{1}{2}\log(2\pi) - \log(\sigma) -
  \frac{(x-\mu)^2}{2\sigma^2},\\
  \deriv{\log\p{x;\theta}}{\mu} &= \frac{x-\mu}{\sigma^2} =
  \frac{\epsilon}{\sigma},\\
  \deriv{\log\p{x;\theta}}{\sigma} &= \frac{(x-\mu)^2}{\sigma^3} -
  \frac{1}{\sigma} = \frac{\epsilon^2}{\sigma} - \frac{1}{\sigma},\\
  \textup{where }~~&x=\mu + \epsilon\sigma \textup{ and }\epsilon\sim\mathcal{N}(0,1).
\end{aligned}
\end{equation}

For a Beta $\p{x;\theta}$:

\begin{equation}
\begin{aligned}
  \p{x;\theta} &=
  \frac{(x-x^2)^{\alpha-1}}{B(\alpha,\alpha)}~~\textup{for }x\in[0,1]\\
  \log\p{x;\theta} &=  -\log\left(B(\alpha,\alpha)\right)
  + (\alpha-1)\log(x-x^2),\\
  \deriv{\log\p{x;\theta}}{x} &= \frac{\alpha-1}{x-x^2}(1-2x),\\
  \deriv{\log\p{x;\theta}}{\mu} &= -\deriv{\log\p{x;\theta}}{x},~~\\
  \textup{where }\mu&\textup{ is a shifting parameter for the mean.}
\end{aligned}
\end{equation}

In practice, we sample an
$\epsilon_\beta = x - 0.5 \Rightarrow x = \epsilon_\beta+0.5$, then
the gradient estimator becomes:
$\frac{2\epsilon_\beta(\alpha-1)}{0.25-\epsilon_\beta^2}\frac{1}{k}$, where the
additional $k$ factor comes if a stretching is applied: $z = \mu + k\epsilon_\beta$.

\paragraph{Baselines to reduce gradient variance:} The LR gradient
estimator on its own has a large variance, and techniques have to be used
to stabilize it. A common technique is to subtract a constant baseline $b$
from the $\phi(x)$ values, so that the gradient estimator becomes

\begin{equation}
  \deriv{}{\theta}\expectw{x\sim\p{x;\theta}}
  {\deriv{\log\p{x;\theta}}{\theta}\left(\phi(x)-b\right)}.
\end{equation}

In practice, using $b = \expectw{x\sim\p{x;\theta}}{\phi(x)}$ works
well, but one can also derive an optimal baseline
\citep{weaver2001optimalbaseline}. We outline the derivation below. The
gradient variance when a baseline is used can be expressed as

\begin{equation}
\label{eq:gradvarderiv}
  \begin{aligned}
  \variancew{x\sim\p{x;\theta}}
  {\deriv{\log\p{x;\theta}}{\theta}\left(\phi(x)-b\right)} &=
  ~~\expectw{x\sim\p{x;\theta}}{\left(\deriv{\log\p{x;\theta}}{\theta}\phi(x)\right)^2}\\
  &-2\expectw{x\sim\p{x;\theta}}{\left(\deriv{\log\p{x;\theta}}{\theta}\right)^2\phi(x)b}
  + \expectw{x\sim\p{x;\theta}}{\left(\deriv{\log\p{x;\theta}}{\theta}b\right)^2}.
  \end{aligned}
\end{equation}

Taking the derivative of Eq.~(\ref{eq:gradvarderiv}) w.r.t. $b$ and setting
to zero gives the optimal baseline as

\begin{equation}
  b_{opt} = \frac{
    \expectw{ x\sim\p{x;\theta} }
    { \left(\deriv{\log\p{x;\theta}}{\theta}
      \right)^2\phi(x) }
  }
  {\expectw{ x\sim\p{x;\theta} }
    { \left(\deriv{\log\p{x;\theta}}{\theta}
      \right)^2}}.
\end{equation}

In practice, for example if $\phi(x)$ is linear and $\p{x;\theta}$ is
 Gaussian then $b_{opt} = \expectw{x\sim\p{x;\theta}}{\phi(x)}$, so
the gain from trying to use an optimal baseline is often small.  What
would happen to the optimal baseline derivation for our importance
sampling case (Sec.~\ref{sliceratio})? The sampling distribution has
to be swapped with $q(x)$, and $\deriv{\log\p{x;\theta}}{\theta}$ has
to be swapped with $\nicefrac{\deriv{\p{x;\theta}}{\theta}}{q(x)}$,
giving the optimal baseline \citep{jie2010connection} as

\begin{equation}
  b_{opt} = \frac{
    \expectw{ x\sim q(x) }
    { \left(\nicefrac{\deriv{\p{x;\theta}}{\theta}}{q(x)}
      \right)^2\phi(x) }
  }
  {\expectw{ x\sim q(x) }
    { \left(\nicefrac{\deriv{\p{x;\theta}}{\theta}}{q(x)}
      \right)^2}}.
\end{equation}

Note that if the slice ratio distribution is used, then
$q = \left|\deriv{\p{x;\theta}}{\theta}\right|/\sqrt{\lambda}$, and
$b_{opt} = \expectw{x\sim q(x)}{\phi(x)}$.

\paragraph{Antithetic sampling:} An often used technique is to sample
points $x$ in pairs opposite to each other, s.t.
$x_+ = \mu + \sigma\epsilon$ and $x_- = \mu - \sigma\epsilon$. This
technique is particularly often used in evolution strategies' research
\citep{salimans2017oaies,mania2018simplers}. We will explain that when
this technique is used, then a baseline has no effect because it cancels. Thus, using
antithetic sampling allows us to disentangle any effect of the
baselines from the effect of the importance sampling, which is why we
use it in all of our experiments. The derivation is easy to see by
considering that for a Gaussian:
$\deriv{\log\p{x;\theta}}{\mu} = \frac{\epsilon}{\sigma}$, so
$\deriv{\log\p{x_+;\theta}}{\mu}(\phi(x_+)-b) +
\deriv{\log\p{x_-;\theta}}{\mu}(\phi(x_-)-b) =
\frac{\epsilon}{\sigma}\left(\phi(x_+) -b - (\phi(x_-) - b)\right) =
\frac{\epsilon}{\sigma}\left(\phi(x_+) - \phi(x_-)\right)$. In
general, this result holds for any symmetric
$\nicefrac{\deriv{\p{x;\theta}}{\theta}}{q(x)}$.

\paragraph{Relationship to finite difference methods:} Finite
difference methods also use the function values $\phi(x)$ to estimate
a derivative, so it may appear that the LR gradient estimator is a
finite difference estimator. Finite difference estimators work by
estimating the slope of the function, by evaluating the change between
two points, i.e.
\begin{equation}
\deriv{\phi(x)}{x} \approx \frac{\phi(x_+) - \phi(x_-)}{\Delta x}.
\end{equation}
In the antithetic sampling case, $\Delta x = 2\sigma\epsilon$, so
the estimator is 
\begin{equation}
\deriv{\phi(x)}{x} \approx \frac{\phi(x_+) - \phi(x_-)}{2\sigma\epsilon}.
\end{equation}
Clearly, this is different to the LR gradient estimator
\begin{equation}
  \frac{\epsilon\left(\phi(x_+) - \phi(x_-)\right)}{2\sigma},
\end{equation}
because the $\epsilon$ is in the wrong place. In Sec.~\ref{boxtheor} we explain
that the LR gradient estimator is a different concept to finite differences, which
is not trying to fit a linear function onto $\phi(x)$.

\section{Derivations for the probability flow theory}
\label{probflowderivapp}
Here we illustrate the background information in 3 dimensions, but it generalizes
straightforwardly to higher dimensions.

\paragraph{Notation:}\mbox{}\\
$\bfv{F} = [F_x(x,y,z), F_y(x,y,z), F_z(x,y,z)]$ is a vector field. \\
$\phi(x,y,z)$ is a scalar field (a scalar function)\\
Div operator:
$\nabla\cdot \bfv{F} = \pderivw{F_x}{x} + \pderivw{F_y}{y} +
\pderivw{F_z}{z}$. \\Grad operator:
$\nabla\phi = [\pderivw{\phi}{x}, \pderivw{\phi}{y},
\pderivw{\phi}{z}]$.

\subsection{Basic vector calculus and fluid mechanics}

\begin{figure*}[!t]
        \centering
\includegraphics[width=.6\textwidth]{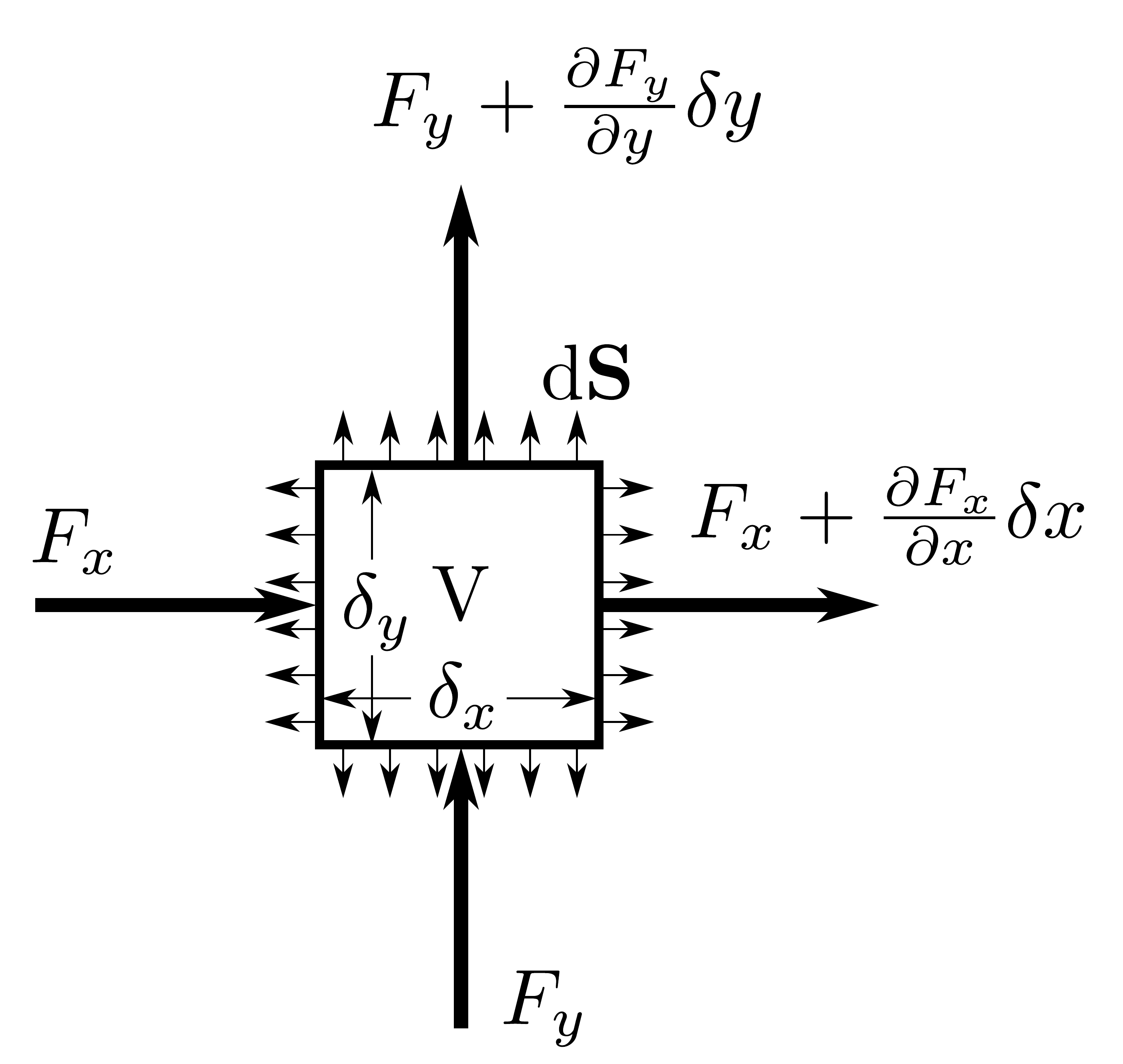}
\caption{Illustration of the divergence theorem.}
          \label{divtheorem}
\end{figure*}

The vector field $\bfv{F}$ could be for example thought of as a local
flow velocity for some fluid. If $\bfv{F}$ is the density flow rate, then
the div operator essentially measures how much the density is decreasing at
a point. If the outflow is larger than the inflow, the density would
decrease and vice versa. The divergence theorem, illustrated in Fig.~\ref{divtheorem}
illustrates how this change in density can be measured in two separate ways:
one could integrate the divergence across the volume, or one could integrate the
in and and outflow across the surface. The divergence theorem states:

\begin{equation}
  \int_V \nabla\cdot\bfv{F}\textup{d}V = \int_S\bfv{F}\cdot\textup{d}\bfv{S}
\end{equation}

To prove the claim, consider the infinitesimal box in
Fig.~\ref{divtheorem}.  The divergence can be calculated as
$\delta x\delta y(\pderivw{F_x}{x} + \pderivw{F_y}{y})$. On the other
hand, to take the integral across the surface, note that the surface
normals point outwards, and the integral becomes
$\delta_y(-F_x + F_x - \pderivw{F_x}{x}\delta x) + \delta x(-F_y + F_y
+ \pderivw{F_y}{y}\delta y) = \delta x\delta y(\pderivw{F_x}{x} +
\pderivw{F_y}{y})$, which is the same as the divergence. To generalize this
to arbitrarily large volumes, notice that if one stacks the boxes next to each
other, then the surface integral across the area where the boxes meet cancels
out, and only the integral across the outer surface remains. For an
incompressible flow, the density does not change, and the divergence
must be zero.

\subsection{Derivation of probability surface integral}
  \label{surfDeriv}
We will show that the LR estimator tries to
integrate
$\int_S\phi( \tilde{\bfv{x}})\nabla_\theta
\tilde{g}(\epsilon_x,\epsilon_h)\textup{d}\bfv{S}$. First, note that
$\textup{d}\bfv{S} = \hat{\bfv{n}}\textup{d}S$, and it is necessary to
express the normalized surface vector $\hat{\bfv{n}}$. To do so, we first
express the tangent vector $\bfv{t}$, then change the height component
of this vector to obtain a vector perpendicular to the tangent vector
(this is exactly the normal vector).

A vector tangent and downhill to the surface is given by
$\bfv{t} = [-\dpx,-\left(\dpx\right)\left(\dpx\right)^T]$. The normal
vector $\bfv{n}$ is $[-\dpx;h]$, such that $\bfv{t}\cdot\bfv{n} = 0$.
Therefore,
$\left(\dpx\right)\left(\dpx\right)^T -
\left(\dpx\right)\left(\dpx\right)^Th = 0 ~~\Rightarrow~~ h = 1$.
Finally, we normalize the vector:
\begin{equation}
  \hat{\bfv{n}} = \nicefrac{[-\dpx,1]}{\sqrt{\left(\dpx\right)
      \left(\dpx\right)^T+ 1}}~.
  \label{eq:n}
\end{equation}

Next, we perform a change of coordinates from the surface elements
$\textup{d}S$ to cartesian coordinates $\textup{d}\bfv{x}$.  When
projecting a surface element $\textup{d}S$ with unit normal
$\hat{\bfv{n}}$ to a plane with unit normal $\hat{\bfv{m}}$, the
projected area is given by
$\textup{d}\bfv{x} =
\left|\hat{\bfv{n}}\cdot\hat{\bfv{m}}\right|\textup{d}S$, therefore
$\textup{d}\bfv{x} =
\textup{d}S\left|\frac{1}{\sqrt{\left(\dpx\right)
      \left(\dpx\right)^T+ 1}}[-\dpx,1]\cdot[\bfv{0},1]\right| =
\nicefrac{\textup{d}S}{\sqrt{\left(\dpx\right) \left(\dpx\right)^T+ 1}}
$, from which we get
\begin{equation}
  \textup{d}S =
  \sqrt{\left(\dpx\right) \left(\dpx\right)^T+ 1}~~\textup{d}\bfv{x}.
  \label{eq:dS}
\end{equation}

Plugging Eqs.~(\ref{eq:n}) and (\ref{eq:dS}) into the right-hand side
of Eq.~(\ref{surfint}) we get
\begin{equation}
\begin{aligned}
\int_{X} \phi(
\tilde{\bfv{x}})\nabla_\theta \tilde{g}(\epsilon_x,\epsilon_h)
\cdot
\frac{[-\dpx,1]}{\sqrt{\left(\dpx\right)
      \left(\dpx\right)^T+ 1}}
  \sqrt{\left(\dpx\right) \left(\dpx\right)^T+ 1}~~\textup{d}\bfv{x}
  &=\\
\int_{X} \phi(
\tilde{\bfv{x}})\nabla_\theta \tilde{g}(\epsilon_x,\epsilon_h)
&\cdot
[-\dpx,1]~\textup{d}\bfv{x}.
\label{eq:normprod}
\end{aligned}
\end{equation}

Recall that the last element of $\tilde{g}(\epsilon_x,\epsilon_h)$ is
$\epsilon_h\p{g(\epsilon_x);\theta}$, and that $\epsilon_h$ at the
boundary surface is $1$, then the
$\nabla_\theta \tilde{g}(\epsilon_x,\epsilon_h)\cdot [-\dpx,1]$ term turns
into
$-\nabla_\theta g(\epsilon_x)\cdot\dpx +
\pderiv{\epsilon_h\p{g(\epsilon_x);\theta}}
{\theta}{\epsilon_x=const,\epsilon_h=1}$. The last term
$\pderiv{\p{g(\epsilon_x);\theta}}
{\theta}{\epsilon_x=const}$ can be thought of as the rate
of change of the probability density while following a point moving in the flow
induced by perturbing $\theta$. This quantity can be expressed with
the material derivative
$\pderiv{\p{g(\epsilon_x);\theta}}
{\theta}{\epsilon_x=const} =
\deriv{\p{\bfv{x};\theta}}{\theta} + \nabla_\theta
g(\epsilon_x)\cdot\dpx$. Finally, substituting into Eq.~(\ref{eq:normprod}):

\begin{equation}
\begin{aligned}
  \int_{S} \phi(\tilde{\bfv{x}})\nabla_\theta
  \tilde{g}(\epsilon_x,\epsilon_h)~\textup{d}\bfv{S} =
  \int_{X}
  \phi(\bfv{x})\deriv{\p{\bfv{x};\theta}}{\theta}
  ~\textup{d}\bfv{x}.
\label{eq:subsapp}
\end{aligned}
\end{equation}

\remove{We have already seen that a Monte Carlo integration of
the right hand side of Eq.~(\ref{eq:subs}) using samples from
$\p{\bfv{x};\theta}$ gives rise to the LR gradient
estimator. Thus, the RP gradient estimator and the LR gradient
estimator are duals under the divergence theorem. To further strengthen this
claim we prove that the LR gradient estimator is the unique estimator
that takes weighted averages of the function values $\phi(\bfv{x})$.}

\section{Reparameterization gradients are not unique}
\label{rpNotUnique}

What happens if we perform the same kind of analysis as in
Theorem~\ref{lrunique} for the RP gradient?  Similarly, suppose that
there exist $u(\bfv{x})$ and $v(\bfv{x})$, s.t.
$\int \nabla\phi(\bfv{x})\cdot u(\bfv{x})~\textup{d}\bfv{x} = \int
\nabla\phi(\bfv{x})\cdot v(\bfv{x})~\textup{d}\bfv{x}$ for any
$\phi(\bfv{x})$.  Rearrange the equation into
$\int
\nabla\phi(\bfv{x})\cdot\left(u(\bfv{x})-v(\bfv{x})\right)~\textup{d}\bfv{x}
= 0$.  Then, if we can pick
$\nabla\phi(\bfv{x}) = u(\bfv{x})-v(\bfv{x})$ it would lead to $u=v$,
which would show the uniqueness. However, it is not necessarily
possible to pick such $\phi(\bfv{x})$. In particular, the integral of
$\nabla\phi(\bfv{x})$ over any closed path is 0, but this is not
necessarily the case for $u-v$. Therefore, the same kind of analysis
does not lead to a claim of uniqueness. Indeed, concurrent work
\citep{jankowiak2018rpflow} showed that there are an infinite amount
of possible reparameterization gradients, and the minimum
variance\footnote{By minimum variance, we mean the minimum variance
  achievable without assuming knowledge of $\phi(\bfv{x})$, or
  alternatively that it is approximately linear in the sampling range,
  $\nabla\phi(\bfv{x}) \approx \bfv{A}$. Their result holds for
  arbitrary dimensionality.} is achieved by the optimal transport
flow.

\section{Slice integral importance sampling}
\label{sliceintegral}

From Theorem~\ref{lrunique} we saw that unlike the RP gradient case,
the weighting $\psi$ for function values $\phi(\bfv{x})$ with
$\bfv{x} \sim \p{\bfv{x};\theta}$ to obtain an unbiased estimator for
the gradient $\deriv{}{\theta}\expect{\phi(\bfv{x})}$ is unique. The
only option to reduce the variance by changing the weighting would
then be to sample from a different distribution
$q(\bfv{x})$ via importance sampling. Motivated by the
resemblance of the ``boxes" theory in Sec.~\ref{boxtheor} to the
Riemann integral, we propose to sample horizontal slices of
probability mass resembling the Lebesgue integral. Such an approach
appears attractive, because if the location of the slice is moved by
modifying the parameters of the distribution (e.g., by changing the
mean), then the derivative of the expected value of the integral over
the slice will depend only on the value at the edges of the slice
(because the probability density in the middle would not change). To
obtain the gradient estimator, it will only be necessary to compute
the probability density $\pind{L}{\bfv{x};\theta}$. We derive such a
``slice integral" distribution corresponding to the Gaussian
distribution. The method resembles the seminal work by
\cite{neal2003slice} on slice sampling in Markov chain Monte Carlo methods. We call our
new distribution the L-distribution, and it is plotted in
Fig.~\ref{ldist}.

\paragraph{Derivation of the pdf of the L-distribution:} One way to
sample whole slices of a probability distribution would be to sample a
height $h$ between 0 and $p_{\mathrm{max}}$ proportionally to the probability
mass at that height. The probability mass at a height $h$ is just
given by $2|x-\mu|$ where $x$ is such that $\p{x;\mu,\sigma} = h$,
i.e., $2|x-\mu|$ is the distance between the edges of
$\p{x;\mu,\sigma}$. The probability mass corresponding to $x$ is then
given by $2|x-\mu|\textup{d}h$. Performing a change of coordinates to the
$x$-domain, and splitting the mass between the two edges of the slice, we get
$|x-\mu|\textup{d}h = |x-\mu|\left|\deriv{\p{x;\mu,\sigma}}{x}\right|\textup{d}x$.  This
gives a closed-form normalized pdf for the L-distribution:

\begin{equation}
  \begin{aligned}
  \pind{L}{x;\mu,\sigma} = |x-\mu|\left|\deriv{\p{x;\mu,\sigma}}{x}\right|
  &= |x-\mu|\p{x;\mu,\sigma}\frac{|x-\mu|}{\sigma^2}\\
  &=\frac{|x-\mu|^2}{\sqrt{2\pi}~\sigma^3}
  \exp\left(\frac{-(x-\mu)^2}{2\sigma^2}\right).
\end{aligned}
\label{maxwell}
\end{equation}

One can recognize that Eq.~(\ref{maxwell}) is actually just a
Maxwell-Boltzmann distribution reflected about the origin with the
probability mass split between the two sides.

\paragraph{Sampling from the L-distribution:} To sample from this
distribution, it is necessary to sample points proportionally to the
length of the slices. It suffices to sample uniformly in the area
under the curve in the space augmented with the height dimension $h$,
then selecting the slice on which the sampled point lies. This can be
achieved with the three steps: 1) sample a point from the base
distribution: $x_s \sim \p{x;\mu,\sigma}$, 2) sample a height:
$h \sim \textup{unif}\left(0,\p{x_s;\mu,\sigma}\right)$, 3) compute where the
edge of slice is $x = p^{-1}(h;\mu,\sigma)$, where
$p^{-1}(h)$ inverts the pdf, and computes the location $x$
that gives a probability density $h$. For the L-distribution, this can be
achieved by sampling $\epsilon_x \sim \mathcal{N}(0,1)$ and
$\epsilon_h \sim \textup{unif}(0,1)$ and transforming these by the equation:
\begin{equation}
  x = \mu \pm\sigma\sqrt{-2\log(\epsilon_h) + \epsilon_x^2}~.
\end{equation}
Now it is straightfoward to obtain the LR gradient estimator:
\begin{equation}
  \begin{aligned}
  \deriv{}{\mu}\expectw{x\sim\p{x;\theta}}{\phi(x)} =
  \expectw{x\sim q(x;\theta)}{\frac{\deriv{p}{\mu}}{q(x;\theta)}\phi(x)} &=
  \expectw{x\sim q(x;\theta)}{\frac{1}{x-\mu}\phi(x)} \\
  &= \expect{\frac{\textup{sgn}(x-\mu)}{\sigma\sqrt{-2\log(\epsilon_h) + \epsilon_x^2}}\phi(x)}.
  \end{aligned}
\end{equation}

\section{Slice ratio gradient derivations}

\subsection{Slice ratio gradients for general distributions}

\begin{figure*}[!t]
        \centering
\includegraphics[width=.6\textwidth]{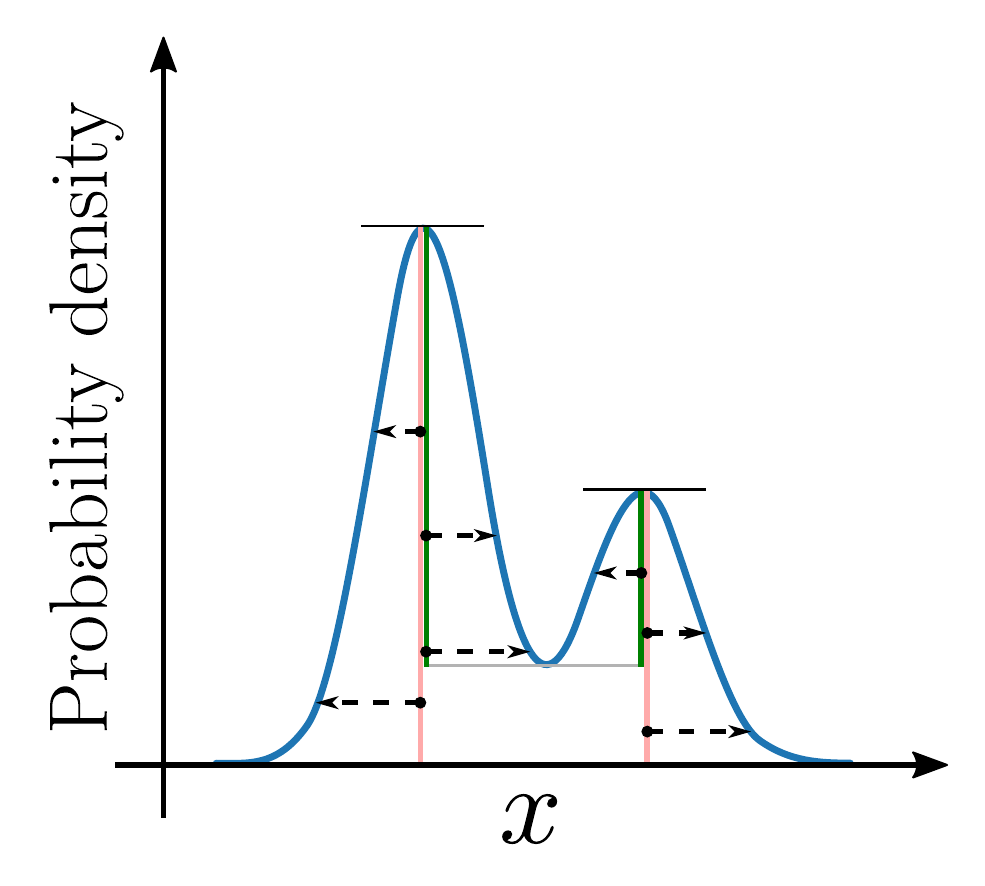}
\caption{Slice ratio sampling method for general multimodal distributions.}
          \label{fig:genslrg}
\end{figure*}

So far we have introduced slice ratio gradients for unimodal
distributions.  Here we explain that the technique works for arbitrary
distributions. The process is illustrated in
Fig.~\ref{fig:genslrg}. The curve $\p{x;\theta}$ is projected onto the
vertical dimension. Then one samples uniformly in this projected area,
and maps the sampled points back onto the curve via $x =
p^{-1}(h)$. The probability density in the $h$-space is uniformly
$1/H$, where $H$ is the total length of the vertical lines. Changing
coordinates will give $\textup{d}h/H =
|\deriv{p}{x}|\textup{d}x/H$. Thus, this sampling method will always
sample proportionally to $|\deriv{p}{x}|$. Because
$\deriv{p}{x} = -\deriv{p}{\mu}$ for arbitrary distributions, this
sampling method gives the desired importance sampling distribution to
minimize the variance of the gradient w.r.t. $\mu$ for arbitrary
distributions. The probability density becomes
\begin{equation}
q(x) = \frac{1}{H}\left|\deriv{p}{x}\right|,
\end{equation}
and the gradient estimator for one sample becomes
\begin{equation}
  \frac{\deriv{p}{\mu}}{\left|\deriv{p}{\mu}\right|/H}\phi(x) =
  \textup{sgn}\left(\deriv{p}{\mu}\right)H\phi(x).
\end{equation}

\subsection{Slice ratio gradient for a Gaussian distribution}
\label{slrgderiv}
The pdf is
\begin{equation}
\p{x} = \frac{1}{\sqrt{2\pi}\sigma}\exp\left(-\frac{(x-\mu)^2}{2\sigma^2}\right).
\end{equation}

The maximum probability density is at $x=\mu$:
\begin{equation}
p_{\mathrm{max}} =  \frac{1}{\sqrt{2\pi}\sigma}.
\end{equation}

The probability density for the slice ratio distribution can be
derived by performing a change in coordinates from the $h$ value to
$x$. The probability mass at a slice $\textup{d}h$ split between two sides is
$\nicefrac{\textup{d}h}{2p_{\mathrm{max}}}$, so

\begin{equation}
  \frac{1}{2p_{\mathrm{max}}}\textup{d}h =
  \frac{1}{2p_{\mathrm{max}}}\left|\deriv{p}{x}\right|\textup{d}x.
\end{equation}

From this we get

\begin{equation}
  \begin{aligned}
    \label{slrgqeq}
  q(x) &= \frac{\sqrt{2\pi}\sigma}{2}
  \left|\frac{1}{\sqrt{2\pi}\sigma}\exp\left(-\frac{(x-\mu)^2}{2\sigma^2}\right)
    \frac{-(x-\mu)}{\sigma^2}\right|\\
  &= \frac{\left|x-\mu\right|}{2\sigma^2}\exp\left(-\frac{(x-\mu)^2}{2\sigma^2}\right),
\end{aligned}
\end{equation}

which is the pdf in Eq.~(\ref{eq:bdist}).

To derive the sampling method, first derive the inverse of the
probability density $p^{-1}(h)$ as
\begin{equation}
\begin{aligned}
  h &=  \frac{1}{\sqrt{2\pi}\sigma}\exp\left(-\frac{(x-\mu)^2}{2\sigma^2}\right)\\
  \log(h) &= -\frac{1}{2}\log(2\pi) - \log(\sigma) -\frac{(x-\mu)^2}{2\sigma^2}\\
  (x-\mu)^2 &= -2\sigma^2\left(\frac{1}{2}\log(2\pi) + \log(\sigma) + \log(h)\right)\\
  x &= \mu \pm \sigma\sqrt{-\log(2\pi) - 2\log(\sigma) - 2\log(h)}~.
\end{aligned}
\end{equation}

Now, noting $h = p_{\mathrm{max}}\epsilon_h$, where $\epsilon_h\sim\textup{unif}(0,1)$,
we end up with the sampling method:

\begin{equation}
\begin{aligned}
  x &= \mu \pm \sigma\sqrt{-\log(2\pi) - 2\log(\sigma) -
    2\log(\frac{1}{\sqrt{2\pi}\sigma}\epsilon_h)}\\
   &= \mu \pm \sigma\sqrt{-2\log(\epsilon_h)}~, \textup{where }\epsilon_h\sim\textup{unif}(0,1).
\end{aligned}
\end{equation}

\subsection{Slice ratio gradient for a symmetric Beta distribution}
\label{slicebetaapp}

The pdf is
\begin{equation}
  \pind{\beta}{x} =
  \frac{x^{\alpha-1}(1-x)^{\alpha-1}}{B(\alpha,\alpha)} =
  \frac{(x-x^2)^{\alpha-1}}{B(\alpha,\alpha)}.
\end{equation}

The maximum probability density is at $x=0.5$:
\begin{equation}
p_{\mathrm{max}} =  \frac{0.25^{\alpha-1}}{B(\alpha,\alpha)}
\end{equation}

Similarly to Eq.~(\ref{slrgqeq}), the pdf of the slice ratio
distribution is $q(x) = \nicefrac{|\deriv{p}{x}|}{2p_{\mathrm{max}}}$:

\begin{equation}
\begin{aligned}
  q(x) &= \frac{B(\alpha,\alpha)}{2\times0.25^{\alpha-1}}
  \left|\frac{(x-x^2)^{\alpha-2}}{B(\alpha,\alpha)}(\alpha-1)(1-2x)\right|\\
  &= \frac{\alpha-1}{2\times0.25^{\alpha-1}}
  \left|(x-x^2)^{\alpha-2}(1-2x)\right|,
\end{aligned}
\end{equation}

which is the pdf in Eq.~(\ref{eq:betadist}).

To derive the sampling method, first derive the inverse of the
probability density $p^{-1}(h)$ as
\begin{equation}
\begin{aligned}
  h &=  \frac{(x-x^2)^{\alpha-1}}{B(\alpha,\alpha)}\\
  h^{\nicefrac{1}{(\alpha-1)}} &= \frac{(x-x^2)}{B(\alpha,\alpha)^{\nicefrac{1}{(\alpha-1)}}}\\
  x^2-x + \left(hB(\alpha,\alpha)\right)^{\nicefrac{1}{(\alpha-1)}} &= 0\\
  x &= \frac{1}{2} \pm \frac{1}{2}\sqrt{1 - 4(hB(\alpha,\alpha))^{\nicefrac{1}{(\alpha-1)}}}~.
\end{aligned}
\end{equation}

Now, noting $h = p_{\mathrm{max}}\epsilon_h$, where $\epsilon_h\sim\textup{unif}(0,1)$,
we end up with the sampling method:

\begin{equation}
\begin{aligned}
  x &= \frac{1}{2} \pm \frac{1}{2}\sqrt{1 -
    4\left(\epsilon_h\frac{0.25^{\alpha-1}}{B(\alpha,\alpha)}
      B(\alpha,\alpha)\right)^{\nicefrac{1}{(\alpha-1)}}}\\
   &= \frac{1}{2} \pm \frac{1}{2}\sqrt{1 - \epsilon_h^{\nicefrac{1}{(\alpha-1)}}}~, \textup{where }\epsilon_h\sim\textup{unif}(0,1).
\end{aligned}
\end{equation}

\paragraph{Stretching factor k to achieve variance $\sigma^2$:}
The variance of the Beta distribution is given by
$\frac{\alpha^2}{(2\alpha)^2(2\alpha+1)} = \frac{1}{4(2\alpha+1)}$.
We need $k^2\frac{1}{4(2\alpha+1)}=\sigma^2 \Rightarrow k = 2\sigma\sqrt{2\alpha+1}$.

\subsection{W-distribution for minimizing variance of
  $\deriv{}{\sigma}$}
\label{wdistDeriv}

For completeness, for a Gaussian we also derive the optimal sampling
distribution for the derivative w.r.t. $\sigma$. First note that
$\deriv{\p{x}}{\sigma} = \sigma\deriv{^2\p{x}}{x^2}$.  This expression
means that if we apply the same height sampling concept as used for
$\mu$ on the distribution proportional to
$\left|\deriv{\p{x}}{x}\right|$, we would obtain samples with
probability density proportional to
$\left|\deriv{^2\p{x}}{x^2}\right|$, and would hence be sampling from
the desired distribution. The required base distribution is just the
B-distribution (Eq.~(\ref{eq:bdist})), so we can perform the required
derivation.

The result
is given below:

\begin{equation}
\begin{aligned}   
  \pind{W}{x;\mu,\sigma} &= \frac{\sqrt{e}}{4\sigma}
  \left|
    \exp\left(\frac{-(x-\mu)^2}{2\sigma^2}\right)
    \left(
      \frac{|x-\mu|^2}{\sigma^2} - 1
    \right)
  \right|,
\\
x &= \mu \pm \sigma\sqrt{W(-\epsilon_h^2/e)} ~~~\textup{where}~~~
\epsilon_h \sim \textup{unif}(0,1),\\
\deriv{}{\sigma}\expectw{x\sim\p{x}}{\phi(x)} &=
\expectw{x\sim \pind{W}{x}}{\textup{sgn}\left(\frac{(x-\mu)^2}{\sigma^2}
    -1\right)
    \frac{2\sqrt{2}}{\sigma\sqrt{e\pi}}\phi(x)}.
\end{aligned}
\label{Weq}
\end{equation}

In the above equation, $W(x)$ is the Lambert W function
\citep{corless1996lambertw}---a function s.t. $z = W(ze^z)$. The
solution for $W$ is picked with equal probability from the $-1$ and
$0$ branches of $W$, and the $\pm$ is also sampled randomly with equal
probability. Efficient implementations of $W$ are available in common
numerical computation packages, such as scipy \citep{scipy} or MATLAB. We call the
result the W-distribution, and it is plotted in Fig.~\ref{wdist}. To
the best of our knowledge, this distribution does not exist in the
literature.

\paragraph{Derivation of W-distribution:}
We first derive the probability density $\pind{W}{x}$, then the
sampling scheme. The base distribution is $\pind{B}{x}$, and we apply
a transformation by which we sample the height $h$, and transorm this
to a point $x$ by using the inverse $x = p_B^{-1}(h)$, and
sampling uniformly between the $x$ values that satisfy the equation,
e.g. for the B-distribution in Fig.~\ref{bdist} there are usually 4
points for each $h$ value. Therefore
$\textup{d}h = \frac{1}{4\textup{max}(p_B)}
\left|\deriv{\pind{B}{x}}{x}\right|\textup{d}x$ and
$\pind{W}{x} = \frac{1}{4\textup{max}(p_B)}
\left|\deriv{\pind{B}{x}}{x}\right|$. The required derivative is given
by

\begin{equation}
\deriv{\pind{B}{x}}{x} = \frac{\textup{sgn}(x-\mu)}{2\sigma^2}
\exp\left(\frac{-(x-\mu)^2}{2\sigma^2}\right)
-
\frac{|x-\mu|}{2\sigma^2}
\exp\left(\frac{-(x-\mu)^2}{2\sigma^2}\right)\frac{(x-\mu)}{\sigma^2}.
\end{equation}

Setting the derivative to 0 gives the locations of the peaks at
$x = \mu \pm \sigma$. Evaluating $\pind{B}{x}$ at these locations in
Eq.~(\ref{eq:bdist}) gives the peak value as
\begin{equation}
  \textup{max}(p_B) =
  \frac{1}{2\sigma}\exp\left(-\nicefrac{1}{2}\right).
  \label{Bmax}
\end{equation}
Combining these results
gives the density in Eq.~(\ref{Weq}).

Deriving the sampling method, requires inverting $\pind{B}{x}$:

\begin{equation}
\begin{aligned}
h &= \frac{|x-\mu|}{2\sigma^2}
\exp\left(\frac{-(x-\mu)^2}{2\sigma^2}\right), ~~~\textup{let}~~~
t = \frac{(x-\mu)^2}{2\sigma^2}, \textup{then}\\
h &= \frac{t^{\nicefrac{1}{2}}}{\sigma\sqrt{2}}\exp\left(-t\right)\\
h^2 &= \frac{t}{2\sigma^2}\exp\left(-2t\right)\\
-4\sigma^2h^2 &= -2t\exp\left(-2t\right)\\
W\left(-4\sigma^2h^2\right) &= -2t.\\
\end{aligned}
\end{equation}

Now recalling that $h = p_{max}\epsilon_h, ~~~
\epsilon_h \sim \textup{unif}(0,1)$, where
$p_{max} = \frac{1}{2\sigma}\exp\left(-\nicefrac{1}{2}\right)$ from
Eq.~(\ref{Bmax}), and plugging in the value of $t$, gives the sampling
method in Eq.~(\ref{Weq}).

The gradient estimator can be found by computing
$\deriv{p}{\sigma}/p_W$. The derivative is given by
\begin{equation}
\begin{aligned}
  \deriv{}{\sigma}\left(\frac{1}{\sqrt{2\pi}\sigma}
    \exp\left(-\frac{(x-\mu)^2}{2\sigma^2}\right)\right) &=
  -\frac{1}{\sqrt{2\pi}\sigma^2}
  \exp\left(-\frac{(x-\mu)^2}{2\sigma^2}\right)
  +
  \frac{1}{\sqrt{2\pi}\sigma}
  \exp\left(-\frac{(x-\mu)^2}{2\sigma^2}\right)
  \left(\frac{(x-\mu)^2}{\sigma^3}\right)\\
  &= \frac{1}{\sqrt{2\pi}\sigma^2}
  \exp\left(-\frac{(x-\mu)^2}{2\sigma^2}\right)\left(
  \frac{(x-\mu)^2}{\sigma^2} - 1\right).
\end{aligned}
\end{equation}
Dividing this derivative with the density in Eq.~(\ref{Weq}) gives
the gradient estimator in Eq.~(\ref{Weq}).

\paragraph{Experimental results for W-distribution:}

\begin{figure*}
        \centering
	\begin{subfigure}{.49\textwidth}
		\includegraphics[width=\textwidth]{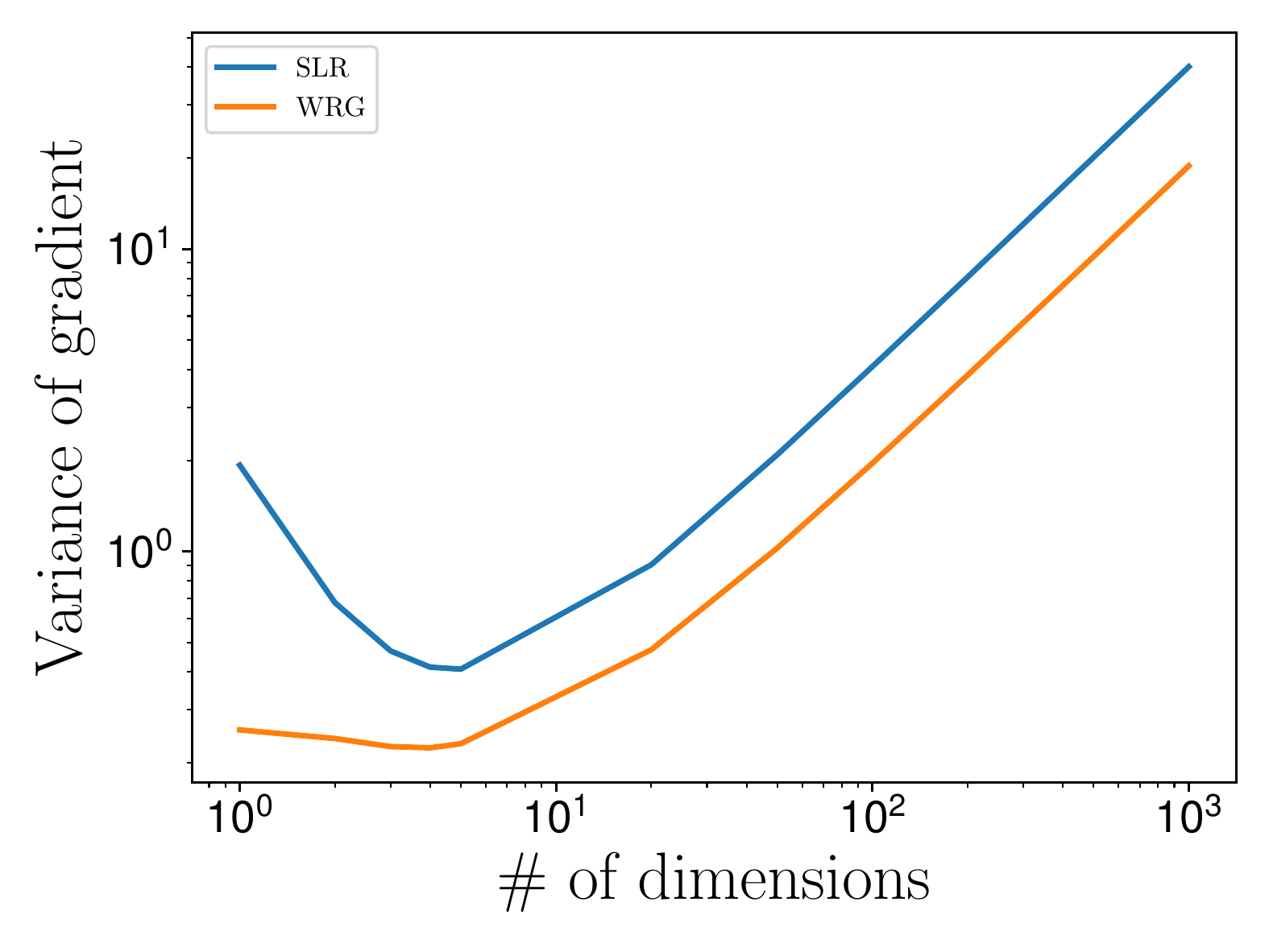}
		\caption{Deterministic, $\bfv{a} = \bfv{1}$}
          \label{detquadW1}
        \end{subfigure}
%%%%%%%%%%%%%%
	\begin{subfigure}{.49\textwidth}
		\includegraphics[width=\textwidth]{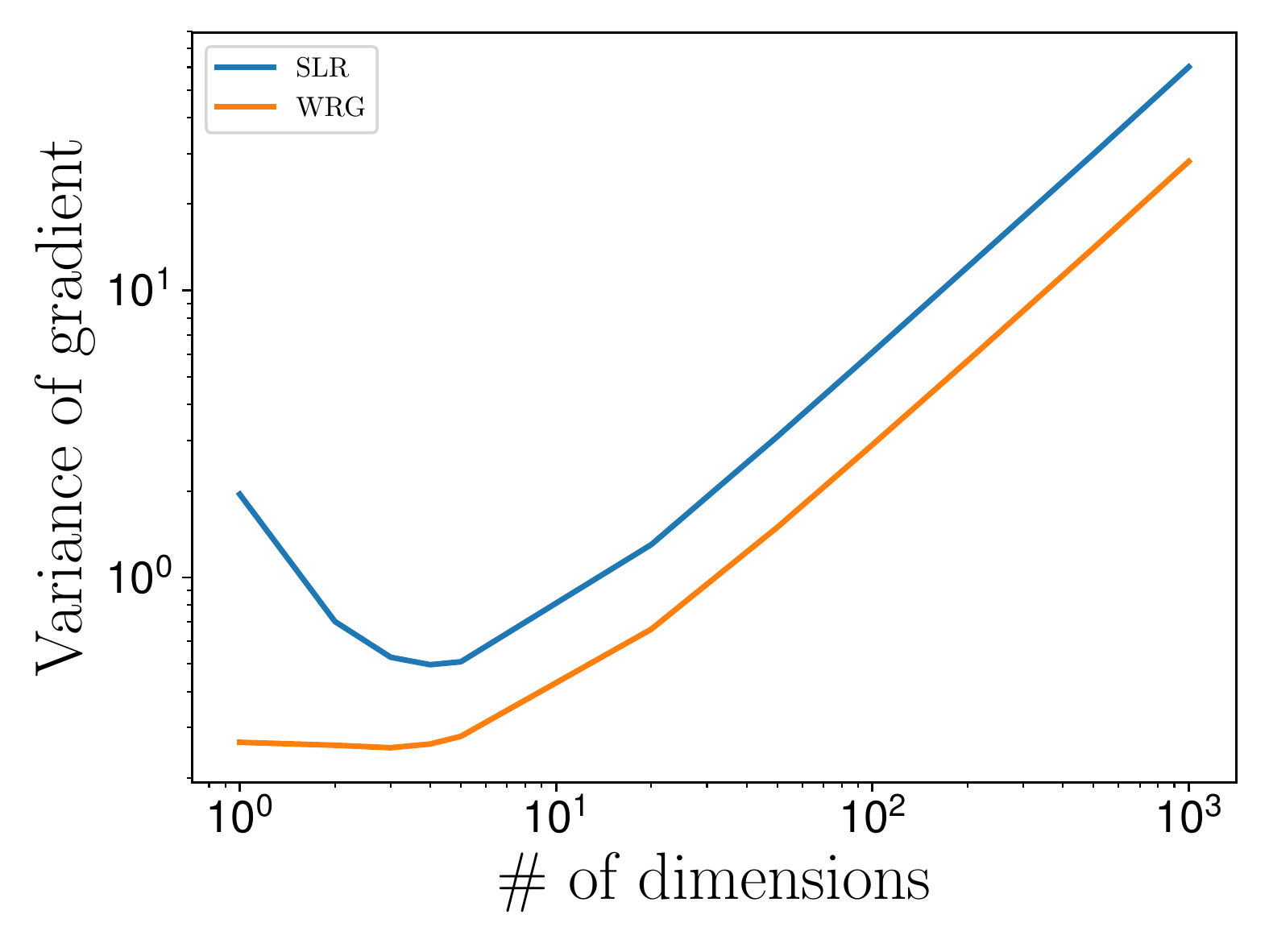}
		\caption{Noisy, $\bfv{a} = \bfv{1}$}
          \label{noisyquadW1}
	\end{subfigure}
%%%%%%%%%%%%%%
	\begin{subfigure}{.49\textwidth}
		\includegraphics[width=\textwidth]{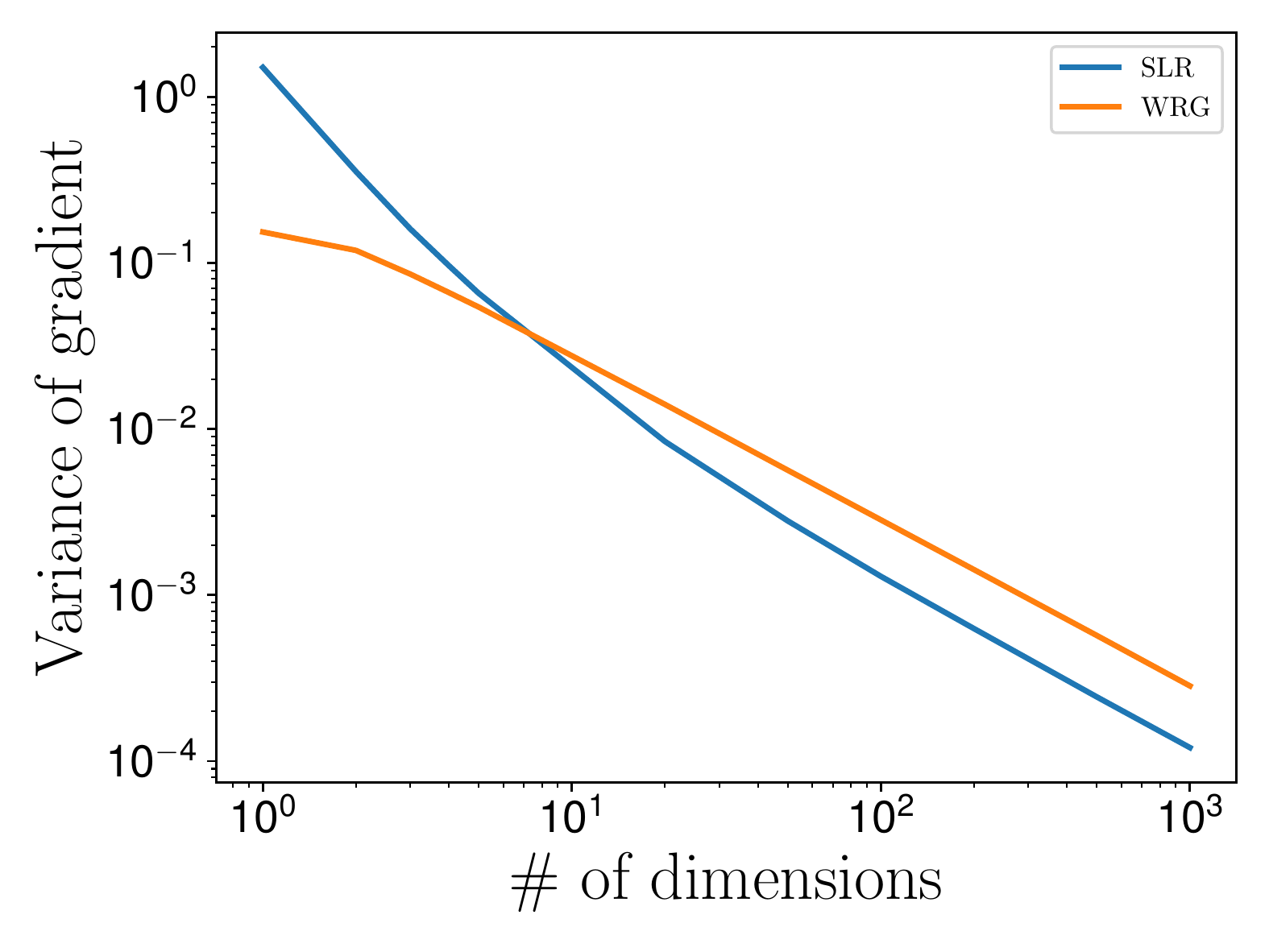}
		\caption{Deterministic, $\bfv{a} = \bfv{0}$}
          \label{detquadW2}
        \end{subfigure}
%%%%%%%%%%%%%%
	\begin{subfigure}{.49\textwidth}
		\includegraphics[width=\textwidth]{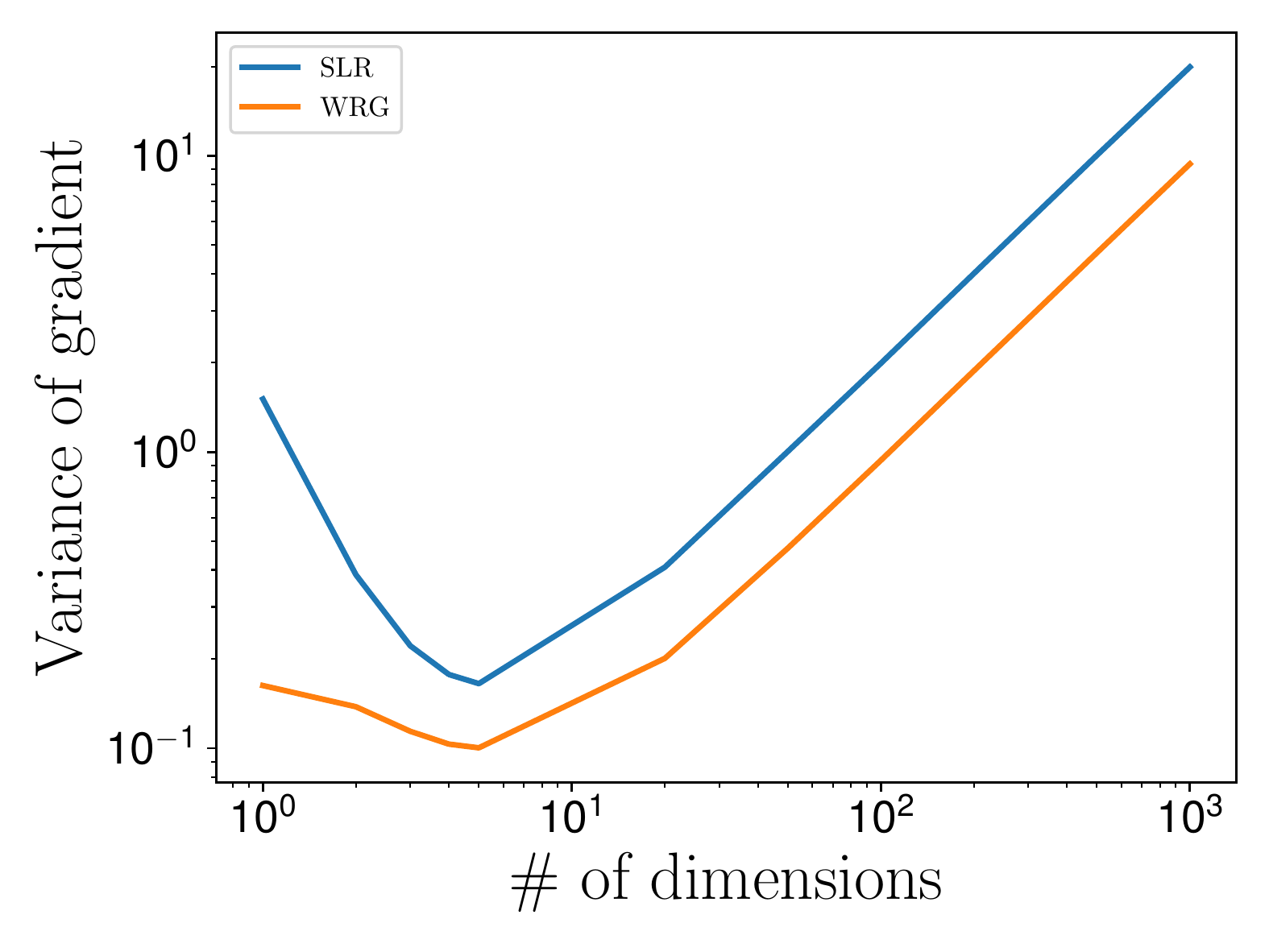}
		\caption{Noisy, $\bfv{a} = \bfv{0}$}
          \label{noisyquadW2}
	\end{subfigure}
        \caption{The confidence
          intervals correspond to one standard deviation of the
          estimate.}
          \label{quadWexps}
\end{figure*}

We performed experiments similar to the experiment in the main section
of the article by comparing the standard LR gradient with the
estimator by sampling from the W-distribution. The setup was
such that $\phi(\bfv{x})$ is a quadratic
$(\bfv{x}-\bfv{a})^TQ(\bfv{x}-\bfv{a})$, where 
$Q = \textup{ones}(D,D)/D^2$ is a matrix of ones, which is scaled,
such that $\phi(\bfv{x})$ remains constant at $\bfv{x}=\bfv{0}$.
We considered two options for $\bfv{a}$: $\bfv{a} = \bfv{1}$ or
$\bfv{a} = \bfv{0}$. We
evaluate a deterministic case, as well as a case when Gaussian noise
$\sigma_n^2=1$ is added on $\phi$. We vary the dimension between
1--1000, and plot the variance of the gradient estimators: SLR---LR
gradient with a Gaussian $p(x)$ and estimating the gradient w.r.t. $\sigma$;
WRG---slice ratio gradient with a
Gaussian $p(x)$, and using the W-distribution to importance sample
and estimate the gradient w.r.t. $\sigma$. We used antithetic
sampling, so that the effect of any baseline could be ignored. The
gradient was estimated by averaging 100 samples, and this was repeated
for a large number of times to estimate the variance of the gradient
estimator. Bootstrapping was used to obtain confidence intervals. The
results (Fig.~\ref{quadWexps}) confirm that the W-distribution increases
the accuracy.

\subsection{Multidimensional Gaussian Slice ratio gradient}
\label{multdim}
In multiple dimensions the optimality equation in
Eq.~(\ref{eq:optdist}) is still valid, but the method to derive the
normalized distribution and sampling method have to be modified. For
simplicity, we consider the case of optimal sampling for the
derivative w.r.t. $\mu$ for a spherical Gaussian. Motivated from the
 derivation for a single dimension, consider a method which would sample a unit
vector on a sphere for a direction $\hat{\bfv{r}}$, as well as a
height $h$, then invert the distribution s.t.
$\bfv{x} = p^{-1}(h,\hat{\bfv{r}})$, where $p^{-1}$ is a function
s.t. $\p{\bfv{x}}=h$ and $\bfv{x} = r\hat{\bfv{r}}$, i.e., it picks
$\bfv{x}$ in the direction $\hat{\bfv{r}}$, which gives the desired
probability density.  The conversion from the $h$-coordinate to the
$\bfv{x}$-coordinate would still give the desired
$\left|\deriv{\p{\bfv{x}}}{\bfv{x}}\right|$ term; however, due to the
change in the surface area as the radius $r$ is increased, there is an
additional factor $r^{-(D-1)}$, where $D$ is the dimensionality. In
other words, the sampling method has to be modified to cancel out this
new factor, and the required distribution must have the property:
$q(\bfv{x}) \propto r^{D-1}\left|\deriv{\p{\bfv{x}}}{\bfv{x}}\right|$.
For a Gaussian base distribution we get
$q(\bfv{x}) \propto r^{D}\exp(-\frac{r^2}{2\sigma^2})$. The required
distribution is the chi distribution:
\begin{equation}
\label{eq:chi}
  q(z;k) = \frac{1}{2^{(k/2-1)}\Gamma(k/2)}z^{k-1}\exp(-\frac{z^2}{2})
  ~~~\textup{where}~~~r = \sigma z ~~~\textup{and}~~~ k=D+1.
\end{equation}

In fact, the Rayleigh distribution is a special case of this
distribution for $D=1$, and the Maxwell-Boltzmann distribution is the
case for $D=2$. Note that if one performs this sampling procedure, but
while using $\tilde{D} = D-1$, then the sample comes exactly from the
original Gaussian distribution $\p{\bfv{x};\mu,\Sigma}$. This remark
highlights that there are diminishing returns to changing the sampling
distribution as the dimensionality of the space is increased, because
the optimal sampling distribution tends to the original Gaussian
distribution. 

\paragraph{Derivation of the directional ratio gradient estimator
  (DRG):} The chi distribution $q(z;k)$ with k degrees of freedom is a
distribution, s.t. $z$ is distributed according to the random variable
$z = \sqrt{\sum_{i=1}^kx_i}$, where $x_i$ are distributed according to
a Gaussian distribution with mean 0 and standard deviation 1. In other
words, $z$ is distributed according to the length of the distance from
the origin, when sampling from a spherical Gaussian with $k$
dimensions, and to sample from a Gaussian distribution, it suffices to
sample a direction $\hat{\bfv{r}}$ on the unit sphere, then sample the
distance $z$ according to the chi distribution, and add a factor
$\sigma z$ to correct for the scaling from the variance parameter. We
can write the probability density of a Gaussian in spherical
coordinates as
\begin{equation}
  \mathcal{N}\left(\bfv{x};0,\sigma^2I\right) =
  \frac{1}{A}q(z;D),
\end{equation}
where $A$ is the area of a D-dimensional hypersphere at radius
$\sigma z$ given by
$A = \frac{2\pi^{D/2}}{\Gamma\left(\frac{1}{2}D\right)}(\sigma
z)^{D-1}$, and $\bfv{x} = \sigma z\hat{\bfv{r}}$, where
$\hat{\bfv{r}}$ is a vector sampled on the unit sphere. In cartesian
coordinates, the gradient w.r.t. $\bfv{x}$ can be written as
\begin{equation}
  \deriv{\p{\bfv{x};0,\Sigma}}{\bfv{x}} =
  \frac{1}{(2\pi)^{D/2}|\Sigma|^{1/2}}\exp\left(-\bfv{x}^T\Sigma^{-1}\bfv{x}\right)
  \times\left(-\Sigma^{-1}\bfv{x}\right) = -\Sigma^{-1}\bfv{x}p(x;0,\Sigma).
\end{equation}
Translating this result to spherical coordinates, we have
\begin{equation}
\label{eq:sphergderiv}
  \deriv{\p{\bfv{x};0,\Sigma}}{\bfv{x}} =
  -\frac{z\hat{\bfv{r}}}{\sigma}\frac{1}{A}q(z;D),
\end{equation}
and
\begin{equation}
\label{eq:dpdxz}
  \left|\deriv{\p{\bfv{x};0,\Sigma}}{\bfv{x}}\right| = \frac{z}{\sigma}\frac{1}{A}q(z;D).
\end{equation}
If one applies the same directional sampling scheme, but instead of
sampling from $q(z;D)$, one samples from a distribution proportional
to $zq(z;D)$, one would be sampling from the desired distribution. By
inspecting Eq.~(\ref{eq:chi}), it is clear that increasing the degrees
of freedom by one adds the additional $z$ factor, so the optimal
importance sampling distribution is
\begin{equation}
\label{eq:drg}
q(\bfv{x}) = \frac{1}{A}q(z;D+1),
\end{equation}
where $\bfv{x} = \sigma z\hat{\bfv{r}}$. To obtain the gradient
estimator, divide $\deriv{p}{\mu} = -\deriv{p}{x}$ given in
Eq.~(\ref{eq:sphergderiv}), with $q(\bfv{x})$ in Eq.~(\ref{eq:drg}):
\begin{equation}
\label{eq:drggrad}
  \left.\frac{z\hat{\bfv{r}}}{\sigma}\frac{1}{A}q(z;D)\middle/
    \frac{1}{A}q(z;D+1)\right. = \frac{\hat{\bfv{r}}}{\sigma}
  \frac{zq(z;D)}{q(z;D+1)} = \frac{\hat{\bfv{r}}}{\sigma}
  \frac{2^{1/2}\Gamma\left((D+1)/2\right)}{\Gamma\left(D/2\right)}
\end{equation}

\paragraph{Derivation of the directional ratio gradient estimator while
  assuming a linear $\phi$ (DLRG):}
The above derivation made our standard assumption that $\phi$ is ignored.
Another option is to assume that $\phi$ varies linearly. The derivation is
easily modified. From Eq.~(\ref{eq:dpdxz}), we saw that it was necessary
to sample from a distribution proportional to $zq(z)$. In the
new derivation, it will be necessary to sample proportionally to
$zq(z)\phi(z;D)$, which when $\phi$ is linear, is equivalent to sampling
proportionally to $z^2q(z;D)$. Based on the same argument as in
Eq.~(\ref{eq:drg}), the necessary sampling can be done by increasing
the degrees of freedom by 2, i.e., one must sample from
\begin{equation}
\label{eq:dlrg}
q(\bfv{x}) = \frac{1}{A}q(z;D+2).
\end{equation}
Similarly to Eq.~(\ref{eq:drggrad}), the gradient estimator can be
derived:
\begin{equation}
\label{eq:dlrggrad}
  \left.\frac{z\hat{\bfv{r}}}{\sigma}\frac{1}{A}q(z;D)\middle/
    \frac{1}{A}q(z;D+2)\right. = \frac{\hat{\bfv{r}}}{\sigma}
  \frac{zq(z;D)}{q(z;D+2)} = \frac{2\hat{\bfv{r}}}{z\sigma}
  \frac{\Gamma\left((D+2)/2\right)}{\Gamma\left(D/2\right)} =
  \frac{D\hat{\bfv{r}}}{z\sigma},
\end{equation}
where the last line follows from the property $\Gamma(n+1) = n\Gamma(n)$.

\subsection{Sufficient conditions for an unbiased gradient
  estimator while ignoring importance weights
from other dimensions}
\label{suffcon}
First we consider functions of the form
$\phi(\bfv{x}) = \sum_{i=1}^D\phi_i(x_i)$, and show that ignoring the
importance weights from dimension $j\neq i$ for the derivative
w.r.t. $\theta_i$, still gives an unbiased gradient estimator.
Note that $\expectw{x_i\sim q(x_i)}
{\frac{\p{x_i}}{q(x_i)}
  \deriv{\log\p{x_i;\theta_i}}{\theta_i}
  \expectw{x_j\sim q(x_j)}{\phi_j(x_j)}
} =
\expectw{x_i\sim p(x_i)}
{
  \deriv{\log\p{x_i;\theta_i}}{\theta_i}
  \expectw{x_j\sim q(x_j)}{\phi_j(x_j)}
}
= 0$, because
$\expectw{x_i\sim p(x_i)}
{
  \deriv{\log\p{x_i;\theta_i}}{\theta_i}
  Y} = \deriv{}{\theta_i}\expect{Y}=0$, for $Y$ statistically independent
from $x_i$. This result means that if $\phi$ has a structure, such
that different dimensions affect $\phi$ independently, then
the gradient estimator will still be unbiased.

Next we show that even if the dimensions are not independent, in some
cases the gradient estimator is unbiased. Notably, for a quadratic
function $\phi(\bfv{x}) =\bfv{a}^T\bfv{x} + \bfv{x}^TQ\bfv{x} + c$, the gradient
estimator will be unbiased. First note that the diagonal terms in the
quadratic function are independent, so the gradient of that portion of
the cost will be unbiased based on the previous example.  Next
consider the off-diagonal terms of $\bfv{x}^TQ\bfv{x}$, which are 
$x_iQ_{ij}x_j$. Note that the distributions we considered, namely the
B, W, L and Beta distributions were all symmetric about the mean value
$\mu_j$. Therefore
$\expectw{x_j\sim q(x_j)}{Q_{ij}x_j} = \expectw{x_j\sim
  p(x_j)}{Q_{ij}x_j}$, and the derivative
$\deriv{}{\theta}
\expectw{x_i\sim p(x_i)}
{
  x_i\expectw{x_j\sim q(x_j)}{
    \frac{p(x_j)}{q(x_j)}Q_{ij}x_j}
}
$
remains unchanged
even if one ignores the $p(x_j)/q(x_j)$ importance
weights. This result implies that if the variance of the distribution
$\sigma^2$ is small, such that $\phi$ is roughly quadratic in the range
of the sampling distribution, then the gradient estimator will remain
roughly unbiased.

\section{Additional justifications for approximations in
the derivations}

\subsection{Ignoring importance weights in multidimensional
slice ratio sampling}
\label{sec:multjust}

In Sec.~\ref{sliceratio} for the multidimensional case we considered
factorized distributions
$\p{\bfv{x};\theta} = \prod_i\pind{i}{x_i;\theta_i}$, and for
estimating the gradient w.r.t.  $\theta_i$, we chose to ignore the
importance weights from the other dimensions $j \neq i$. We justified
the omission by noting that the unbiased gradient estimator would be
given by
\begin{equation}
\frac{\pind{\backslash
          i}{\bfv{x}_{\backslash i}}\pind{i}{x_i}} {q_{\backslash
          i}(\bfv{x}_{\backslash i})q_i(x_i)}\deriv{\log
        \pind{i}{x_i}}{\theta_i}\phi(\bfv{x}),
\end{equation}
and that the variance would grow exponentially as the dimension increases,
because of the growth of the variance of the $\frac{\pind{\backslash
i}{\bfv{x}_{\backslash i}}} {q_{\backslash
i}(\bfv{x}_{\backslash i})}$ term. The assumption of
factorized distributions may appear restrictive; however, note that this
is the most common scenario in practice, and finding a solution in this
setting is important. Moreover, note that by making the factorization
assumption, we ended up with a worst case scenario, where as the dimension
increases, the optimal unbiased importance sampling distribution will
tend to the original distribution, thus showing that no gains are possible
without adding in bias. Replacing the importance weights with their
expected value is not just a convenience, but a necessity. If the distribution
does not factorize, then such an omission may not be necessary, and
good unbiased importance sampling distributions may exist, but our methods
would not be directly applicable, and this is a topic for future work.

\subsection{Omission of $\phi$ in optimality of slice ratio sampling derivation theoretical reasons}
\label{sec:phiom}

In the derivation of the slice ratio gradients, the optimization of
the variance of $\deriv{\p{x;\theta}}{\theta}\phi(x)/q(x)$ was
replaced with optimizing the variance of
$\deriv{\p{x;\theta}}{\theta}/q(x)$. Here we explain the various
reasons, which justify this omission, and show that in most realistic
settings it is almost exactly the correct thing to do. We introduce
three realistic settings to which this omission corresponds: 1. the
estimation of $\phi$ is very noisy, 2. $\phi$ is high dimensional,
3. $\phi$ has high frequency variations at a length scale smaller than
the range of the sampling distribution. In addition, note that another
reasonable assumption might be to assume that $\phi(x)$ is linear, but
the L-distribution (App.~\ref{sliceintegral}) turns out to be optimal
in this setting (in low dimensions).

\paragraph{$\phi$ is very noisy:} If $\phi$ is noise uncorrelated
with $x$, then $\variance{\deriv{\p{x;\theta}}{\theta}\phi(x)/q(x)} =
\variance{\deriv{\p{x;\theta}}{\theta}/q(x)}\variance{\phi(x)}$, and
one can ignore $\phi$ in the optimization. The same reasoning holds if
$\phi(x) = \hat\phi(x) + \epsilon_n$, where $\epsilon_n$ is random noise
with magnitude much larger than the variation of $\phi(x)$.

\paragraph{$\phi$ is high dimensional:} Consider the independent
multidimensional sampling scenario justified in App.~\ref{sec:multjust},
and estimating the variance of the gradient of one dimension $i$. The
general gradient estimator is given by
\begin{equation}
\begin{aligned}
&\deriv{}{\theta_i}\expectw{\bfv{x}\sim\p{\bfv{x}}}{\phi(\bfv{x})}
    = 
    \expectw{\bfv{x}\sim q(\bfv{x})}
    {\frac{\pind{\backslash
          i}{\bfv{x}_{\backslash i}}\pind{i}{x_i}} {q_{\backslash
          i}(\bfv{x}_{\backslash i})
        q_i(x_i)}\deriv{\log
        \pind{i}{x_i}}{\theta_i}\phi(\bfv{x})}, ~~~~~~~\textup{where }p_{\backslash i}
    \textup{ is } \prod_{j\neq i}\pind{j}{x_j;\theta_j}.
\end{aligned}
\end{equation}
In the independent sampling case, we justified that
$\frac{\pind{\backslash i}{\bfv{x}_{\backslash i}}}{q_{\backslash
    i}(\bfv{x}_{\backslash i})}$ should be ignored if one hopes to
make any gains in terms of variance reduction.\footnote{Potentially
  other variance reduction techniques besides completely ignoring the
  weights may also work, e.g., clipping the weights, but the analysis
  regarding omitting $\phi$ is not affected.} We are left with
estimating the variance of
$\frac{\pind{i}{x_i}}{q_i(x_i)}\deriv{\log
  \pind{i}{x_i}}{\theta_i}\phi(\bfv{x})$. Note that $\phi(\bfv{x})$
still contains all dimensions other than $i$, i.e. $x_j,$ where
$j\neq i$ still matter; however, they are statistically independent of
the gradient estimator, and thus the variation caused by $x_j$ acts as
noise on the gradient signal. We call this, the {\it sampling
  interference noise}. If one assumes that the dimensionality is $D$,
and that the variation of $\phi(x)$ is roughly the same in each
dimension, then roughly a fraction $(D-1)/D$ of $\phi$ can be
considered as noise for each gradient estimator. Thus, as the
dimension increases, the variation in $\phi(\bfv{x})$ rapidly
approaches equivalence to random noise, and rejecting the noise will
be most important for reducing gradient variance.

\paragraph{$\phi$ has high frequency components:} Consider
$\phi(x) = \hat{\phi}(x) + a\sin(\omega x)$. If $\omega$ is large
compared to the sampling range, then $\sin(\omega x)$ is almost
statistically independent to $x$, and can be viewed as noise. Such
high frequency components occur when applying LR gradients to chaotic
systems \citep{pipps}, and correspond to the situation when LR vastly
outperforms RP. Thus, reducing the variance of LR gradients in this
scenario is important.

\subsection{Additional experiments showing downsides of
alternative approaches}

\begin{figure*}[!t]
        \centering
	\begin{subfigure}{.49\textwidth}
		\includegraphics[width=\textwidth]{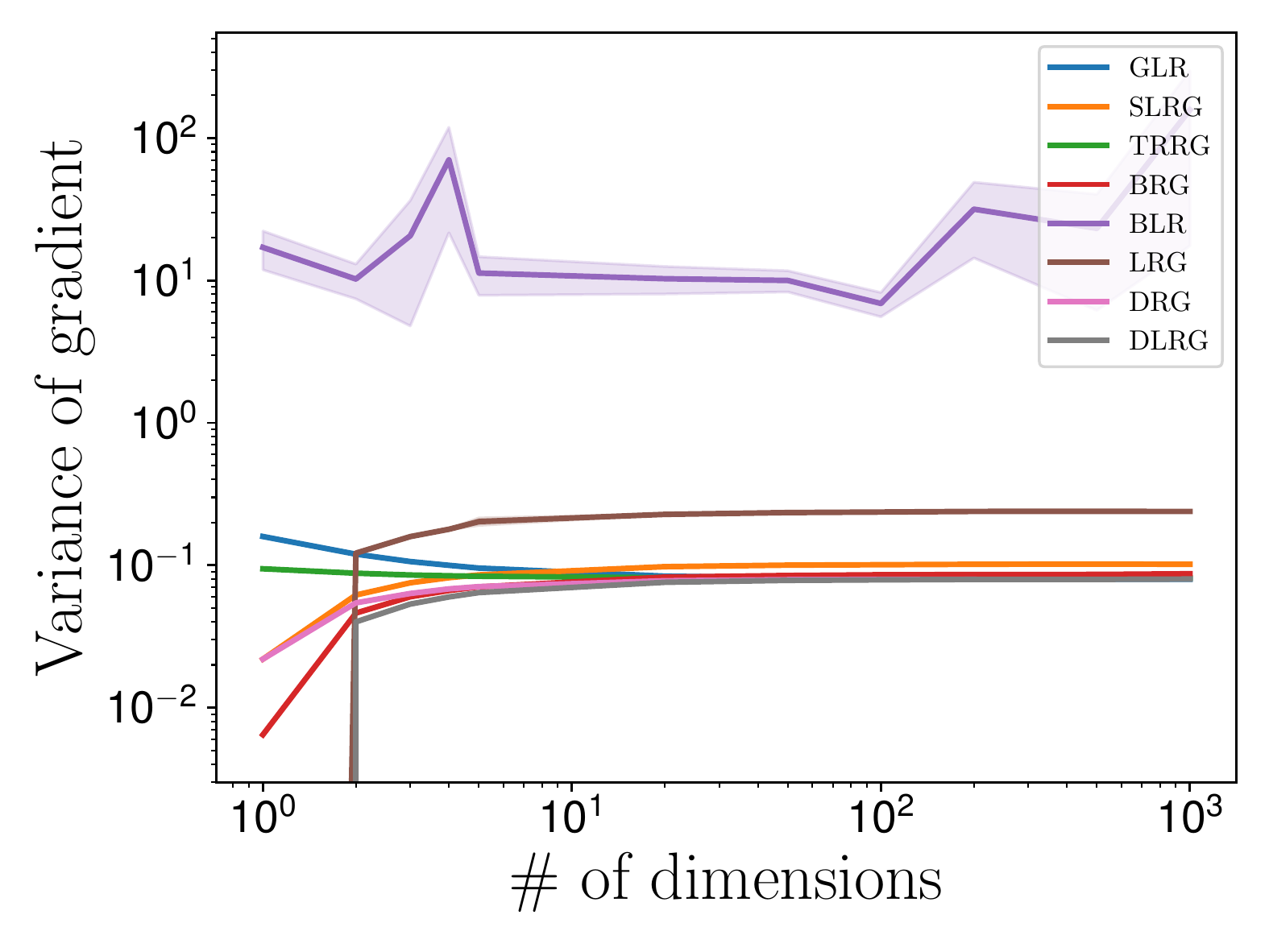}
		\caption{Deterministic}
          \label{detquadnew}
        \end{subfigure}
%%%%%%%%%%%%%%
	\begin{subfigure}{.49\textwidth}
		\includegraphics[width=\textwidth]{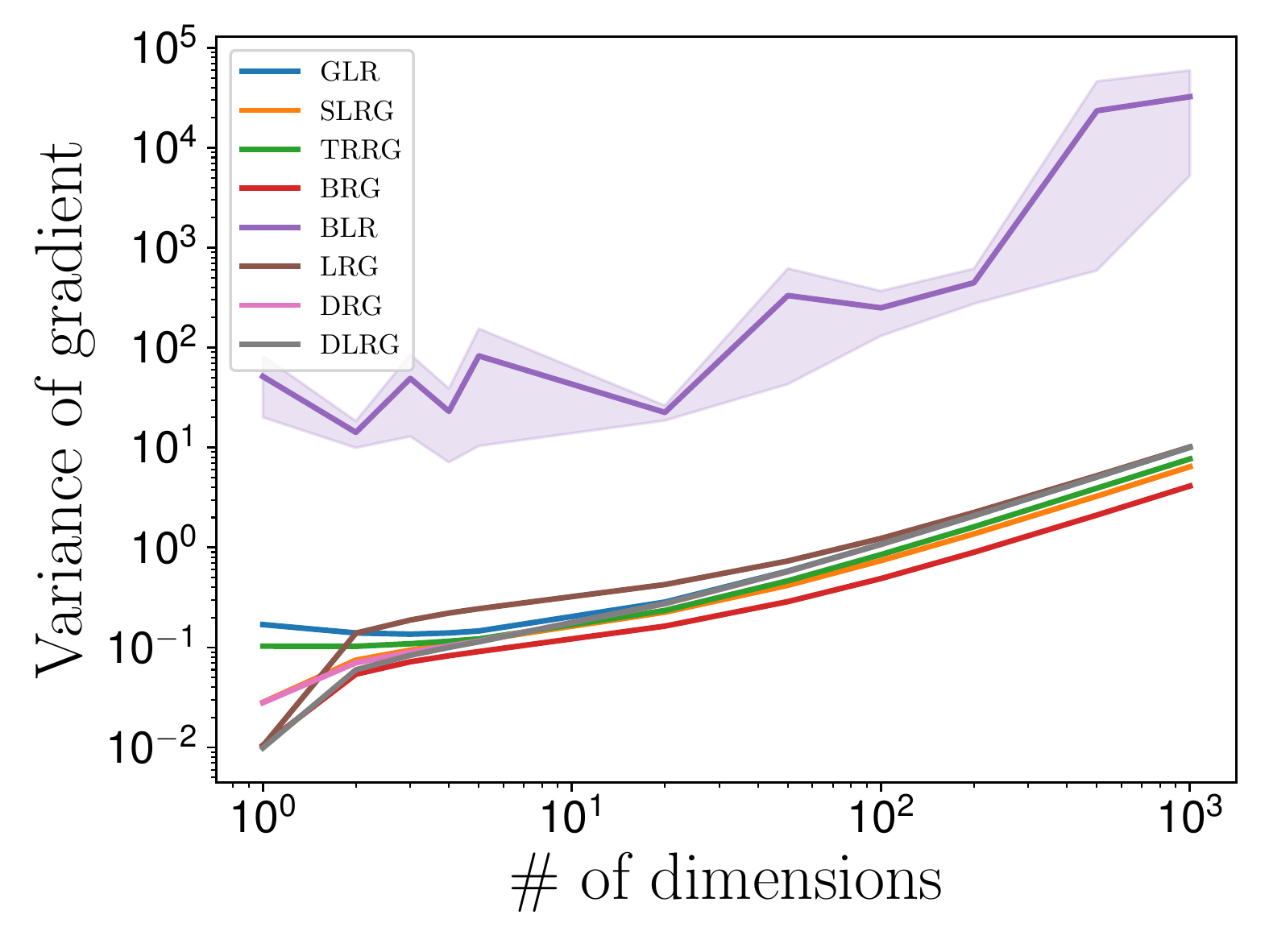}
		\caption{Noisy}
          \label{noisyquadnew}
	\end{subfigure}
%%%%%%%%%%%%%%
	\caption{yadayada. The confidence intervals correspond to one
          standard deviation of the estimate.}
          \label{quadexpsnew}
\end{figure*}

In Fig.~\ref{quadexpsnew} we show experiments evaluating alternative
optimal gradient estimators based on different assumptions, and
explain that our approach in the main paper is better.
We evaluated importance sampling based on the L-distribution (LRG),
as well as the optimal importance sampling distribution in multiple
dimensions without ignoring the importance weights from the other
dimensions. DRG stands for directional ratio gradient, which omits
$\phi$ in the derivation, but considers importance weights from all
dimensions. DLRG assumes $\phi$ is roughly linear. The other results are
the same as in the main section of the article: GLR---LR
gradient with a Gaussian $p(x)$; SLRG---slice ratio gradient with a
Gaussian $p(x)$; TRRG---truncated ratio gradient with $c=0.5$;
BRG---slice ratio gradient with a Beta $p(x)$, and $\alpha=1.5$,
plotted in Fig.~\ref{betaplot}; BLR---LR gradient with a Beta $p(x)$,
and $\alpha=1.5$.

\paragraph{Results:} The alternative methods (LRG, DRG, DLRG) converge
to GLR in the noisy setting for high dimensions, and show no gain in
terms of variance reduction, whereas our proposed methods, SLRG and
TRRG are able to show some advantage. In the deterministic case, at
high dimensions LRG has higher variance than the other methods,
because it has a larger sampling variance, which increases the {\it
  sampling interference noise} from the other dimensions, explained in
Sec.~\ref{sec:phiom}. A point which deserves discussion is that in the
1-dimensional deterministic case, LRG and DLRG show extremely low
variance. The gradient estimator in this situation is given by
$\frac{\textup{sgn}(x-\mu)}{\sigma\sqrt{-2\log(\epsilon_h) +
    \epsilon_x^2}}\phi(x) =
\frac{\textup{sgn}(x-\mu)}{x}\phi(x)$. Note that in the antithetic
sampling setting, this estimator just becomes a finite difference
$(\phi(x^+) - \phi(x^-))/(2\Delta x)$, where the 2 comes from
averaging two samples, and if $\phi$ is linear, then the gradient
estimator would be exact with just one sampled pair (unlike the
standard finite difference estimator, this estimator would be unbiased
for non-linear $\phi$ as well). In the experiment, the curvature of
$\phi$ was quite low, so LRG and DLRG gave extremely low variance in
the 1-dimensional setting, but this advantage does not hold up, when
the dimensionality is increased. In conclusion, the approximations
we made are well justified.

\section{Truncated ratio gradient derivations}
\label{truncderivapp}

Recall that the truncated ratio gradient probability density function, sampling method and
gradient estimator are given by the result below:

\begin{equation}
  \begin{aligned}
  \label{trunkappendix}
  \pind{tr}{x;c,\mu,\sigma} &= \frac{\exp(-\frac{c^2}{2})}{1-\Phi(c)}
  \frac{1}{\sigma2\sqrt{2\pi}}\frac{|x-\mu|}{\sqrt{(x-\mu)^2+\sigma^2c^2}}
  \exp(-\frac{(x-\mu)^2}{2\sigma^2}) \\ &~~~\textup{where}~ \Phi(c) \textup{ is the cdf
  of a unit normal distribution},\\
  x &=
  \mu \pm \sigma\sqrt{\epsilon_c^2 - c^2} ~~~\textup{where}~~~
  \epsilon_c \sim \textup{truncG}(c,\infty)\\
  &~~~\textup{and}~ \textup{truncG}(a,b)
  \textup{ is the unit normal truncated between } a \textup{ and } b,\\
\deriv{}{\mu}\expectw{x\sim\pind{tr}{x}}{\phi(x)} &=
\expectw{x\sim q(x)}{\textup{sgn}(x-\mu)
    \frac{2\epsilon_c}{\sigma}\frac{1-\Phi(c)}{\exp(-\frac{c^2}{2})}\phi(x)}.
\end{aligned}
\end{equation}

The pdf $\pind{tr}{x;c,\mu,\sigma}$ satisfies the optimality
Eq.~(\ref{varconstoptimality}), therefore, as long as it is a proper
probability density, it is correct. We will show that the proposed
sampling method corresponds to this pdf.

Without loss of generality, let $\mu=0$. The pdf of $\epsilon_c$ is
given by

\begin{equation}
  \label{epscdist}
  \p{\epsilon_c} = \frac{1}{\sqrt{2\pi}}\exp\left(-\frac{\epsilon_c^2}{2}\right)
  \frac{1}{1 - \Phi(c)}, \textup{between } \epsilon_c\in[c,\infty].
\end{equation}

Perform a change of coordinates from $\epsilon_c$ to $x$ and account for the stretching
due to the Jacobian:

\begin{equation}
  x = \sigma\sqrt{\epsilon_c^2 - c^2} \Rightarrow \textup{d}x =
  \sigma\frac{\epsilon_c}{\sqrt{\epsilon_c^2 - c^2}}\textup{d}\epsilon_c,
\end{equation}

note that $\sigma\epsilon_c = \sqrt{x^2+\sigma^2c^2}$, so

\begin{equation}
  \frac{x/\sigma}{\sqrt{x^2+\sigma^2c^2}}\textup{d}x = \textup{d}\epsilon_c,
\end{equation}

therefore

\begin{equation}
\begin{aligned}
  \p{\epsilon_c}\textup{d}\epsilon_c &= \frac{1}{\sqrt{2\pi}}
  \exp\left(-\frac{\epsilon_c^2}{2}\right)
  \frac{1}{1 - \Phi(c)} \frac{x/\sigma}{\sqrt{x^2+\sigma^2c^2}}\textup{d}x\\
  &= \frac{1}{\sqrt{2\pi}}\exp\left(-\frac{x^2}{2\sigma^2}+\frac{c^2}{2}\right)
  \frac{1}{1 - \Phi(c)} \frac{x/\sigma}{\sqrt{x^2+\sigma^2c^2}}\textup{d}x\\
  &= \frac{\exp\left(-\frac{c^2}{2}\right)}{1-\Phi(c)}\frac{1}{\sigma\sqrt{2\pi}}
  \frac{x}{\sqrt{x^2 +\sigma^2c^2}}\exp\left(-\frac{x^2}{2\sigma^2}\right)\textup{d}x.
\end{aligned}
\end{equation}

This result is the desired probability distribution on the half-plane $[0,\infty]$.
It is a normalized pdf by construction. Symmetrizing
the distribution about 0, and shifting by a mean parameter $\mu$ gives the desired result.
The gradient estimator is easily derived by $\nicefrac{\deriv{p}{\theta}}{q}$.

\paragraph{Variance and gradient accuracy derivations:}

The variance is most easily derived by working with the distribution on $\epsilon_c$ in
Eq.~(\ref{epscdist}). Note that if we symmetrize the distribution about 0, then the
mean will be 0, and the variance can be estimated as the expectation of $x^2$ when sampling
from half the distribution:

\begin{equation}
  \variance{x} = \int x^2\p{\epsilon_c}\textup{d}\epsilon_c =
  \int \sigma^2(\epsilon_c^2-c^2)\p{\epsilon_c}\textup{d}\epsilon_c
  = \int \sigma^2\epsilon_c^2\p{\epsilon_c}\textup{d}\epsilon_c - \sigma^2c^2.
\end{equation}

So, we just need to find
$\expect{\epsilon_c^2} = \variance{\epsilon_c} +
\expect{\epsilon_c}^2$. Denote $\mathcal{N}(x)$ is the unit variance
Gaussian distribution, and $\Phi(x)$ is the cdf of the unit variance
Gaussian, then the mean and variance of the 0 mean truncated Gaussian
between $[c, \infty]$ can be written as
$\expect{\epsilon_c} = \frac{\mathcal{N}(c)}{1-\Phi(c)}$ and
$\variance{\epsilon_c} = 1 + \frac{c\mathcal{N}(c)}{1 - \Phi(c)} -
\left(\frac{\mathcal{N}(c)}{1 - \Phi(c)}\right)^2$. Combining these two results:

\begin{equation}
\expect{\epsilon_c^2} = 1 + \frac{c\mathcal{N}(c)}{1 - \Phi(c)}.
\end{equation}

Hence, the variance is

\begin{equation}
v(c) = \variance{x} = \sigma^2\left(1 + \frac{c\mathcal{N}(c; 0, 1)}{1 - \Phi(c)} - c^2\right).
\end{equation}

Next, we derive the variance of the gradient term $\deriv{p}{\mu}/q$. Note that
this term is given in Eq.~(\ref{trunkappendix}) as $\textup{sgn}(x-\mu)
\frac{2\epsilon_c}{\sigma}\frac{1-\Phi(c)}{\exp(-\frac{c^2}{2})}$, so the variance is

\begin{equation}
  \begin{aligned}
  \variance{\deriv{p}{\mu}/q} =
  \expect{\left(\frac{2\epsilon_c}{\sigma}\frac{1-\Phi(c)}{\exp(-\frac{c^2}{2})}\right)^2}
  &= \expect{\epsilon_c^2}\left(\frac{2}{\sigma}\frac{\left(1-\Phi(c)\right)}
    {\exp(-\frac{c^2}{2})}\right)^2 \\&=
  \left(1 + \frac{c\mathcal{N}(c;0,1)}{1 - \Phi(c)}\right)
  \frac{4}{\sigma^2}\frac{\left(1-\Phi(c)\right)^2}
  {\exp(-c^2)}.
  \end{aligned}
\end{equation}

Finally, note that the gradient accuracy $t(c)$ is defined as
$1/\variance{\deriv{p}{\mu}/q}$.

\section{Evolution strategies in reinforcement learning}
\label{evolstrategies}

\begin{figure*}[!t]
        \centering
	\begin{subfigure}{.49\textwidth}
		\includegraphics[width=\textwidth]{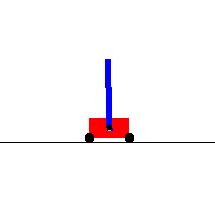}
		\caption{Cart-pole swing-up}
          \label{cartpole}
        \end{subfigure}
%%%%%%%%%%%%%%
	\begin{subfigure}{.49\textwidth}
		\includegraphics[width=\textwidth]{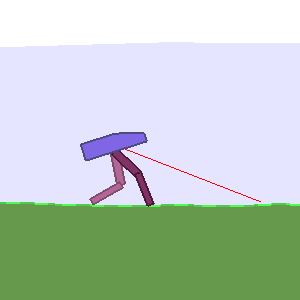}
		\caption{Biped walker}
          \label{biped}
	\end{subfigure}
%%%%%%%%%%%%%%
	\caption{Environments used in evolution strategies experiments.}
          \label{envpics}
\end{figure*}

Evolution strategies are a technique based on sampling in the
parameter space of a problem $\p{\bfv{w};\theta}$, and applying LR
gradients to optimize the objective
$\expectw{\p{\bfv{w};\theta}}{\phi(w)}$.  For example $\bfv{w}$ may be
the parameters of a neural network policy in reinforcement learning,
and the objective is to find the distribution $\p{\bfv{w};\theta}$
over the parameters $\bfv{w}$, which gives the behavior with the
largest expected reward.  In this case, $\phi(\bfv{w})$ would be the
return function for a particular parameter set $\bfv{w}$. One would
first sample parameters $\bfv{w}$, these would be kept fixed for one
episode of the agent's behavior, the behavior would be evaluated based
on a reward function, and the sum of the reward would be returned to
the algorithm as $\phi(\bfv{w})$. LR gradients can be used to evaluate
$\deriv{}{\theta}\expectw{\bfv{w}\sim\p{\bfv{w};\theta}}{\phi(\bfv{w})}$,
and the objective can be optimized directly using gradient ascent. We
implemented our new importance sampling schemes into David Ha's
Evolution Strategies code available from
https://github.com/hardmaru/estool\citep{ha2017evolving} (note that our methods are
not available from the link yet), and tested our methods on cart-pole
swing-up and biped walker tasks illustrated in Fig.~\ref{envpics}.

\subsection{Experiments}

In all experiments we used spherical Gaussian base distributions
$\p{\bfv{w};\theta}$ for the GLR, SLRG and TRRG methods, while the
sampling distributions $q(\bfv{w})$ varied based on the importance
sampling scheme. For BRG, we used a Beta base distribution, and
applied the Beta slice ratio gradient method. We used antithetic
sampling, i.e. we always sampled $\bfv{w}$ in pairs, which are located
opposite of each other in the distribution. If such a scheme is used,
then any constant baseline $b$ \citep{greensmith2004cv},
which is subtracted from the $\phi(\bfv{w})$ values will cancel out
from the opposite pairs, and the effect of such baselines can be
ignored. We did not use a weight decay. For TRRG, $c=0.5$, and for
BRG, $\alpha=1.1$ in all cases. We used a CPU cluster for our
experiments. Biped tasks were run on 33 cores, and cart-pole tasks were
run on 5 cores. All tasks were run for 2000 policy improvement
iterations (gradient steps), and repeated for several different random
number seeds (details in tables). Because the samples from $q(x)$ do
not correspond to the objective $\expectw{\p{x;\theta}}{\phi(x)}$, we
separately evaluated the performance by sampling from $\p{x;\theta}$
after every 10 iterations. Note that this was done only for evaluation
purposes, and did not have any effect on the learning.

\paragraph{Cart-pole setup} State dimension: 5; Action dimension: 1;
Policy: neural network with one hidden layer with 10 neurons and tanh
activations, total number of parameters $\bfv{w}$: 71; Optimizer:
basic stochastic gradient ascent with one learning rate parameter;
Number of samples per iteration: 32; Std $\sigma$ of Gaussian: 0.5.
In addition to the standard cart-pole task, we considered a setting where
we artificially add noise onto the $\phi(\bfv{w})$ values to simulate
a setting where the rewards can only be observed stochastically, and
test how our importance sampling methods cope with such noisy measurements.
There are additional details in the table and figure captions.

\paragraph{Biped walker setup} State dimension: 24; Action dimension:
4; Policy: neural network with two hidden layers with 40 neurons each
and tanh activations, total number of parameters $\bfv{w}$: 2804;
Optimizer: Adam with $\beta_1=0.99$ and $\beta_2=0.999$; Number of
samples per iteration: 256; Std $\sigma$ of Gaussian: 0.04. We used
reward normalization \citep{mania2018simplers}, which is a technique to
ensure that scale of the rewards stays roughly constant by normalizing
these with the standard deviation of the sampled returns $\phi$. This
appeared to perform better for GLR than rank standardization as used
in \citep{salimans2017oaies}.

\paragraph{Results} The results are in the tables and figures. The
errorbars in the tables correspond to the sample standard deviation
(so divide by the square root of the sample size to obtain a
confidence interval), while the errorbars in the figures are already
the standard deviation of the mean. The results act as a sanity check
and show that our methods do work, while as expected the difference
with standard GLR is small, because the improvement in accuracy is
modest. On the other hand, the cart-pole swing-up experiments show
that using the slice ratio gradient method allows the Beta base
distribution to be competitive with Gaussian distributions. The
experiments also show that SLRG can indeed have trouble with systems
with a low stochasticty, e.g. the cart-pole. Moreover, the results
confirm that our methods reduce gradient variance in stochastic
settings. An important topic of future work will be finding distributions,
which outperform Gaussians, and importance sampling techniques like our
slice ratio gradient method will be crucial in such a pursuit.

\begin{table*}
  \caption{Cart-pole swing-up and balancing, no added noise, 32 samples per batch,
    SGD optimizer, average reward over whole training run, 10 experimental runs for each setting}
\vskip 0.1in
\label{cpolent32}
\centering
\begin{tabular}{rlllll}%{lcccccccccc}
\toprule
Learn. rate  & 0.001 & 0.003 & 0.005 & 0.008 & 0.01\\
\midrule
  GLR & 492.7  $\pm$  15.6 & 694.4  $\pm$  43.2 & 716.9  $\pm$  77.4 & 792.6  $\pm$  31.9
                                               & 782.9  $\pm$  43.1\\
  SLRG & 439.0  $\pm$  12.2 & 580.3  $\pm$  51.4 & 664.9  $\pm$  55.2 & 763.2  $\pm$  52.1
                                               & 747.4  $\pm$  72.7\\
  TRRG & 464.1  $\pm$  5.6 & 676.6  $\pm$  52.4 & 771.2  $\pm$  31.7 & 809.5  $\pm$  21.7
                                               & 747.9  $\pm$  102.1\\
\bottomrule
\end{tabular}
\vskip -0.1in
\end{table*}

\begin{table*}
  \caption{Cart-pole swing-up and balancing, no added noise, 32 samples per batch,
    SGD optimizer, average reward of last 100 parameter values, 10 experimental runs for each
  setting}
\vskip 0.1in
\label{cpolent32}
\centering
\begin{tabular}{rlllll}%{lcccccccccc}
\toprule
Learn. rate  & 0.001 & 0.003 & 0.005 & 0.008 & 0.01\\
\midrule
  GLR & 596.7  $\pm$  39.7 & 881.2  $\pm$  15.3 & 840.1  $\pm$  103.2 & 904.4  $\pm$  3.4
                                               & 889.6  $\pm$  47.4\\
  SLRG & 548.8  $\pm$  7.6 & 723.1  $\pm$  133.8 & 845.4  $\pm$  82.0 & 887.0  $\pm$  57.0
                                               & 895.8  $\pm$  24.7\\
  TRRG & 561.8  $\pm$  16.0 & 867.1  $\pm$  67.4 & 901.1  $\pm$  3.5 & 905.8  $\pm$  1.4
                                               & 840.0  $\pm$  136.0\\
\bottomrule
\end{tabular}
\vskip -0.1in
\end{table*}

\begin{table*}
  \caption{Cart-pole swing-up and balancing, 90 added noise standard deviation, 32
    samples per batch, SGD optimizer, average reward over whole
    training run, the number of experimental runs for GLR, SLRG, TRRG
    were [10,10,50,50,10] for the learning rates from left to right
    respectively, and 20 runs for BRG in all cases.}
\vskip 0.1in
\label{cpolent32Noise}
\centering
\begin{tabular}{rlllll}%{lcccccccccc}
\toprule
Learn. rate  & 0.001 & 0.003 & 0.005 & 0.008 & 0.01\\
  \midrule
  GLR & 519.4  $\pm$  36.9 & 668.1  $\pm$  72.6 & 702.7  $\pm$  66.3 & 690.3  $\pm$  60.8
                                               & 637.4  $\pm$  42.8 \\
  SLRG & 459.7  $\pm$  10.4 & 608.5  $\pm$  51.2 & 668.9  $\pm$  70.8 & 710.5  $\pm$  62.6
                                               & 696.1  $\pm$  64.9 \\
  TRRG & 485.5  $\pm$  8.0 & 658.0  $\pm$  64.7 & 708.1  $\pm$  64.8 & 699.2  $\pm$  73.7
                                               & 682.8  $\pm$  71.4 \\
  BRG & 409.1  $\pm$  12.5 & 531.3  $\pm$  16.4 & 600.8  $\pm$  54.4 & 662.7  $\pm$  70.6
                                               & 723.0  $\pm$  67.6 \\
\bottomrule
\end{tabular}
\vskip -0.1in
\end{table*}

\begin{table*}
  \caption{Cart-pole swing-up and balancing, 90 added noise, 32
    samples per batch, SGD optimizer, average reward of last 100
    parameter values, the number of experimental runs for GLR, SLRG,
    TRRG were [10,10,50,50,10] for the learning rates from left to
    right respectively, and 20 runs for BRG in all cases.}
\vskip 0.1in
\label{cpolent32}
\centering
\begin{tabular}{rlllll}%{lcccccccccc}
\toprule
Learn. rate  & 0.001 & 0.003 & 0.005 & 0.008 & 0.01\\
\midrule
  GLR & 639.2  $\pm$  81.6 & 807.6  $\pm$  101.2 & 834.0  $\pm$  82.8 & 810.2  $\pm$  91.2
                                               & 729.2  $\pm$  103.8 \\
  SLRG & 552.0  $\pm$  5.9 & 771.0  $\pm$  115.8 & 810.4  $\pm$  108.9 & 845.1  $\pm$  74.2
                                               & 815.0  $\pm$  83.3 \\
  TRRG & 574.1  $\pm$  21.4 & 818.1  $\pm$  107.8 & 837.0  $\pm$  86.0 & 814.0  $\pm$  99.1
                                               & 794.6  $\pm$  121.0 \\
  BRG & 535.8  $\pm$  8.9 & 626.6  $\pm$  69.9 & 736.8  $\pm$  123.0 & 792.7  $\pm$  115.0
                                               & 873.5  $\pm$  66.3 \\
\bottomrule
\end{tabular}
\vskip -0.1in
\end{table*}

\begin{table*}
  \caption{Biped walker, 256 samples per batch, each parameter sample
    averaged over 4 episodes, Adam optimizer, reward scaled by
    standard deviation of rewards, average reward of whole training
    run, the number of experimental runs were [20,60,40,40,20] for the
    learning rates from left to right respectively}
\vskip 0.1in
\label{cpolent32}
\centering
\begin{tabular}{rlllll}%{lcccccccccc}
\toprule
Learn. rate  & 0.005 & 0.01 & 0.015 & 0.02 & 0.04\\
\midrule
  GLR & 14.6  $\pm$  30.0 & 223.2  $\pm$  62.1 & 264.2  $\pm$  45.7 & 253.0  $\pm$  51.6
                                             & 260.9  $\pm$  56.6 \\
  TRRG & 31.6  $\pm$  42.7 & 230.0  $\pm$  39.5 & 250.2  $\pm$  43.9 & 257.4  $\pm$  46.0
                                             & 251.2  $\pm$  45.5 \\
\bottomrule
\end{tabular}
\vskip -0.1in
\end{table*}

\begin{table*}
  \caption{Biped walker, 256 samples per batch, each parameter sample averaged
    over 4 episodes,
  Adam optimizer, reward scaled by standard deviation of rewards, average reward of last 100 parameter values, the number of experimental runs were [20,60,40,40,20] for the
    learning rates from left to right respectively}
  \vskip 0.1in
\label{cpolent32}
\centering
\begin{tabular}{rlllll}%{lcccccccccc}
\toprule
Learn. rate  & 0.005 & 0.01 & 0.015 & 0.02 & 0.04\\
\midrule
  GLR & 38.4  $\pm$  92.2 & 390.4  $\pm$  61.5 & 376.6  $\pm$  48.6 & 345.5  $\pm$  63.4
                                             & 347.9  $\pm$  63.0 \\
  TRRG & 104.1  $\pm$  154.2 & 394.0  $\pm$  34.1 & 363.3  $\pm$  58.3 & 353.8  $\pm$  62.3
                                             & 352.5  $\pm$  63.1 \\
\bottomrule
\end{tabular}
\vskip -0.1in
\end{table*}

\begin{figure*}[!t]
        \centering
	\begin{subfigure}{.49\textwidth}
		\includegraphics[width=\textwidth]{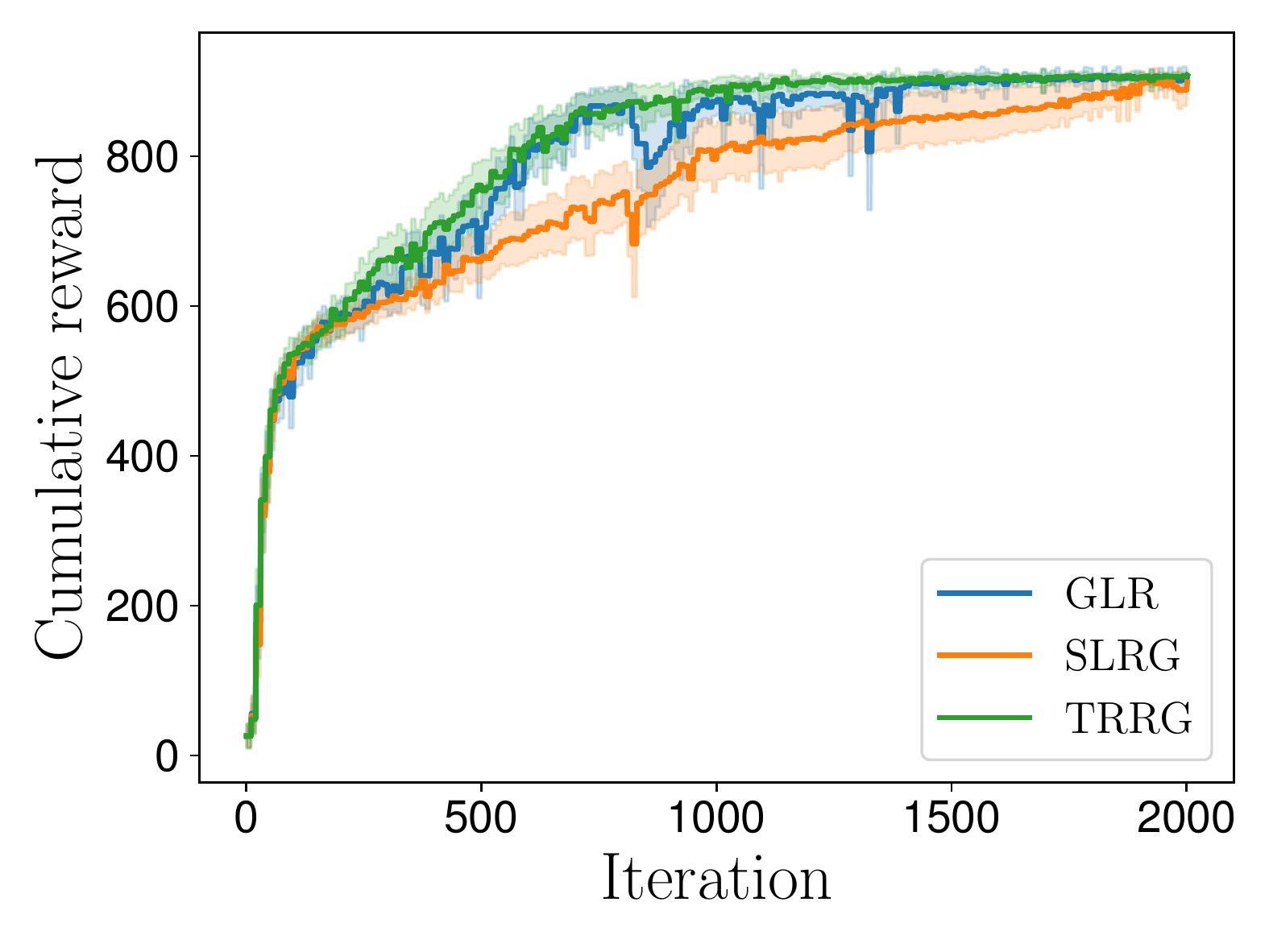}
		\caption{Learning performance}
          \label{}
        \end{subfigure}
%%%%%%%%%%%%%%
	\begin{subfigure}{.49\textwidth}
		\includegraphics[width=\textwidth]{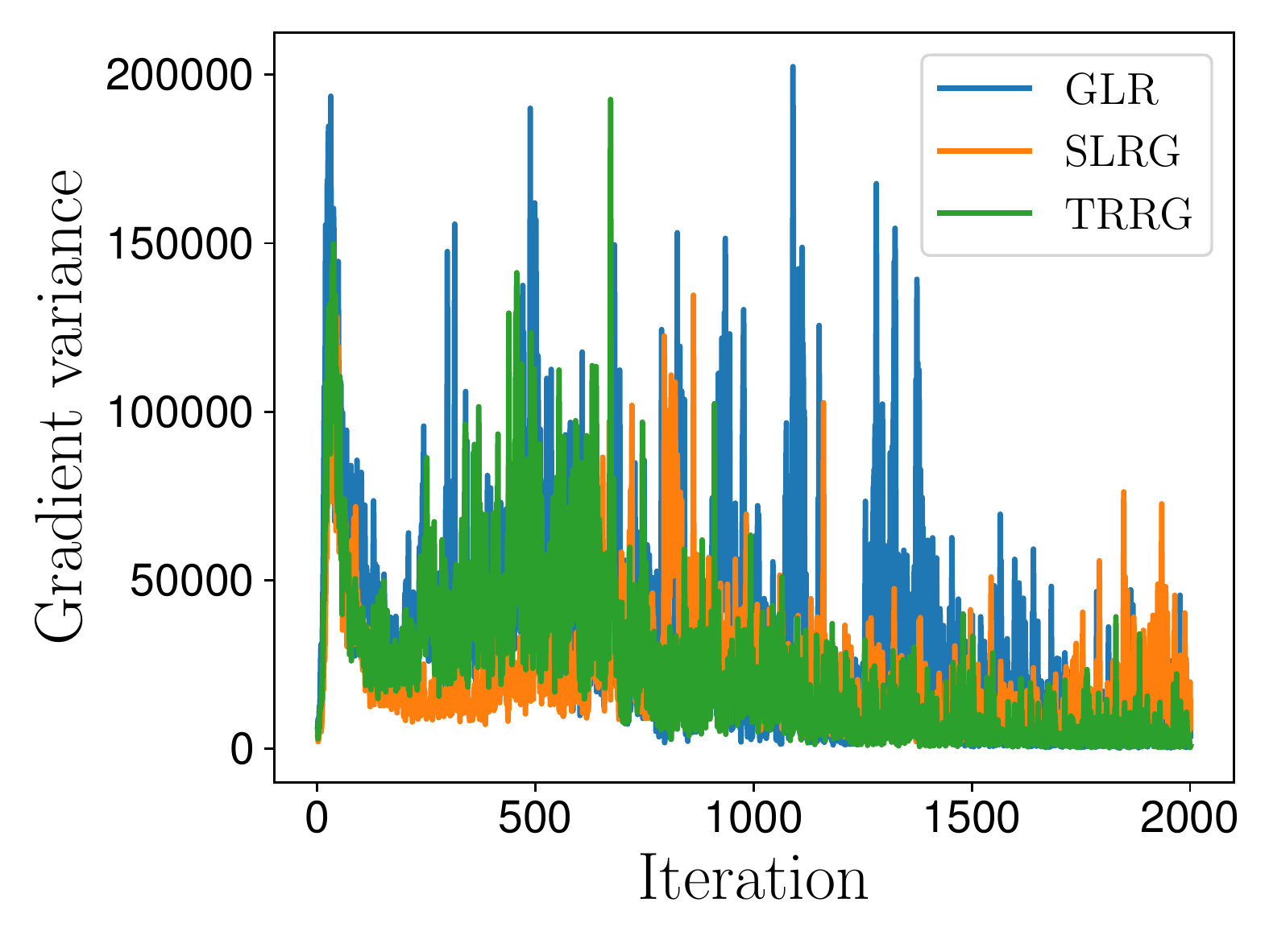}
		\caption{Gradient variance}
          \label{}
	\end{subfigure}
%%%%%%%%%%%%%%
	\caption{Cart-pole swingup, no noise, learning rate: 0.008 for all methods, errorbars show 1 standard deviation of the mean, TRRG's $c=0.5$.}
          \label{cpnonoise}
\end{figure*}

\begin{figure*}[!t]
        \centering
	\begin{subfigure}{.49\textwidth}
		\includegraphics[width=\textwidth]{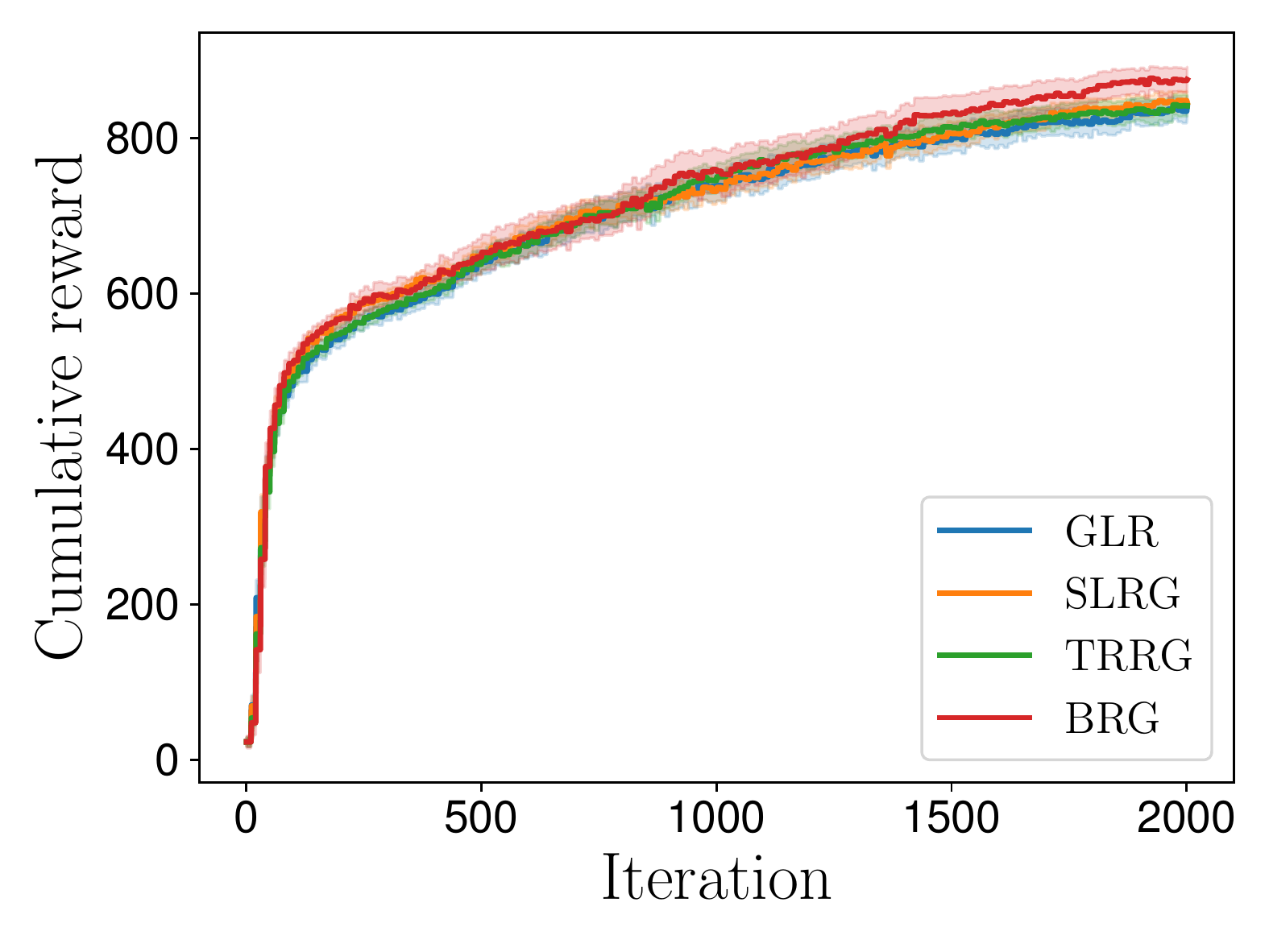}
		\caption{Learning performance}
          \label{}
        \end{subfigure}
%%%%%%%%%%%%%%
	\begin{subfigure}{.49\textwidth}
		\includegraphics[width=\textwidth]{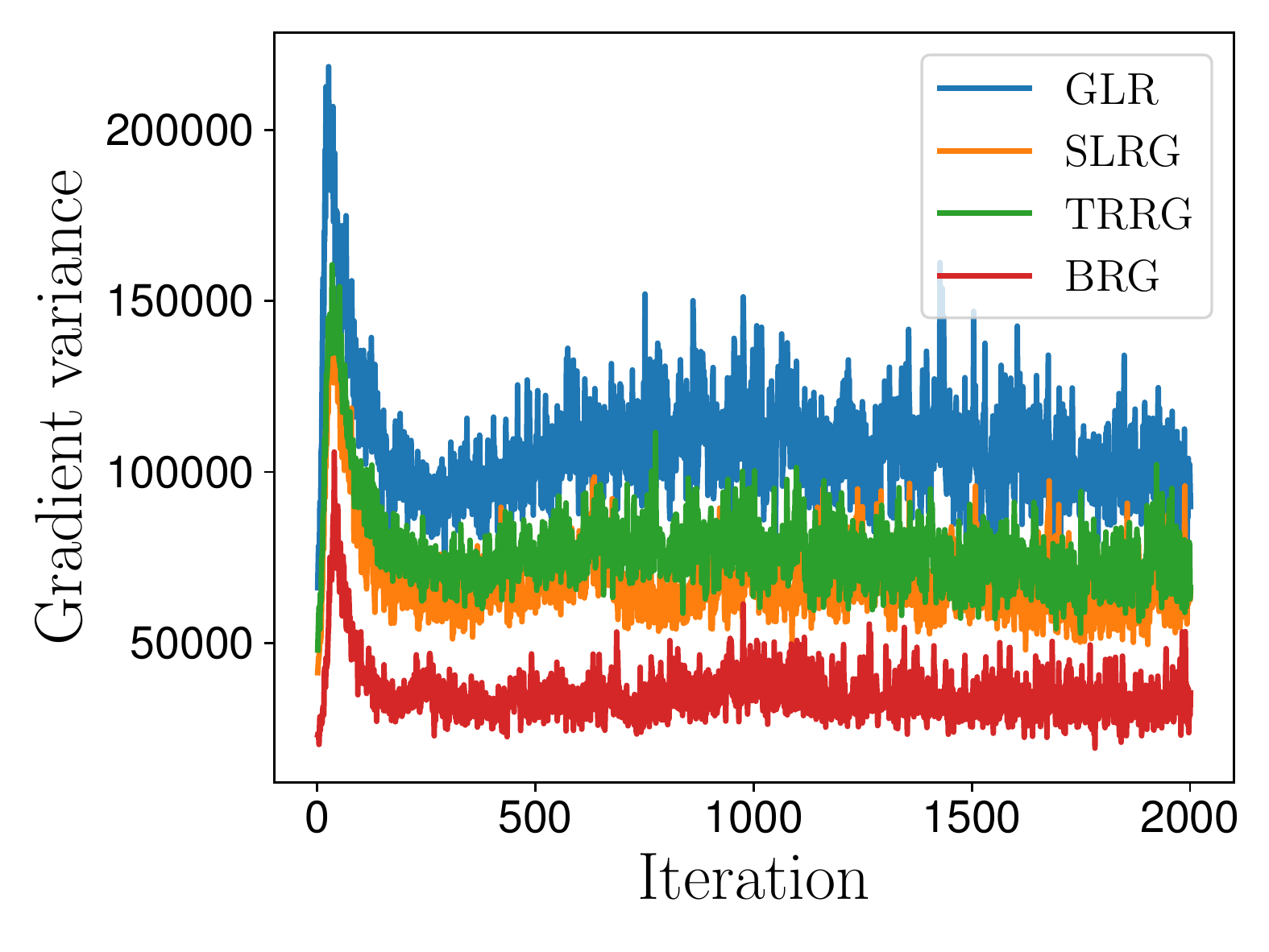}
		\caption{Gradient variance}
          \label{}
	\end{subfigure}
%%%%%%%%%%%%%%
	\caption{Cart-pole swingup; noise with std 90 added on cumulative reward; learning rates: GLR: 0.005, SLRG: 0.008, TRRG:0.005, BRG:0.01; errorbars show 1 standard deviation of the mean, BRG's $\alpha=1.1$, TRRG's $c=0.5$.}
          \label{cpnoise90}
\end{figure*}

\begin{figure*}[!t]
        \centering
	\begin{subfigure}{.32\textwidth}
		\includegraphics[width=\textwidth]{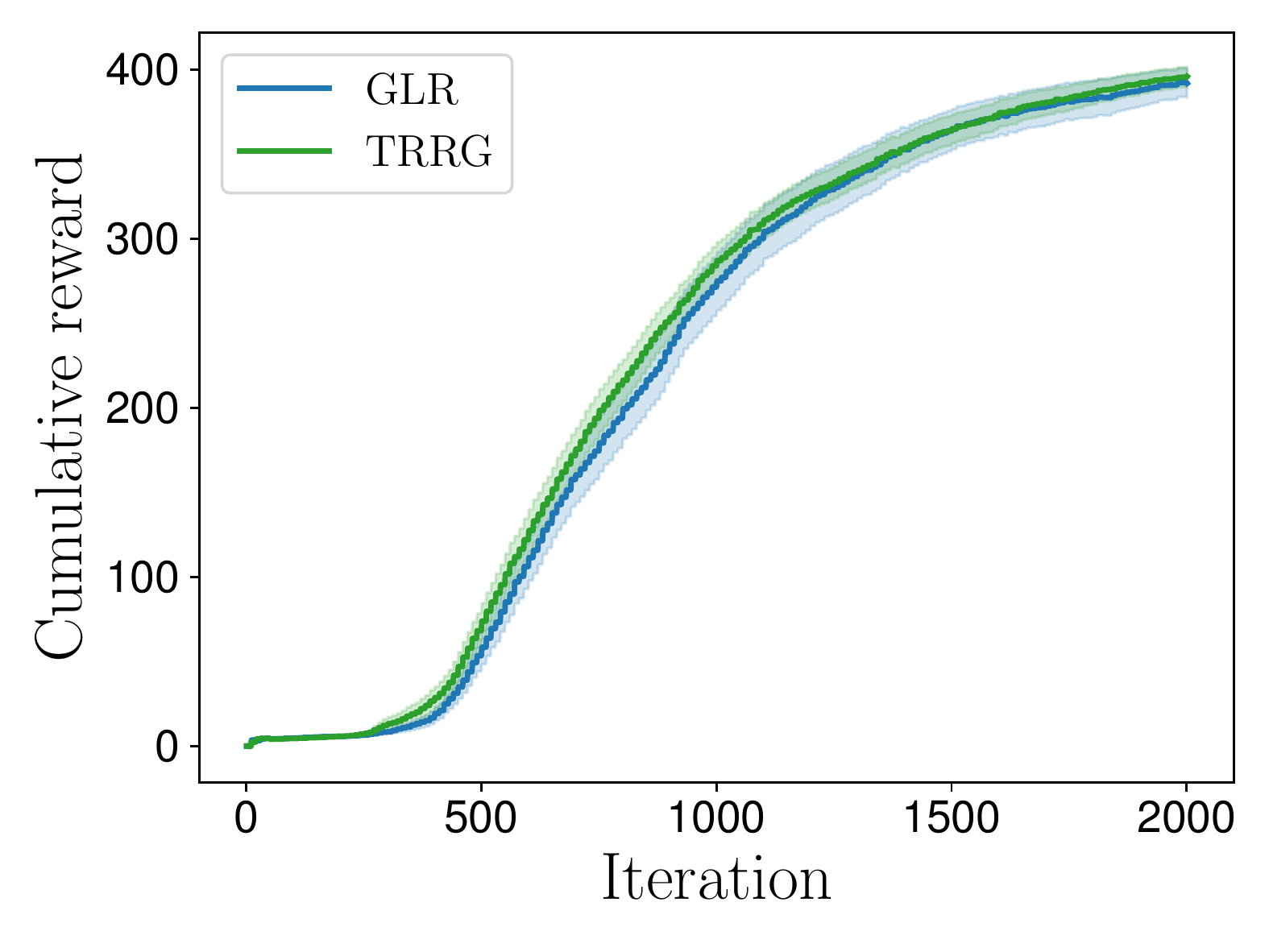}
		\caption{Learning rate: 0.01}
          \label{}
        \end{subfigure}
%%%%%%%%%%%%%%
	\begin{subfigure}{.32\textwidth}
		\includegraphics[width=\textwidth]{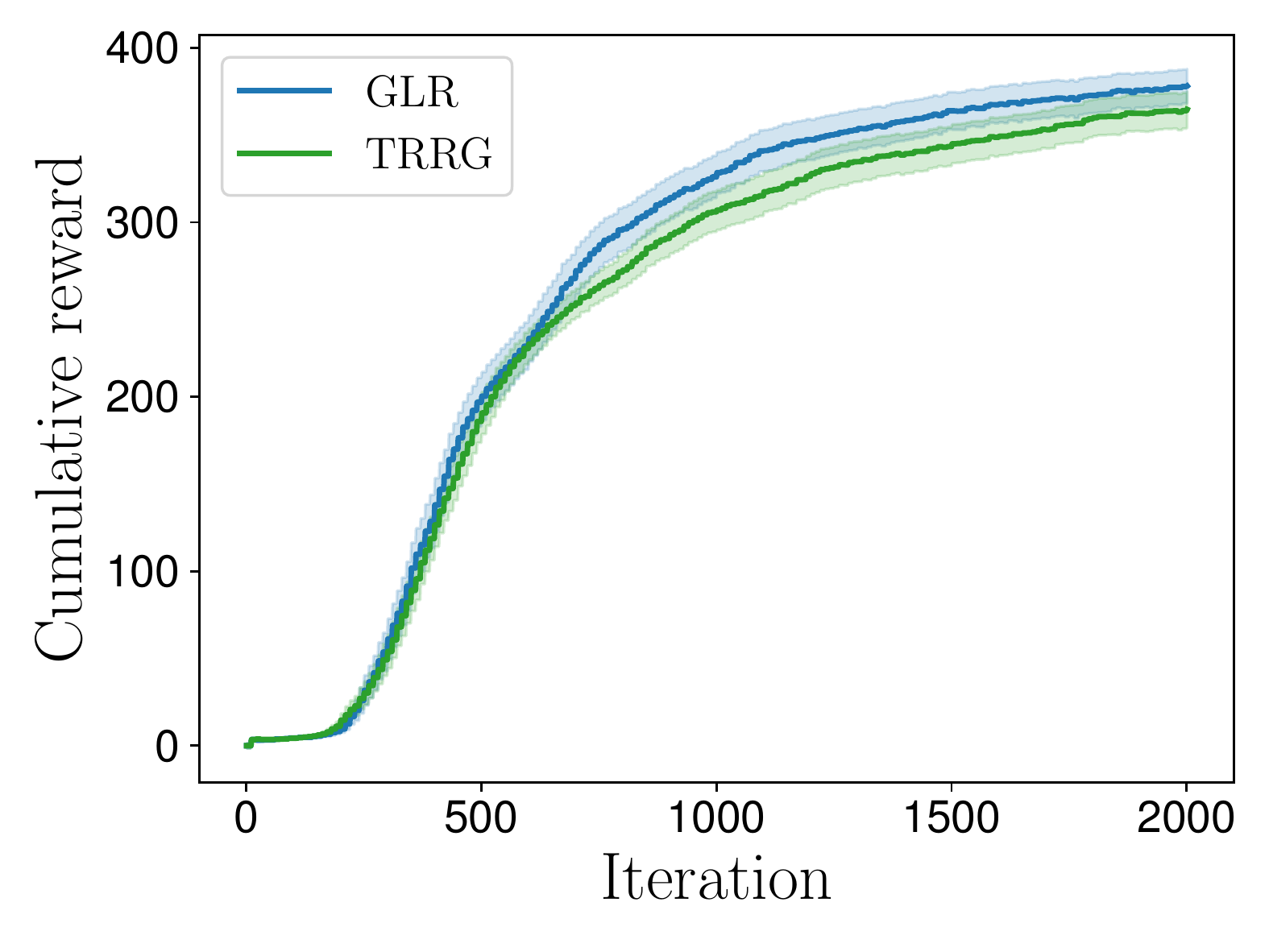}
		\caption{Learning rate: 0.015}
          \label{}
	\end{subfigure}
%%%%%%%%%%%%%%
	\begin{subfigure}{.32\textwidth}
		\includegraphics[width=\textwidth]{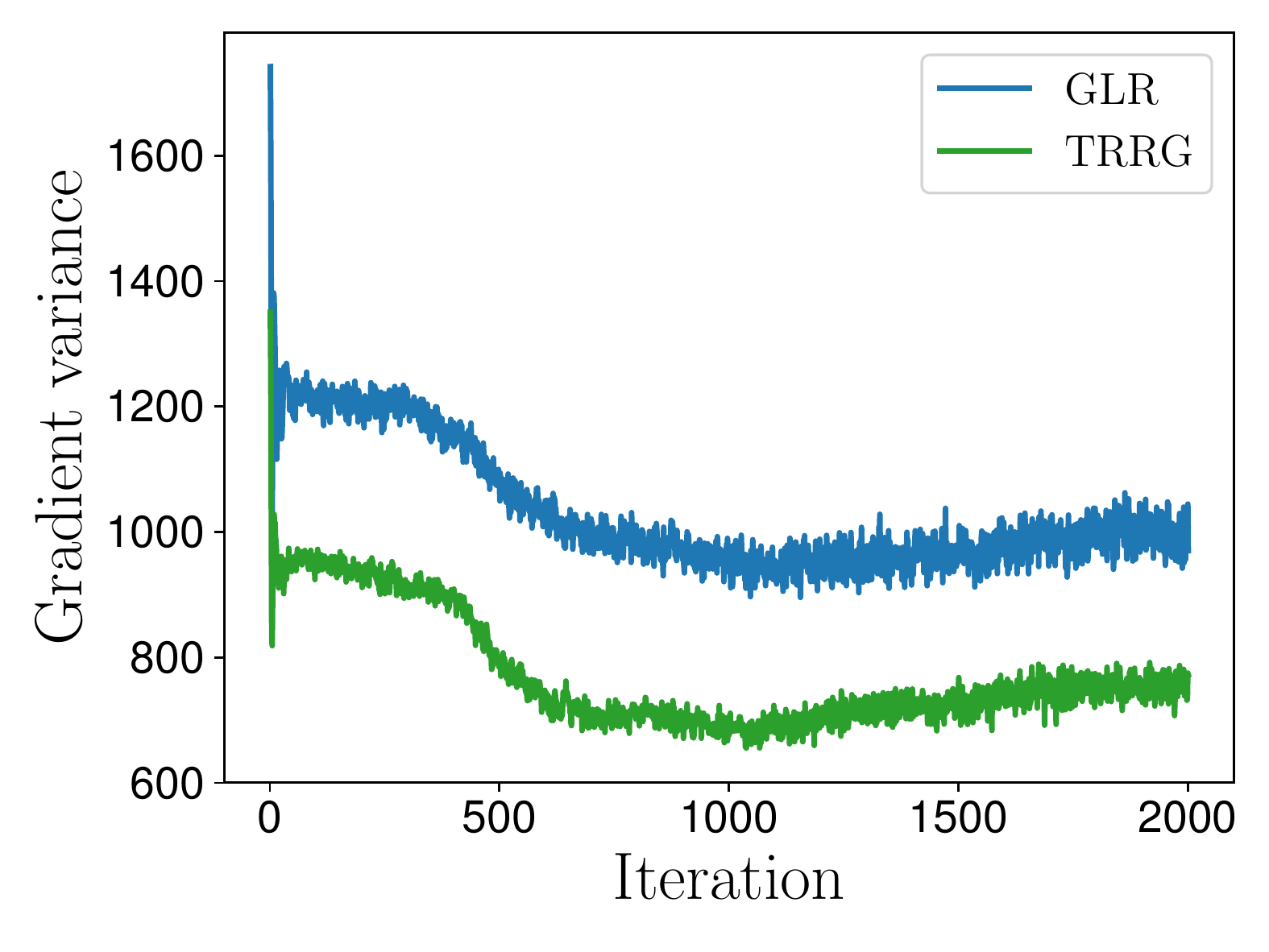}
		\caption{Learning rate: 0.01}
          \label{}
	\end{subfigure}
%%%%%%%%%%%%%%
	\caption{Biped walker; errorbars show 1 standard deviation of
          the mean, each parameter sample averaged over 4 episodes,
          TRRG's $c=0.5$.}
          \label{bipedexp}
\end{figure*}

\begin{figure*}[!t]
        \centering
	\begin{subfigure}{.49\textwidth}
		\includegraphics[width=\textwidth]{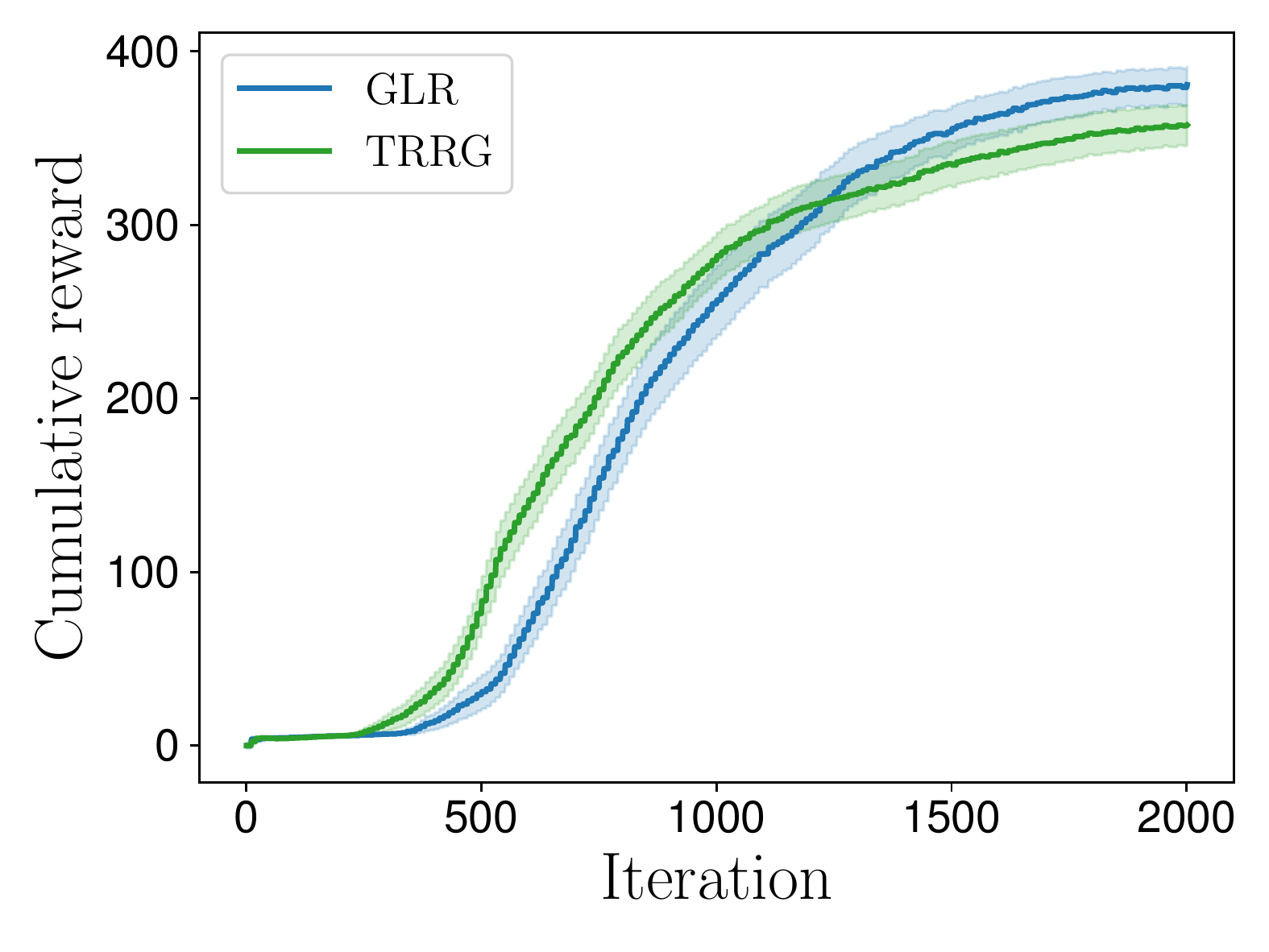}
		\caption{Average results}
          \label{}
        \end{subfigure}
%%%%%%%%%%%%%%
	\begin{subfigure}{.49\textwidth}
		\includegraphics[width=\textwidth]{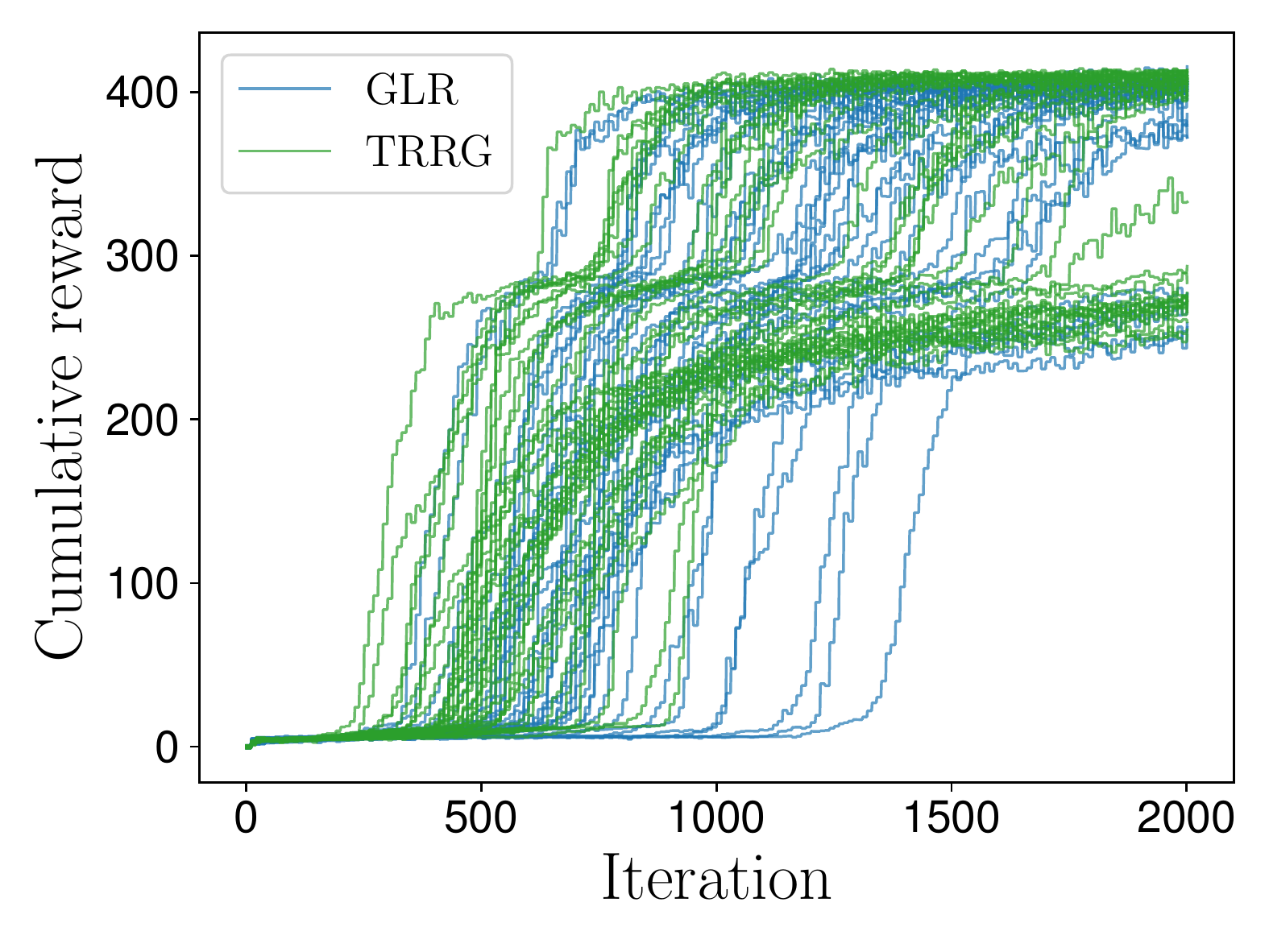}
		\caption{Raw data}
          \label{}
	\end{subfigure}
%%%%%%%%%%%%%%
	\caption{Biped walker; learning rate: 0.01, errorbars show 1
          standard deviation of the mean, each parameter sample from 1
          episode, 40 random number seeds, TRRG's $c=0.5$; the final
          reward was bimodal, and while TRRG learned faster, more
          experiments converged to the lower local minimum.}
          \label{bipedexp}
\end{figure*}

\end{appendices}
\end{document}